\documentclass{article}

    \PassOptionsToPackage{numbers, compress, sort}{natbib}

    \usepackage[preprint]{neurips_2020}

\usepackage[utf8]{inputenc} 
\usepackage[T1]{fontenc}    
\usepackage[colorlinks = true]{hyperref}       
\usepackage{url}            
\usepackage{booktabs}       
\usepackage{amsfonts}       
\usepackage{nicefrac}       
\usepackage{microtype}      
\usepackage{amsmath}
\usepackage[final]{graphicx}
\usepackage{subcaption}
\usepackage[colorinlistoftodos]{todonotes}
\usepackage{threeparttable}
\usepackage{booktabs}
\usepackage{array}
\usepackage{arydshln}
\usepackage{wrapfig}

\newcommand{\FDNet}{\texttt{FD-Net}}
\newcommand{\FDBlock}{\texttt{FD-Block}}
\newcommand{\FDBlocks}{\texttt{FD-Blocks}}
\newcommand{\FDFilters}{\texttt{FD-Filters}}
\newcommand{\FDFilter}{\texttt{FD-Filter}}

\title{Finite Difference Neural Networks: Fast Prediction of Partial Differential Equations}

\author{%
  Zheng Shi \\
  Lehigh University\\
  Bethlehem, PA 18015 \\
  \url{zhs310@lehigh.edu} \\
  \And
  Nur Sila Gulgec \\
  Thornton Tomasetti \\
  San Francisco CA 94105 \\
  \url{sgulgec@gmail.com} \\
  \AND
  Albert S. Berahas \\
  Lehigh University \\
  Bethlehem, PA 18015\\
  \url{albertberahas@gmail.com} \\
  \And
  Shamim N. Pakzad \\
  Lehigh University \\
  Bethlehem, PA 18015\\
  \url{pakzad@lehigh.edu} \\
  \And
  Martin Tak\'a\v c \\
  Lehigh University \\
  Bethlehem, PA 18015\\
  \url{takac.mt@gmail.com} \\
}

\begin{document}

\maketitle

\begin{abstract}
  Discovering the underlying behavior of complex systems is an important topic in many science and engineering disciplines. In this paper, we propose a novel neural network framework, finite difference neural networks (\FDNet), to learn partial differential equations from data. Specifically, our proposed finite difference inspired network is designed to learn the underlying governing partial differential equations from trajectory data, and to iteratively estimate the future dynamical behavior using only a few trainable parameters. We illustrate the performance (predictive power) of our framework on the heat equation, with and without noise and/or forcing, and compare our results to the Forward Euler method. Moreover, we show the advantages of using a Hessian-Free Trust Region method to train the network.
\end{abstract}

\section{Introduction}

Partial differential equations (PDEs) are widely adopted in a plethora of science and engineering fields to explain a variety of phenomena such as heat, diffusion, electrodynamics, fluid dynamics, elasticity, and quantum mechanics, to mention a few. This is primarily due to their ability to model and capture the behavior of complex systems as well as their versatility. However, solving PDEs is far from a trivial task. Often incredible amounts of computing power and time are required to get reasonable results, and the methods used can be complicated and highly-sensitive to the choice of parameters. 

The rapid development in data sensing (collection) and data storage capabilities provides scientists and engineers with another avenue for understanding and making predictions about these phenomena. The massive amounts of data collected from highly complex and multi-dimensional systems have the potential to provide a better understanding of the underlying system dynamics.

Utilizing this abundance of data to solve PDEs has been exploited in several recent studies; see e.g.,  \cite{schmidt2009distilling,bongard2007automated,rudy2017data,schaeffer2017learning,raissi2018hidden,raissi2017physics,chen2018neural,farimani2017deep,long2017pde,pdenet2}. In \cite{schmidt2009distilling,bongard2007automated} and \cite{rudy2017data,schaeffer2017learning}, the authors applied symbolic regression and sparse regression techniques, respectively, to explain nonlinear dynamical systems. In \cite{raissi2018hidden,raissi2017physics}, the authors introduced physics informed neural networks using Gaussian processes. Moreover, Chen et al. \cite{chen2018neural} proposed continuous-depth residual networks and continuous-time latent variable models to train ordinary neural networks. Finally, in \cite{farimani2017deep}, the authors proposed conditional generative adversarial networks to predict solutions for steady state heat conduction and incompressible fluid flow, and, in \cite{long2017pde,pdenet2}, the authors proposed PDE-Net inspired by Wavelet theory to approximate the unknown nonlinear responses of diffusion and convection processes. Possibly the closest work to ours is PDE-Net \cite{long2017pde,pdenet2}, which the authors proposed to learn differential operators by learning convolution filters. The key differentiating features of our approach can be summarized as follows: $(i)$ our approach is computationally efficient since it trains finite difference inspired, small and linear filters; $(ii)$ our network architecture can be adapted and enhanced to learn PDEs with forcing; and, $(iii)$ we use a second-order optimization method to improve accuracy and computation time of training.

In this paper, inspired by finite-difference approximations and residual neural networks \cite{resnet}, we propose a novel neural network framework, finite difference neural networks (\FDNet), to learn the governing partial differential equations from trajectory data, and iteratively estimate future dynamical behavior. Mimicking finite-difference approximations, \FDNet{} employs ``finite-difference’’ block(s) (\FDBlock) with artificial time steps to learn first-, second- and/or higher-order partial derivatives, and thus learn the underlying PDEs from neighboring spatial points over the time horizon. As a proof-of-concept, we deploy our proposed method to learn and predict the underlying dynamics of PDEs using trajectory data from the heat equation in different cases: $(1)$ simple homogeneous heat equation; $(2)$ heat equation with noise; and, $(3)$ heat equation with a forcing term.

Stochastic first-order methods have been very successful in training machine learning models in various applications \cite{bottou2018optimization}. However, there are several drawbacks to using such methods, and it has been shown that, for certain applications, employing stochastic second-order methods can be beneficial \cite{multibatch,subsample,xu2020second,berahas2019quasi,berahas2020investigation}. In this paper, we show that training our networks is one such application; training time can be significantly reduced and the accuracy of the solutions can be drastically improved by using a second-order method. Specifically, we employ a second-order Hessian-Free method, Trust-Region Newton CG \cite{Nocedal,steihaug}.



The paper is organized as follows. In Section \ref{Method-Pre}, we introduce the PDE used in our case study and discuss the four different classes of problems that we investigate. We discuss in detail the fundamentals of \FDNet{} in Section \ref{Method-FD-net}. Extensive numerical results are presented in Section \ref{Experiment}. Finally, in Section \ref{fin_rem}, we make some concluding remarks and discuss avenue for future research.




\section{The Heat Equation}\label{Method-Pre}

Consider a linear partial differential equation (PDE) in canonical form:
\begin{align}
\mathcal {F}(x,t, u,u_t, u_x, u_{xx},u_{xxx}, \dots) = 0, \label{eq1}
\end{align}
where $ \mathcal{F}$ is a linear function of $u$ and its partial derivatives with respect to time and/or space. The objective of our study is to implicitly learn $\mathcal{F}$ given a series of measurements (trajectory data) at specific time and spatial instances, and predict the solutions to the equation throughout the time horizon.

For our case study, we consider the heat equation, one of the most frequently used PDEs in physics, mathematics, engineering and more. The heat equation describes the evolution of heat flow over time in an object \cite{incropera2007fundamentals}. Let $u(x,t)$ denote the temperature at a spatial point $x$ and time $t$. The heat equation for a 1-D bar of length $L$ can be expressed as
\begin{align} 
 \frac{\partial u}{\partial t} = \beta \frac{\partial^2 u}{\partial x^2}, \label{eq2}
\end{align}
where $\beta$ is a rate of diffusion of the material. Under the assumption of perfectly insulated boundaries, the boundary conditions (BCs) can be expressed as
\begin{align}
u(0,t) = 0, \qquad u(L,t) = 0. \label{eq3}
\end{align}
We consider the following initial condition (IC)
\begin{align}
u(x,0) &= \sum_{i=1}^N C_i \sin \left(\frac{i \pi x}{L}\right),\label{eq4}
\end{align}
where $C_i \in \mathbb{R}$ for $i \in \{1,2,...,N\}$.
The exact solution of \eqref{eq2} with BCs \eqref{eq3} and IC \eqref{eq4} can be expressed as
\begin{align}
u(x,t) &= \sum_{i=1}^N C_i \sin \left(\frac{i \pi x}{L}\right) e^ {- \beta  (i \pi/L)^2 t}.\label{eq5}
\end{align}

The reasons we choose this PDE are three-fold: $(1)$ it is an extensively used PDE that will allow us to investigate the merits and limitations of our proposed approach; $(2)$ although the PDE is simple, it has several characteristics that are interesting to investigate (e.g., first- and second-order derivatives) and the behavior of the PDE can be complex in the presence of noise and/or a forcing term; and $(3)$ we can derive the exact solution.

Given $x \in [0,L]$ and $t \in [0,T]$, \eqref{eq2} can be approximately solved via forward Euler method \cite{euler1824institutionum}. To this end, the domain is discretized (both in $x$ and $t$) and $u(x,t)$ is computed recursively as follows:
\begin{align}
u(x,t+\Delta t) = u(x,t) + \delta \left[u(x+\Delta x, t) -2 u(x,t) + u(x-\Delta x, t)\right], \quad \text{where $\delta = \beta \tfrac{ \Delta t}{(\Delta x)^2}$.}\label{eq6}
\end{align}
The performance of the Euler method, in terms of the accuracy of the solution, is highly dependent on the choice of the granularity of the discretization, both in time ($\Delta t$) and space ($\Delta x$). Specifically, the Euler method fails to generate accurate approximations, and may even diverge, if $\Delta t$ and $\Delta x$ do not satisfy $ \delta \leq 0.5$, known as the stability criterion \cite{olsen2011numerical}. We should note that of course higher-order numerical procedures (or even implicit schemes) exist and could be used to solve \eqref{eq2} and mitigate some of the stability problems at the cost of more complex updates.

In addition to the instability associated with sparsely discretized time steps, in real-life applications the measurements of $u(x,t)$ are often contaminated with noise, e.g., Gaussian noise $\varepsilon \sim \mathcal{N}(\mu,\sigma^2)$, which can severely impact the stability and quality of the solutions. Moreover, the PDE could also have a forcing term, e.g.,
\begin{align}
\frac{\partial u}{\partial t} = \beta \frac{\partial^2 u}{\partial x^2} + f(x), \qquad \text{where } \;
f(x) = \sum_{i=1}^{N} D_i \sin\left(\frac{i\pi x}{L}\right), \label{eq7}
\end{align}
and $D_i \in \mathbb{R}$ for $i \in \{1,2,...,N\}$. See Appendix \ref{app_exact} for the exact solution of \eqref{eq7}.

The challenges that arise from instability, noisy measurements and forcing terms can make conventional approaches, such as the Euler method, vulnerable and result in inaccurate approximations. These challenges have inspired researchers in the fields of computational mathematics and machine learning to develop solution techniques that utilize the power of deep neural networks and exploit the massive amounts of measurements (i.e., trajectory data) that are readily available to solve PDEs and make predictions.

In this paper we investigate the performance of our proposed method, \FDNet{}, on the heat equation for all four aforementioned cases: $(1)$ stable; $(2)$ unstable; $(3)$ noisy; and, $(4)$ forced.

\section{Fundamentals of \FDNet{}} \label{Method-FD-net}

In this section, we describe the fundamental components of \FDNet{}.

\begin{figure}
  \centering
  \includegraphics[width=0.95\textwidth]{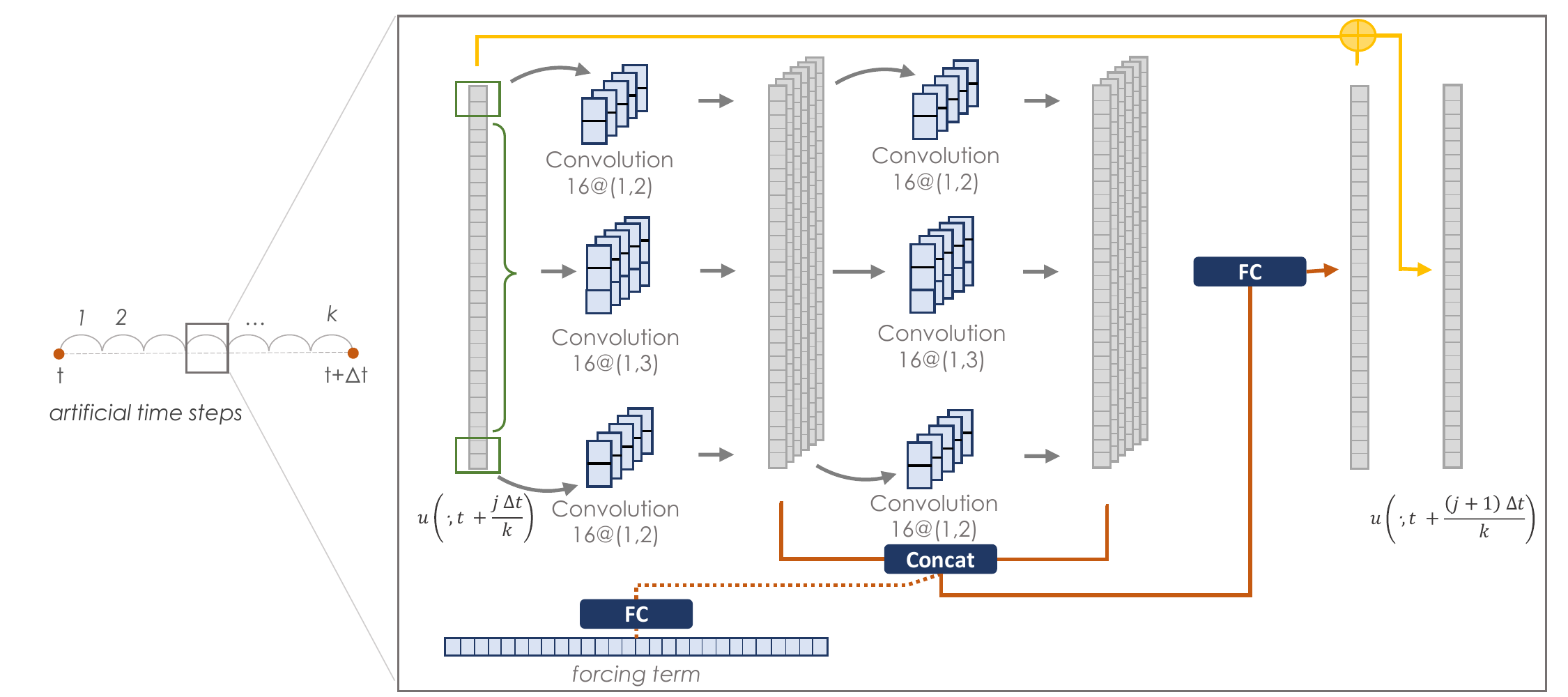}
  \caption{An illustration of  \FDBlock{} and the artificial time step. In this particular instance, there are $k$ \FDBlocks{} defined in the network and thus $k-1$ artificial time steps on $[t,t+\Delta t]$. For each \FDBlock{}, there are $16$ \FDFilters{}, two groups of convolutional layers, an FC layer, a forcing function representation and a skip connection. At the (artificial) time step $t + \frac{j \Delta t}{k}$ for $j = 0,1,...,k-1$, the input $u(\cdot,t+\frac{j\Delta t}{k})$ is passed through the convolutional layers to learn the first- and second-order partial derivatives. Concatenated with the representation of the forcing function, the outputs are then passed through the FC layer with the skip connection to predict the function behavior at $t+\frac{(j+1)\Delta t}{k}$.}
  \label{fig2}
\end{figure}

The building blocks of \FDNet{} are \FDBlocks{}, whose design is inspired by finite-difference approximations of partial derivatives. Figure \ref{fig2} shows an instance of \FDBlock{}. An \FDBlock{} is a deep residual learning block \cite{resnet} 
that aims to learn the evolution of a dynamical system for one artificial time step on $[t,t+\Delta t]$. It is composed of groups of convolutional layers, a fully connected (FC) layer, and a multi-step skip connection. 

Specifically, for each group of convolutional layers, a certain number of ``finite-difference’’ filters (\FDFilters) are defined in space: for $x \in \{\Delta x, 2\Delta x,...,L-\Delta x\}$, the size of the filter is three (one parameter for $x$ itself, one for its left neighbor and one for its right neighbor); for the boundaries, i.e., $x = 0$ or $x = L$, the size of a filter is two as there is only one neighbor, either on the left or right. The outputs of one group of layers with \FDFilters{} are concatenated to form a learned representation of partial derivatives of a certain order. In order to capture and mimic higher-order partial derivatives, multiple groups of convolutional layers with such filters are employed. The representation from a previous group is used as input of the subsequent group in order to learn a higher-order representation. The learned representations of partial derivatives, by all groups, are then concatenated and passed as input to the FC layer in order to learn the evolution (dynamics). Next, a skip connection is applied and the network proceeds to the following artificial time step. 

Moreover, to imitate finite-difference approximations and to capture the behavior of linear equations, \FDNet{} defines the parameters of each layer without bias terms, and the outputs of the layers without applying nonlinear activation functions. In addition to the main architecture of the \FDBlock{}, \FDNet{} constructs a learnable representation via an FC layer and concatenates it with the outputs of the convolutional layers to learn forcing functions that are potentially present in the PDE.

Overall, \FDNet{} is formed by stacking multiple \FDBlocks{} sequentially in order to produce an approximate solution of the PDE at $t+\Delta t$ given a solution at $t$. Incorporating $k\left(>1\right)$ \FDBlocks{} introduces $k-1$ artificial time steps between $t$ and $t+\Delta t$ to \FDNet{}, which enhances the learning capability of \FDNet{}, especially when $\Delta t$ is large. We discuss this further in Section \ref{Experiment}. The number of \FDBlocks{} is the first hyper-parameter of \FDNet{}. 

Furthermore, instead of defining distinct \FDFilters{} for each \FDBlock{}, \FDNet{} shares the same \FDFilters{} along the sequence. As a result, the size of an instance of our networks does not depend on the number of \FDBlocks{} but rather on the number and size of the \FDFilters. \FDNet{} uses the same number of \FDFilters{} across all convolutional layers for a consistent input/output shape. We use this quantity as the second hyper-parameter of \FDNet{} and refer to it as "the number(s) of \FDFilters{}".

Table \ref{tab:1} shows the sizes of the networks, for different numbers of \FDFilters{}, used to learn PDEs with first- and second-order partial derivatives. Our experiments (see Section \ref{Experiment}) indicate that $16$ \FDFilters{} are sufficient for \FDNet{} to produce predictions with high precision for our case-study PDE for all aforementioned cases.

\begin{table}[h!]
\centering 
\caption{Number of Parameters in \FDNet{}.}
\vfill
\label{tab:1}
\begin{threeparttable}
\begin{tabular}{l c c c c c}
\toprule
 \# \FDFilters{}    \; &  4 \; & 8 \;& 16 \; & 32\; & 64\; \\
 \hdashline
 \# Parameters w/o forcing\tnote{$\dagger$}\  \;  & 148 \;& 520 \; & 1936\; & 7456\; & 29248\; \\ 
 \hdashline
 \# Parameters\tnote{$\ddagger$}\  \;  & 468 \;& 840 \; & 2256\; & 7776\; & 29568\; \\ 
 \bottomrule
\end{tabular}
\begin{tablenotes}\scriptsize
\item[$\dagger$] does not count parameters for learning forcing function.
\item[$\ddagger$] includes all parameters in an instance of \FDNet{}, the forcing function is in the form of \eqref{eq7} with $N = 10$, $x \in [0,\pi]$ and $\Delta x = 0.1$.
\end{tablenotes}
\end{threeparttable}
\end{table}

\section{Numerical Experiments} \label{Experiment}
In this section, we present numerical experiments to demonstrate the empirical performance of \FDNet{} on the heat equation under the four scenarios described in Section \ref{Method-Pre}: $(1)$ stable case; $(2)$ unstable case; $(3)$ noisy case; and, $(4)$ forcing case. We first describe the data we used in our experiments, then discuss the optimization methods employed for training the networks and finally show numerical results.

The main goal of the experiments is to study if \FDNet{} is capable of making accurate predictions throughout the time horizon by solely relying on trajectory data and iteratively learning the short-term (i.e., between $t$ and $t+\Delta t$) evolutions. We illustrate the training and testing (prediction) performance of our proposed approach, compare the predictions made by \FDNet{} with the approximate solutions generated by the forward Euler method, demonstrate the advantages of training our networks with Trust-Region (TR) Newton CG method, and investigate the sensitivity of our networks to the hyper-parameters. For brevity, we show only a subset of our results in the main paper and defer the full experimental results to Appendix \ref{app_num_res}.

\subsection{Data, Training and Testing} \label{Experiment-1}
For each aforementioned case, we generated synthetic data using the exact solutions to the heat equations; see Section \ref{Method-Pre}. Specifically, for each case, we generated 200 different trajectories, each with a randomly generated initial condition, i.e., $C_i \sim \mathcal{N}(0,1)$, for $i \in \{1,2,...,10\}$ in \eqref{eq4}. We considered the 1D bar of length $L=\pi$, and the rate of diffusion parameter was set to $\beta = 2\cdot 10^{-4}$. We set the spatial discretization to $\Delta x = 0.1$ on $[0,\pi]$, the time horizon as $[0,1000]$, and the temporal discretization to $\Delta t = 1$ (namely, $\delta = 0.02 < 0.5$ in \eqref{eq6}), for the stable, noisy and forcing cases, and $\Delta t = 200$ (namely, $\delta = 4 > 0.5$ in \eqref{eq6}), for the unstable case. For the noisy case, we considered multiplicative noise of the form $u(x,t) = u(x,t)(1+\gamma_i\epsilon_{x,t})$ for $i \in\{1,2,3\}$, where $\epsilon_{x,t} \sim \mathcal{N}(0,1)$, and  $\gamma_1=10^{-8}$ (low), $\gamma_2 = 10^{-4}$ (medium) and $\gamma_3 = 10^{-2}$ (high). For the case with a forcing function, we generated the function with $D_i \sim \mathcal{N}(0,1)$ for $i \in \{1,2,...,10\}$ in \eqref{eq7} and applied it to the PDEs of all initial conditions.

Let
\begin{align*}
    A = \left\{u_s(x,t)\; | \;s \in S, \; x \in \{0, \Delta x,...,L\}, \; t \in \{0,\Delta t,...,T\}\right\}
\end{align*}
denote a randomly generated data set, where $S$ is the index set of ICs, $u_s(x,t)$ is the data (of measurements) and $\{u_s(x,t)\}_{x,t}$ is the trajectory data for a specific IC. We randomly selected $150$ ICs as our training set, and the remainder ($50$ ICs) were used for testing purposes. We denote $S_{\text{train}}$ and $S_{\text{test}}$ the subsets of indices of ICs for training and testing, respectively.


For training purposes, we adopted a ``one-step ahead’’ procedure. Let 
\begin{align*}
    A_{\text{train}} = \left\{ \left(u_s(x,t), u_s(x,t+\Delta t)\right)\;|\; s\in S_{\text{train}},\;x\in \{0,\Delta x,...,L\},\;t\in \{0,\Delta t,...,T-\Delta t\}\right\}
\end{align*}
define the training data set, where $\left(u_s(x,t), u_s(x,t+\Delta t) \right)$ is a training tuple (sample), $u_s(x,t)$ is the input and $u_s(x,t+\Delta t)$ is the target. We defined the MSE loss of the stochastic mini-batch as
\begin{align}
    \text{MSE}_{\text{mini}} = \frac{1}{|A_{\text{mini}}|} \sum_{s,x,t} \left(u_s(x,t+\Delta t) - \tilde{u}_s(x,t+\Delta t)\right)^2, \label{eq8}
\end{align}
where $A_{\text{mini}} \subseteq A_{\text{train}}$ is a mini-batch and $\tilde{u}_s(x,t+\Delta t)$ is the output of a network. On the other hand, for testing, we used a ``1000-step’’ sequential prediction procedure (and refer to it as "1000-step prediction”). Let 
\begin{align*}
    A_{\text{test}} = \left\{ \left(u_s(x,0), u_s(x,T)\right)\;|\; s\in S_{\text{test}},\;x\in \{0,\Delta x,...,L\}\right\}
\end{align*}
define the testing data set, $u_s(x,0)$ be the input and $u_s(x,T)$ be the target. We used $u_s(x,0)$ as the initial input and sequentially made predictions through the time horizon until reaching $T$, where the final prediction $\tilde{u}_s(x,T)$ was made. Specifically, $u_s(x,0)$ was used as the input to make the next prediction $\tilde{u}_s(x,\Delta t)$, which in turn was used as the input to make the next prediction $\tilde{u}_s(x,2\Delta t)$, and this was repeated throughout the whole time horizon. The error metric we used was MSE and was defined as
\begin{align}
    \text{MSE}_{\text{test}} = \frac{1}{|A_{\text{test}}|} \sum_{s,x} \left(u_s(x,T) - \tilde{u}_s(x,T)\right)^2. \label{eq9}
\end{align} 


For each case, we configured the networks with different numbers of \FDBlocks{} and \FDFilters{} and used two optimization methods, ADAM (with learning rates $10^{-3}$ and $10^{-4}$) \cite{Adam} and Trust-Region (TR) Newton CG method \cite{Nocedal}, both with mini-batch sizes of $64$. We prescribed a fixed budget of $100$ iterations for the TR method on the stable, noisy and forcing cases and $300$ iterations on the unstable case, but allowed the ADAM algorithm to run for $12000$ iterations on all cases. For each configuration, algorithm and case, we used $10$ random seeds to initialize network parameters and to generate stochastic mini-batches. 

\subsection{Results and Discussion}
In this section, we present numerical results and discuss the strengths and limitations of \FDNet{}. We consider all the aforementioned cases: $(1)$ stable; $(2)$ unstable; $(3)$ noisy (medium); and $(4)$ forcing. For brevity, among all configurations investigated, we show results for the best configuration for each case (from an average performance perspective given the budget). Specifically, we show results for $1$ \FDBlock{} \&  $16$ \FDFilters{} for the stable, forcing and noisy cases, and  $10$ \FDBlocks{} \& $16$ \FDFilters{} for the unstable case. Furthermore, we investigate the sensitivity of \FDNet{} to the hyper-parameters, i.e., numbers of \FDBlocks{} and \FDFilters{}. More numerical results with different numbers of \FDBlocks{} and \FDFilters{} and different noise levels can be found in Appendix \ref{app_num_res}.

\begin{figure}
  \centering
  \includegraphics[width=0.98\textwidth]{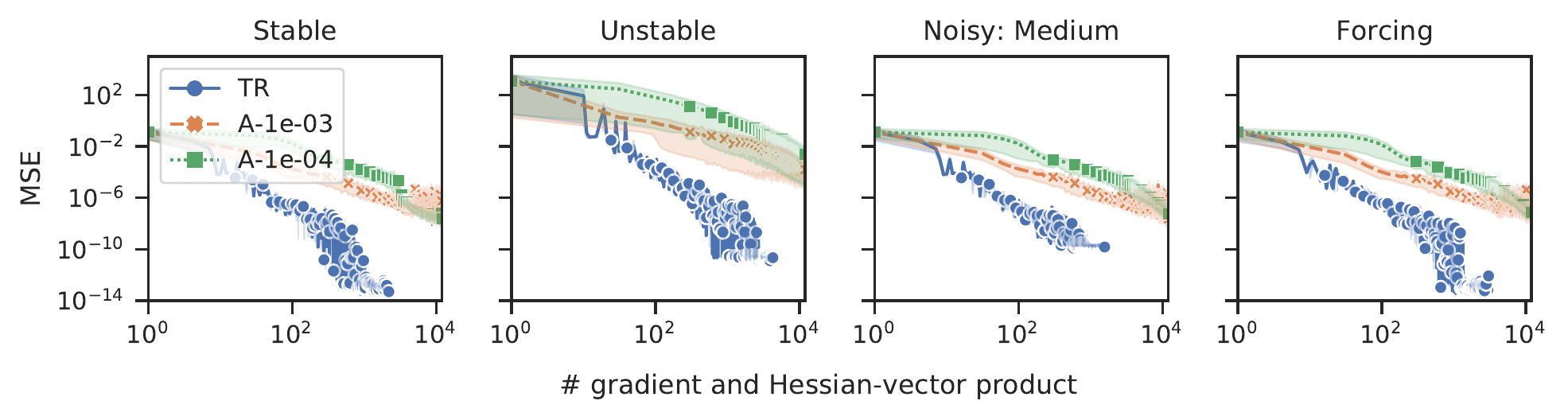}
  \flushleft{\scriptsize \textit{Abbreviations}: TR, Trust-Region Newton CG Method; A-1e-03/A1e-04, ADAM with a learning rate of $10^{-3}/10^{-4}$.}
  \caption{Evolution of training error. The marked dashes represent the average mini-batch MSE loss over $10$ random seeds and the filled areas represent their $95\%$ confidence intervals.}
  \label{fig.main1}
\end{figure}

We begin our presentation by showing the evolution of the training errors, i.e., $\text{MSE}_{\text{mini}}$ \eqref{eq8}, for different optimization algorithms in Figure \ref{fig.main1}.
We compare the performance of the algorithms in terms of the number of gradient and Hessian-vector computations. As is clear from the figure, the TR methods is able to achieve smaller $\text{MSE}_{\text{mini}}$ than ADAM within the given budget for all cases. This is true for other \FDBlock{} and \FDFilter{} configurations, as well as different noise levels; see Appendix \ref{app_num_res} for more results.

\begin{figure}
  \centering
   \includegraphics[width=\linewidth]{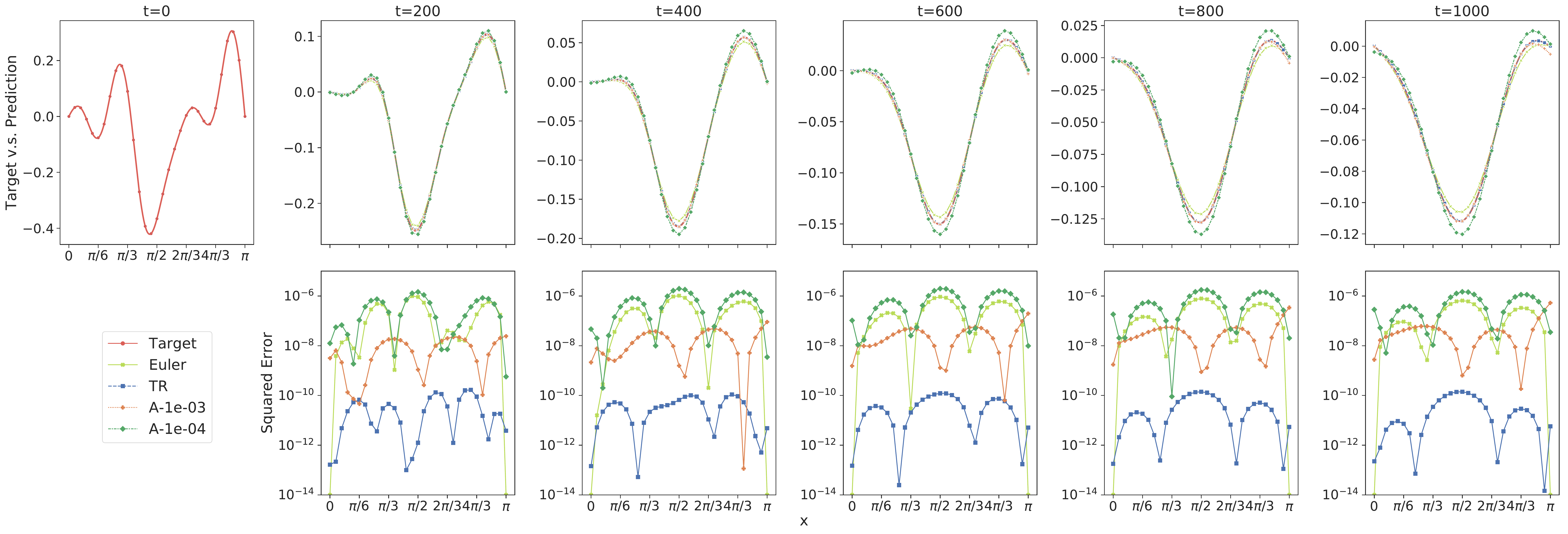}
  \flushleft{\scriptsize \textit{Abbreviation}: Euler, forward Euler method.}
  \caption{Sequence of predictions for the stable case at $t \in \{200,400,600,800,1000\}$ and $x \in \{0,\Delta x, 2\Delta x, ..., \pi\}$ for one specific IC. Top Left corner: $u_s(x,0)$; Rest of top row: Targets ($u_s(x,t)$) and predictions ($\tilde{u}_s(x,t)$); Bottom row: Squared errors ($\left(u_s(x,t) - \tilde{u}_s(x,t)\right)^2$). } 
  
  \label{fig.main2}
\vspace{0.1in}
    \centering
    \begin{minipage}{1.0\textwidth}
    \centering
        \includegraphics[width=0.23\linewidth]{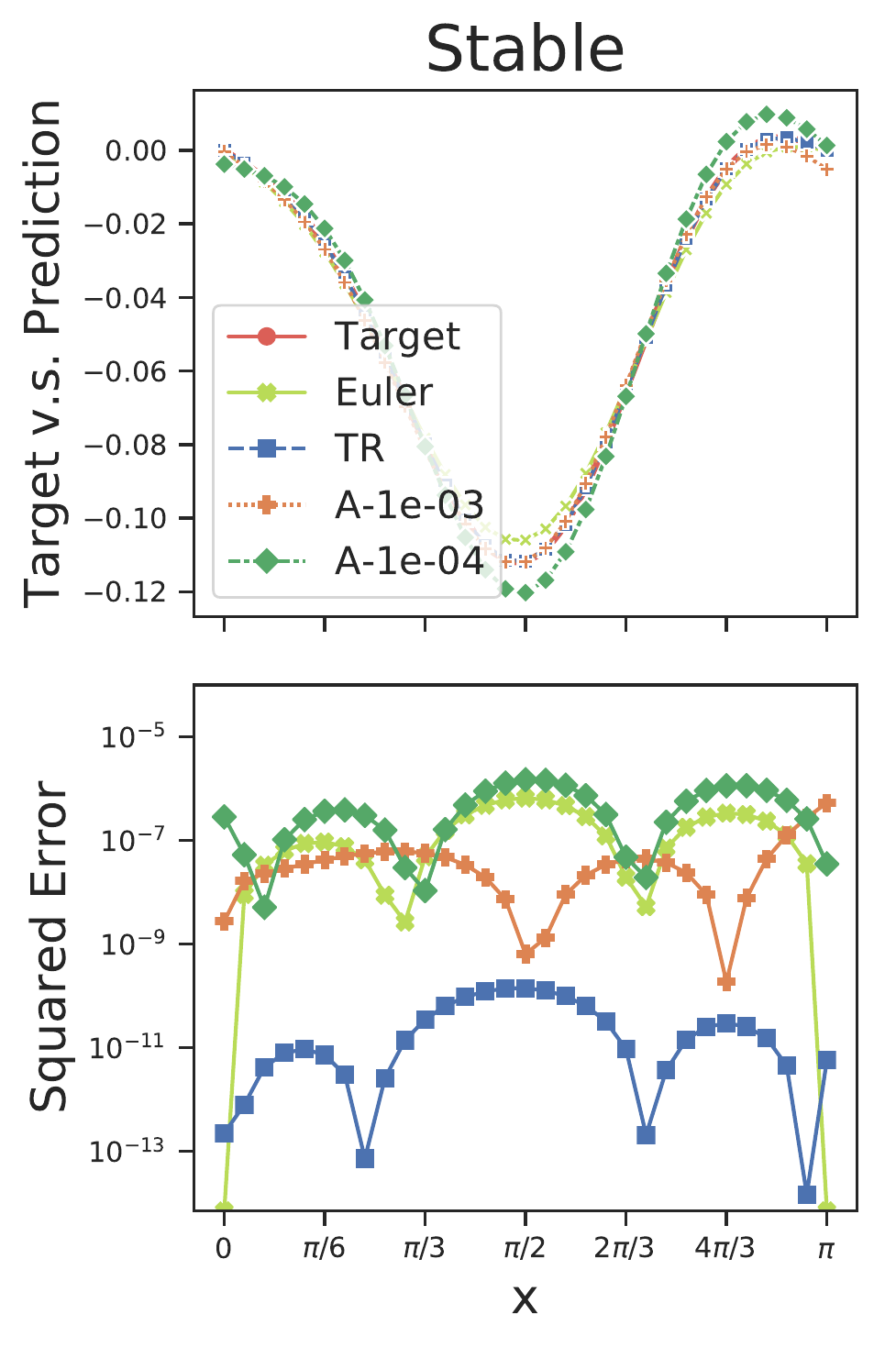}
        \includegraphics[width=0.23\linewidth]{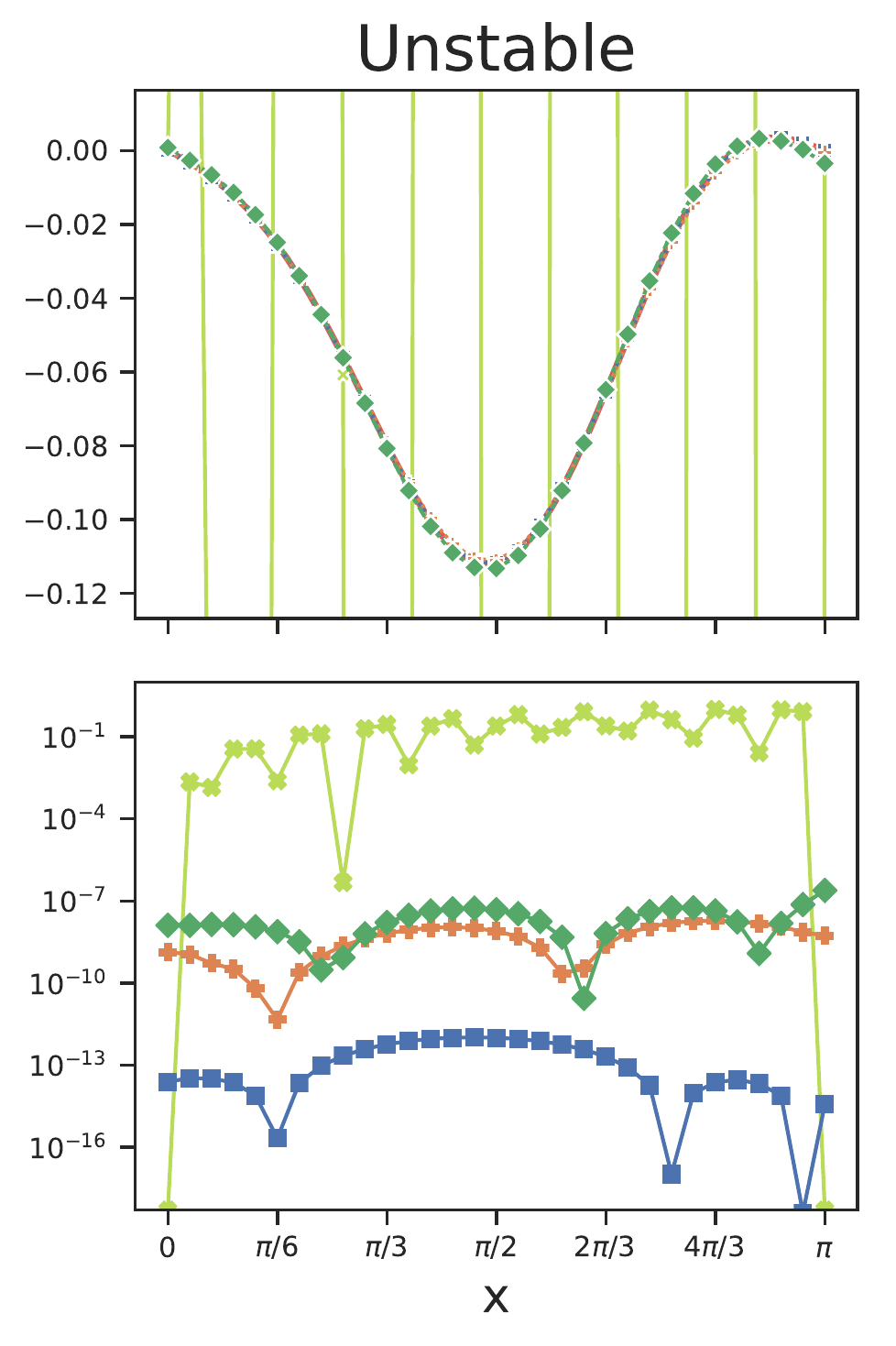}
        \includegraphics[width=0.23\linewidth]{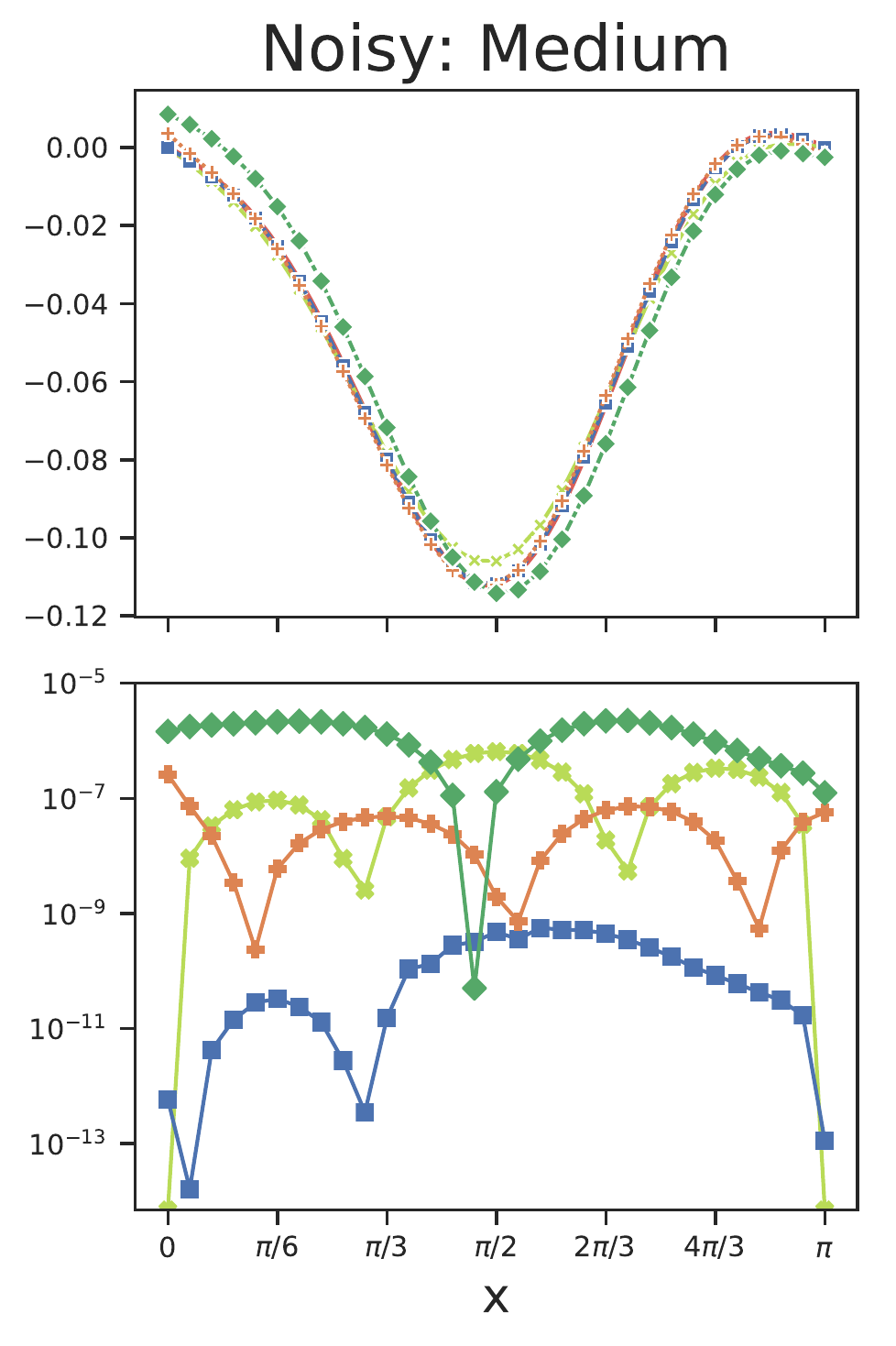}
        \includegraphics[width=0.23\linewidth]{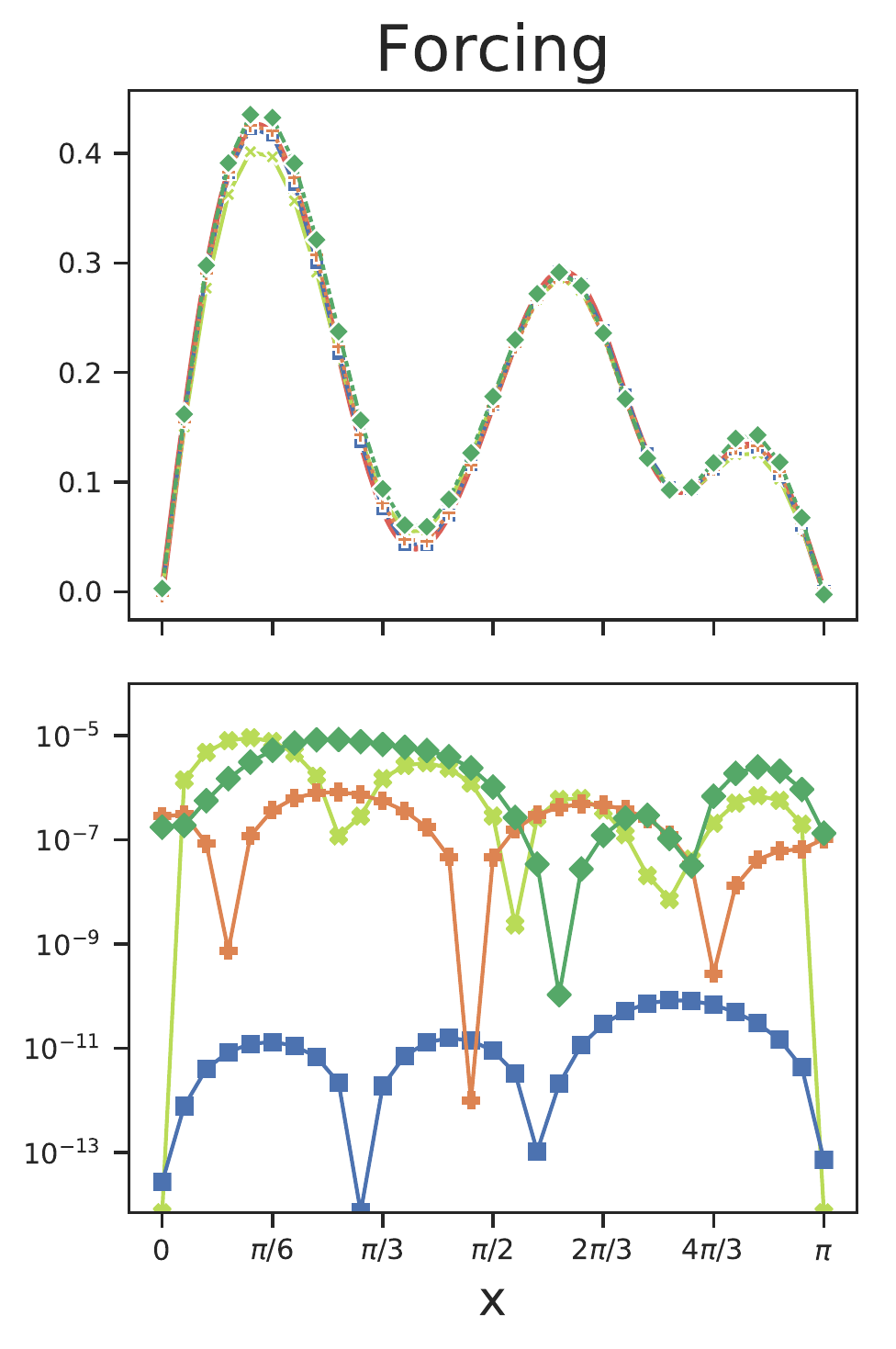}
    \end{minipage}
  \caption{Final predictions for all cases at $t=1000$ and $x \in \{0,\Delta x, 2\Delta x, ..., \pi\}$ for one specific IC. Top row: Targets and predictions; Bottom row: Squared errors.}
  \label{fig.main3}
  \centering
  \includegraphics[width=0.98\textwidth]{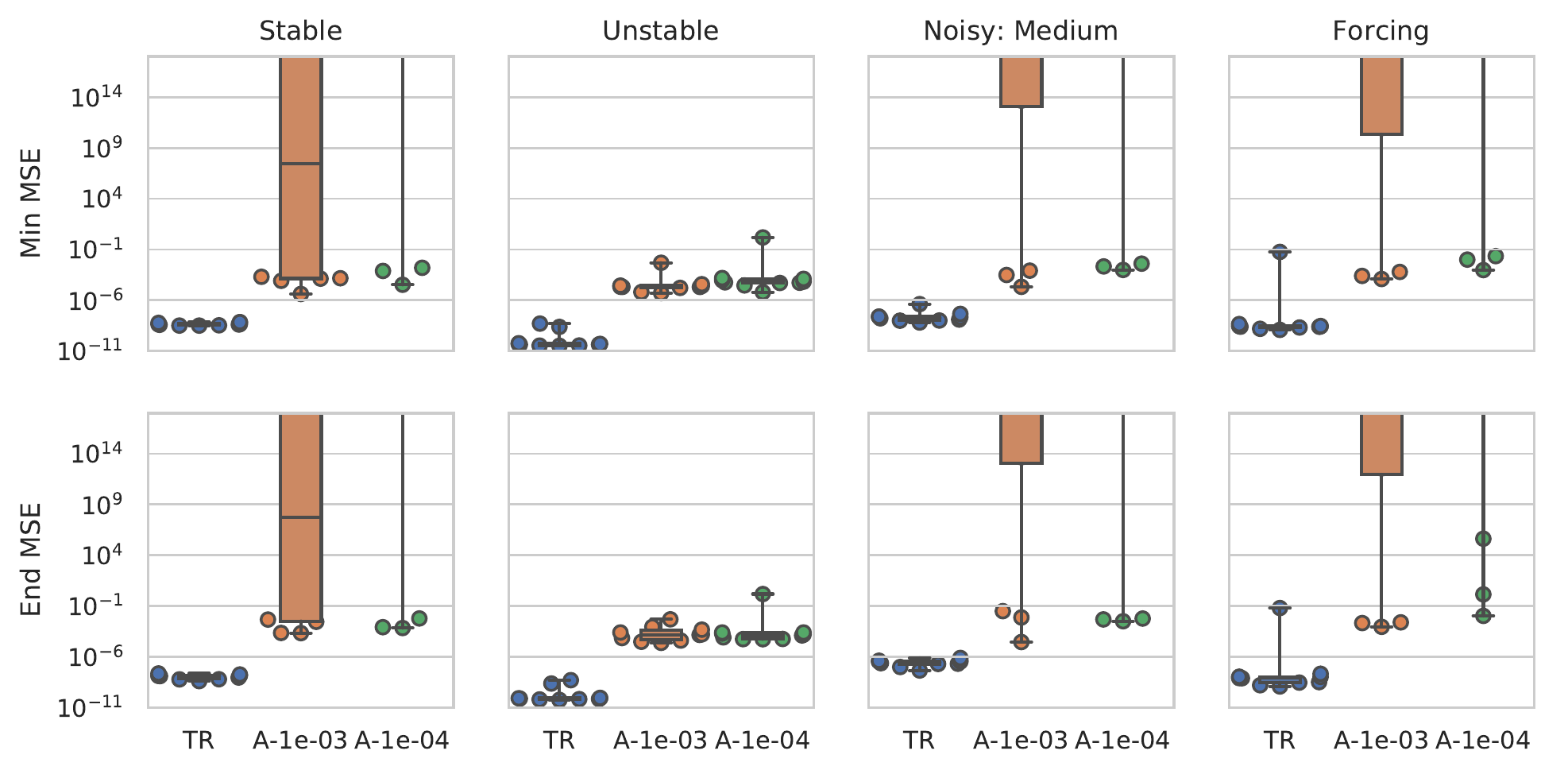}
  \caption{
  Testing error for different cases, algorithms and random seeds. Top row: Minimum testing error; Bottom row: Final testing error. Top/middle/bottom line of boxplot is the upper quartile/median/lower quartile, respectively, the whiskers represent the range, and the circled dots are individual observations.}
  \label{fig.main5}
\end{figure}

Having demonstrated that our networks can be adequately trained within a budget, we proceed to show the testing (prediction) accuracy of \FDNet{} and compare against a standard benchmark numerical scheme, the forward Euler method, in Figures \ref{fig.main2} and \ref{fig.main3}.
Figure \ref{fig.main2} shows the sequence of predictions made by \FDNet{}, for the stable case, at 5 time steps in the horizon. For brevity, in Figure \ref{fig.main3}, we only show the final predictions at $t=1000$ for the remaining cases and defer the rest of the results to Appendix \ref{app_num_res}. We chose the sequences of 1000-step predictions with the minimum testing errors \eqref{eq9} over the course of training, and compared them with the predictions made by the forward Euler method. We show results for a single IC. Figures \ref{fig.main2} and \ref{fig.main3} clearly indicate that: $(1)$ \FDNet{}, when trained sufficiently well, is able to make higher quality predictions across the time horizon than the forward Euler method; $(2)$ training our networks with the TR method allows for better predictions than the ADAM optimizer; $(3)$ the performance of the forward Euler method is highly dependent on the case. Specifically, the Euler method, as predicted by the theory, cannot adequately capture the dynamics of the PDE in the unstable setting.

To further illustrate the testing (prediction) performance of \FDNet{}, in Figure \ref{fig.main5}, we show the minimum testing errors over the training process and the final testing errors for every case, algorithm and random seed. Clearly, training \FDNet{} using the TR method results in higher accuracy predictions with lower variance for all cases. Indeed, this is true for all configurations of \FDNet{}; see Appendix \ref{app_num_res}. It is worth noting that 1000-step prediction is a challenging task. This can be attributed to the fact that the error at each time step propagates throughout the time horizon, and any imperfect intermediate predictions can severely deteriorate the final prediction at $t=1000$. This is evident for the networks trained by ADAM for the stable, noisy and forcing cases, where the testing errors for certain random seeds are very large. This effect is less severe for the unstable case as there are far fewer time steps from $t=0$ to $t=1000$. We should note, however, that the TR method is able to reduce the testing errors for all cases, and the effect of error propagation is not evident. This is true across the different cases so long as the networks are appropriately configured; see Appendix \ref{app_num_res} for more details. 

\begin{wrapfigure}{r}{0.7\textwidth}
\vskip-15pt
  \centering
  \includegraphics[width=0.7\textwidth]{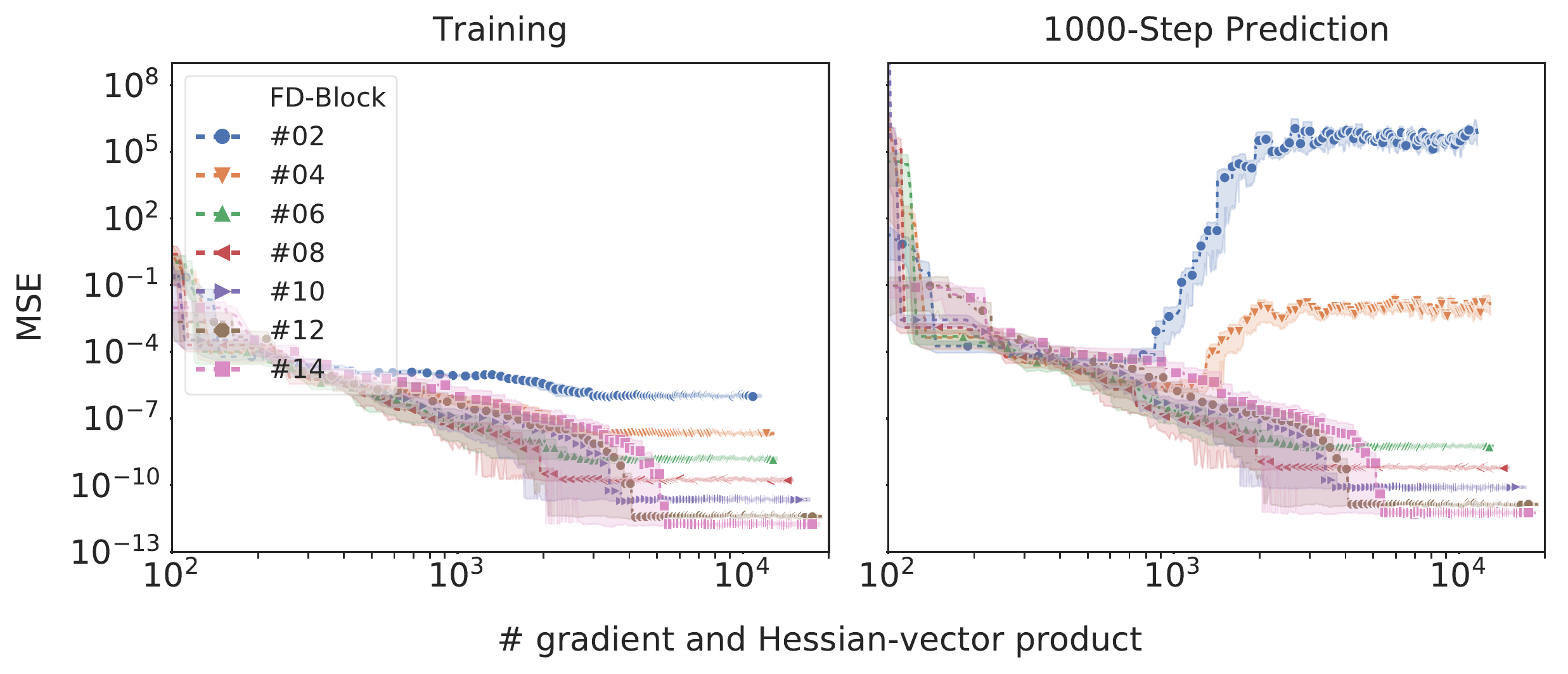}
  \caption{Evolution of the average training \& testing errors (and $95\%$ confidence intervals) for different numbers of \FDBlocks{} \&  $16$ \FDFilters{} for the unstable case.}
  \label{fig.main6}
  \vskip-10pt
\end{wrapfigure}
Next, we investigate the sensitivity of \FDNet{} to choices of the hyper-parameters (the numbers of \FDBlocks{} and \FDFilters{}). The main results are given in Figures \ref{fig.main6} and \ref{fig.main7}; see Appendix \ref{app_num_res} for more results. To fully reveal the learning capabilities of \FDNet{}, we used the TR method without imposing any budget, and trained the network of each configuration with $10$ random seeds.

As mentioned in Section \ref{Method-FD-net}, stacking $k$ \FDBlocks{} introduces $k-1$ artificial time steps to the time interval $[t,t+\Delta t]$. While $k=1$ suffices for the cases with small $\Delta t$ (i.e., $\Delta t \leq 1$), it is crucial to introduce a sufficient number of artificial time steps in order to achieve good training and testing performance in the setting where $\Delta t$ is large. Figure \ref{fig.main6} shows the results of training the networks with different numbers of \FDBlocks{} and $16$ \FDFilters{} for the unstable case ($\Delta t = 200$). As is clear, the larger the number of \FDBlocks{}, the lower the eventual training and testing errors at the cost of training a more complex network.

\begin{wrapfigure}{r}{0.7\textwidth}
\vskip-20pt
  \centering
    \includegraphics[width=0.7\textwidth]{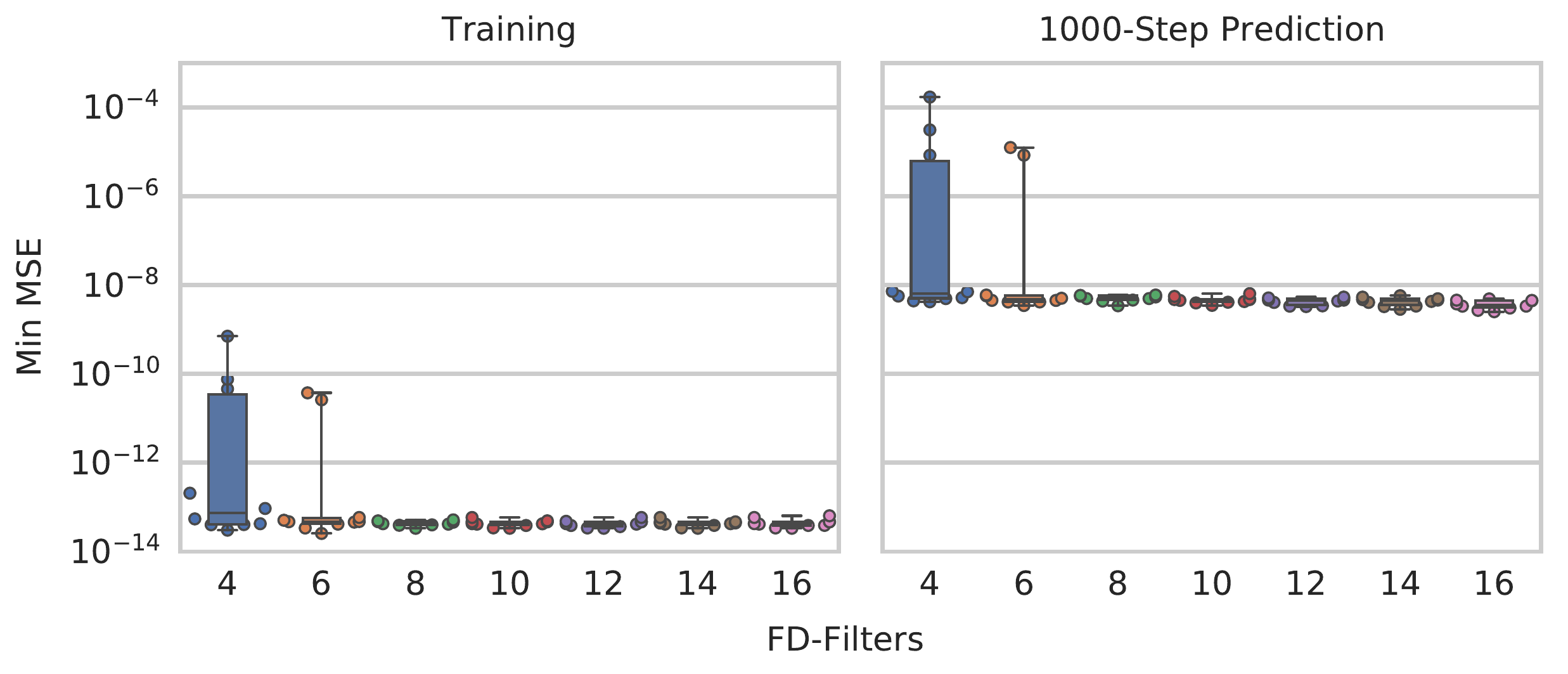}
    \vskip-10pt
  \caption{
  Minimum training \& testing errors (with the lower/upper quartiles, medians and ranges) for different numbers of \FDFilters{} \&  $1$ \FDBlock{} for the stable case.}
  \vskip-10pt
  \label{fig.main7}
\end{wrapfigure}
The size of \FDNet{} models depend on the number of \FDFilters{}. Thus far, we illustrated the performance of \FDNet{} with $16$ \FDFilters{}. Figure \ref{fig.main7} shows the results of training the networks with different numbers of \FDFilters{} and $1$ \FDBlock{} on the stable case. The figure clearly shows that the performance of \FDNet{} with a small number of \FDFilters{} varies by random seeds, and that utilizing a larger number of \FDFilters{} reduces this variance. That being said, the results highlight that there is little benefit to using more that $8$ \FDFilters{}, as the average testing error only improves marginally with more \FDFilters{}. Similar conclusions can be drawn for the other cases; see Appendix \ref{app_num_res}.

\section{Final Remarks}\label{fin_rem}

In this paper, we presented a novel neural network framework, \FDNet{}, for learning the dynamics of PDEs and making predictions solely based on trajectory data. The architecture of \FDNet{} is inspired by finite differences and residual neural networks. \FDNet{} is able to efficiently learn the dynamics and make predictions for the heat equation in the stable, unstable, forcing and noisy settings. However, this was only a proof-of-concept study of our \FDNet{} model. As future work, we aim to study the applicability of \FDNet{} for solving different PDEs (e.g., higher-order, nonlinear, etc) and compare against higher-order and implicit numerical schemes, to extend the territory to discovering hidden PDEs, and to develop customized optimization algorithms for training the networks.

\subsubsection*{Acknowledgements} This work was partially supported by the U.S. National Science Foundation, under award numbers NSF:CCF:1618717, NSF:CMMI:1663256 and NSF:CCF:1740796.

\bibliographystyle{plain}
\bibliography{mybib}

\newpage

\appendix

\section{Exact solutions to PDEs}\label{app_exact}

In this section, we derive the exact solutions to the PDEs we investigate. Although these results are well-known, we state them here for completeness.

\subsection{No Forcing}
Given the heat equation \eqref{eq2} defined on a 1-D bar of length $L$ with a rate of diffusion $\beta$, the boundary conditions \eqref{eq3} and the initial condition \eqref{eq4}, we restate the exact solution \eqref{eq5} in the following
\begin{align*}
u(x,t) &= \sum_{i=1}^N C_i \sin \left(\frac{i \pi x}{L}\right) e^ {- \beta  (i \pi/L)^2 t}.
\end{align*}
\subsection{With Forcing}
Given the forcing function and the corresponding equation \eqref{eq7}, the exact solution is 
\begin{align}
    u(x,t) = \sum_{i=1}^N \left(C_i - \frac{D_i}{\beta(i\pi/L)^2}\right) \sin\left(\frac{i \pi x}{L}\right) e^{-\beta (i \pi / L)^2 t} + \sum_{i=1}^N \frac{D_i}{\beta(i\pi /L)^2} \sin\left(\frac{i \pi x}{L}\right). \label{eq11}
\end{align}

\section{Extended Numerical Results}\label{app_num_res}

\subsection{Detailed Description of Experiments}

In this section, we describe the approach and implementation of the experiments in detail.
\subsubsection{Data Generation} \label{app_data}
Following the descriptions in Section \ref{Experiment-1}, the $200$ ICs were generated randomly with $200$ distinct random seeds. We generated the data set of the solution \eqref{eq5} for the stable case and used it as our base data for the noisy and unstable cases. Specifically, the noisy data set was formed by adding the multiplicative noises to the base data; the unstable data set was formed by extracting the data of $t \in \{0,200,400,600,800,1000\}$ from the base data. For the forcing case, we used a randomly generated forcing function to create the data set of solution (\ref{eq11}). 

\subsubsection{Supplementary Testing Procedures} \label{app_sup_test}
In addition to the 1000-step prediction studied in Section \ref{Experiment}, we adopted two supplementary testing procedures, i.e., the one-step and multi-step predictions, to evaluate the networks' performance of making short-term predictions. 

Given a $\Delta t$, let $\tau'\in \mathbb{N}^+$ s.t. $\tau'\Delta t \leq T$ and consider a generalized testing data set 
\begin{align*}
    A_{\text{test}_{\tau'}} = \left\{ \left(u_s(x,t),u_s(x,t+\tau'\Delta t)\right)\;|\;s \in S_{\text{test}},\;x \in \{0,\Delta x,..., L\},\; t\in \{0,\Delta t,...,T-\tau'\Delta t\}\right\},
\end{align*}
where $S_{\text{test}}$ is the index set of ICs for the testing purposes, $\left(u_s(x,t),u_s(x,t+\tau'\Delta t)\right)$ is a testing sample, $u_s(x,t)$ is the input to the network,  and $u_s(x,t+\tau'\Delta t)$ is the target. 

The one-step prediction procedure is consistent with the training procedure, and we used $\tau'=1$ for all cases. To define the multi-step prediction, we let $\tau'=10$ for the stable, noisy and forcing cases and $\tau' = 3$ for the unstable case. (Note that when $\tau'=5$ for the unstable case or $\tau'=1000$ for one of the others, we have the data set for the 1000-step prediction.)

Accordingly, we defined the testing error as MSE in a generalized form
\begin{align}
    \text{MSE}_{\text{test}_{\tau'}} = \frac{1}{|A_{\tau'}|} \sum_{s,x,t} \left(u_s(x,t+\tau'\Delta t) - \tilde{u}_s(x,t+\tau'\Delta t)\right)^2, \label{eq12}
\end{align}
where $\tilde{u}_s(x,t+\tau'\Delta t)$ denotes a prediction made by the network. 

\subsubsection{Implementation and Design of Experiments}
We implemented the networks in Python with PyTorch and trained the networks on an \textit{NVIDIA K80} GPU. For the optimization methods, we implemented the Trust-Region (TR) Newton CG method in Python and used the ADAM optimizer of PyTorch. 

The primary goal of the experiments conducted in this paper is to evaluate the training and testing performance of the networks. To this end, we chose different configurations (see Table \ref{tab:2}) and adopted three testing procedures for each case. And, for each configuration, we trained the network with the TR methods and ADAM with learning rates $10^{-3}$ and $10^{-4}$, and used $10$ distinct random seeds to initialize the network parameters and to select the stochastic mini-batches. We constrained the training budget of TR to be $100$ iterations for the stable, noisy and forcing cases and $300$ iterations for the unstable case. And, we allowed the ADAM algorithms to run for $12000$ iterations regardless of a case. 
\begin{table}[h!]
\centering 
\caption{Network Configurations.}
\vfill
\label{tab:2}
\begin{threeparttable}
\begin{tabular}{l c c }
\toprule
Hyper-Parameter & Unstable Case & Others\tnote{$\dagger$}\\
 \midrule
 \# \FDBlocks{}  & $1,2,3,4,6,8,10$ & $1,2,3,4$\\
 \hdashline
 \# \FDFilters{} & $2,4,8,16$ & $2,4,8,16$\\
 \bottomrule
\end{tabular}
\begin{tablenotes}\scriptsize
\item[$\dagger$] includes the stable, noisy and forcing cases.
\end{tablenotes}
\end{threeparttable}
\end{table} 

To study the effects of the hyper-parameters, i.e., the number of \FDBlocks{} and \FDFilters{}, and to validate our design of the architecture, we conducted further experiments with different numbers of \FDBlocks{} against the unstable case and with different numbers of \FDFilters{} against the stable case. 

In the following sections, we will present the full experimental results in the order of the stable case, the unstable case, the forcing case and the noisy case.

\clearpage

\subsection{Stable Case}

\begin{figure}[ht!]
  \centering
  \includegraphics[width=0.7\textwidth]{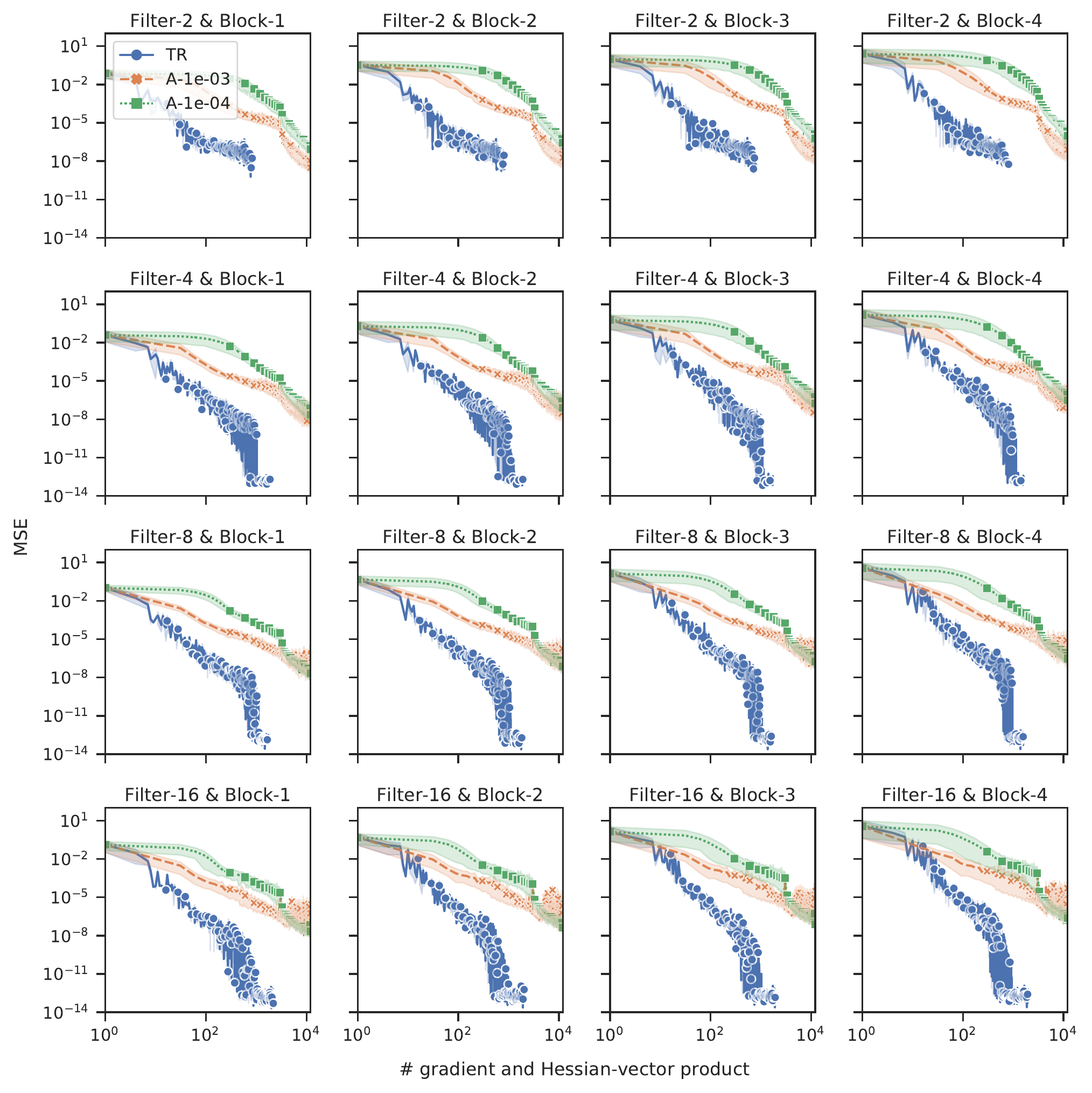}
  \caption{\textbf{Stable Case:} Evolution of the MSE loss of stochastic mini-batch by configuration.}
  \label{fig.stable_train}
  \vspace{0.5in}
  \centering
  \includegraphics[width=0.9\textwidth]{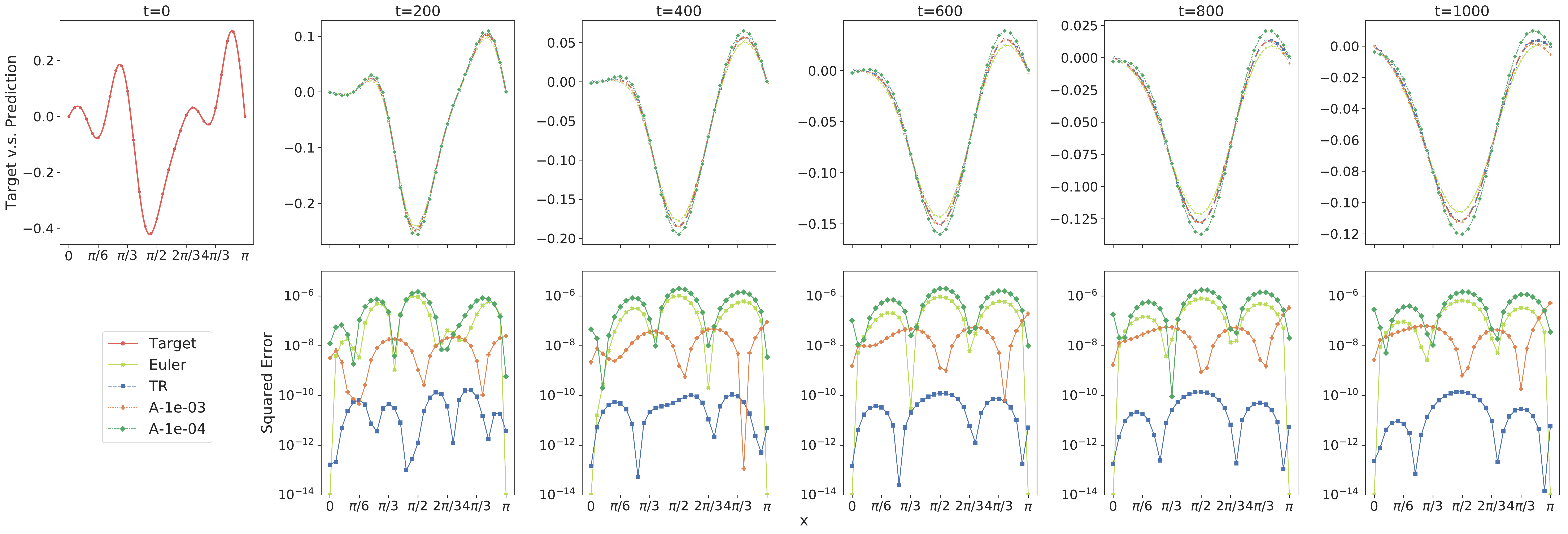}
  \caption{\textbf{Stable Case:} Sequence of predictions.}
  \label{fig.stable_u}
\end{figure}

\begin{figure}
  \centering
  \includegraphics[width=0.7\textwidth]{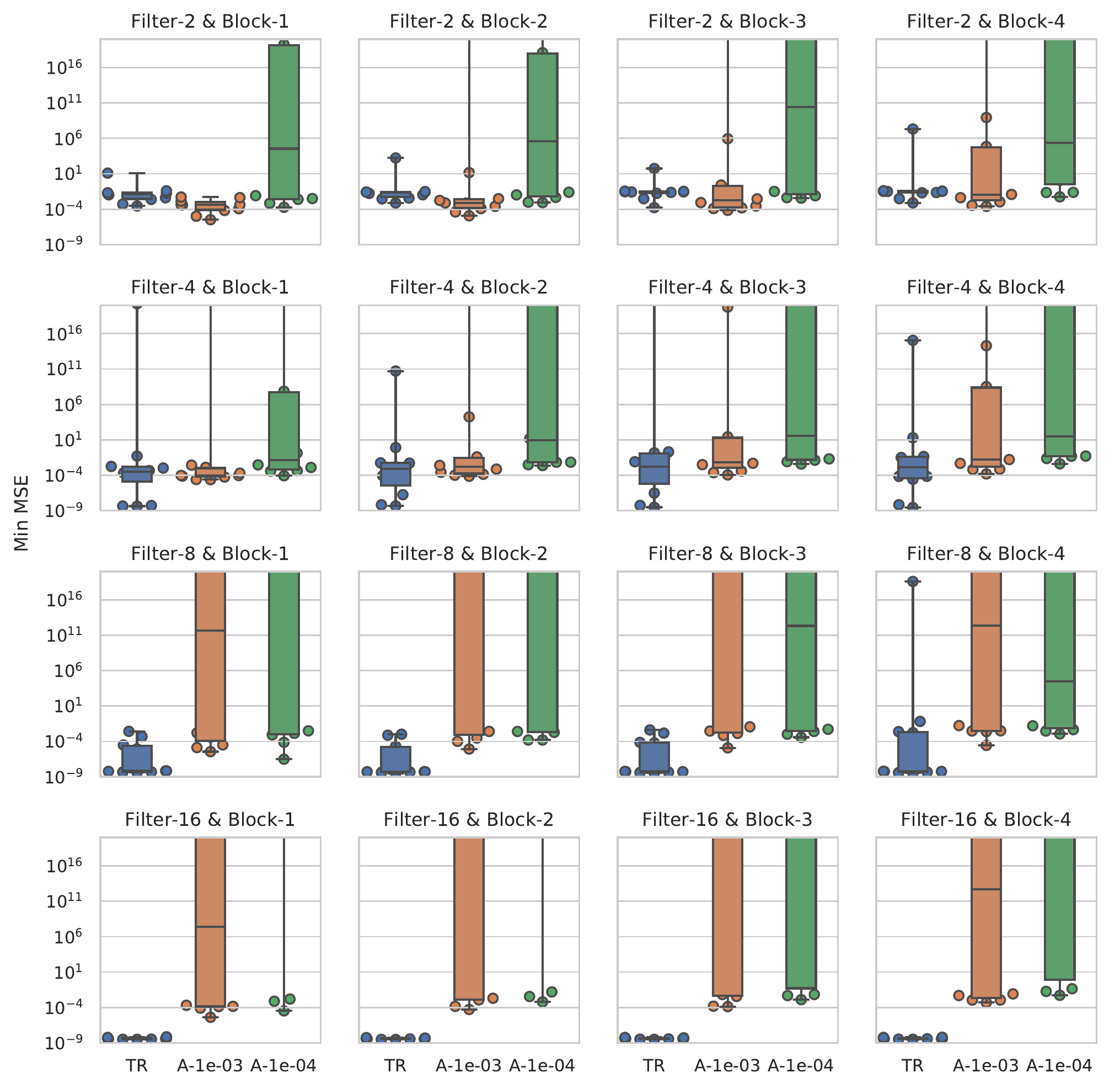}
  \caption{\textbf{Stable Case:} Minimum testing errors by configuration - 1000-step prediction.}
  \label{fig.stable_1000}
  \vspace{0.7in}
  \centering
  \includegraphics[width=0.7\textwidth]{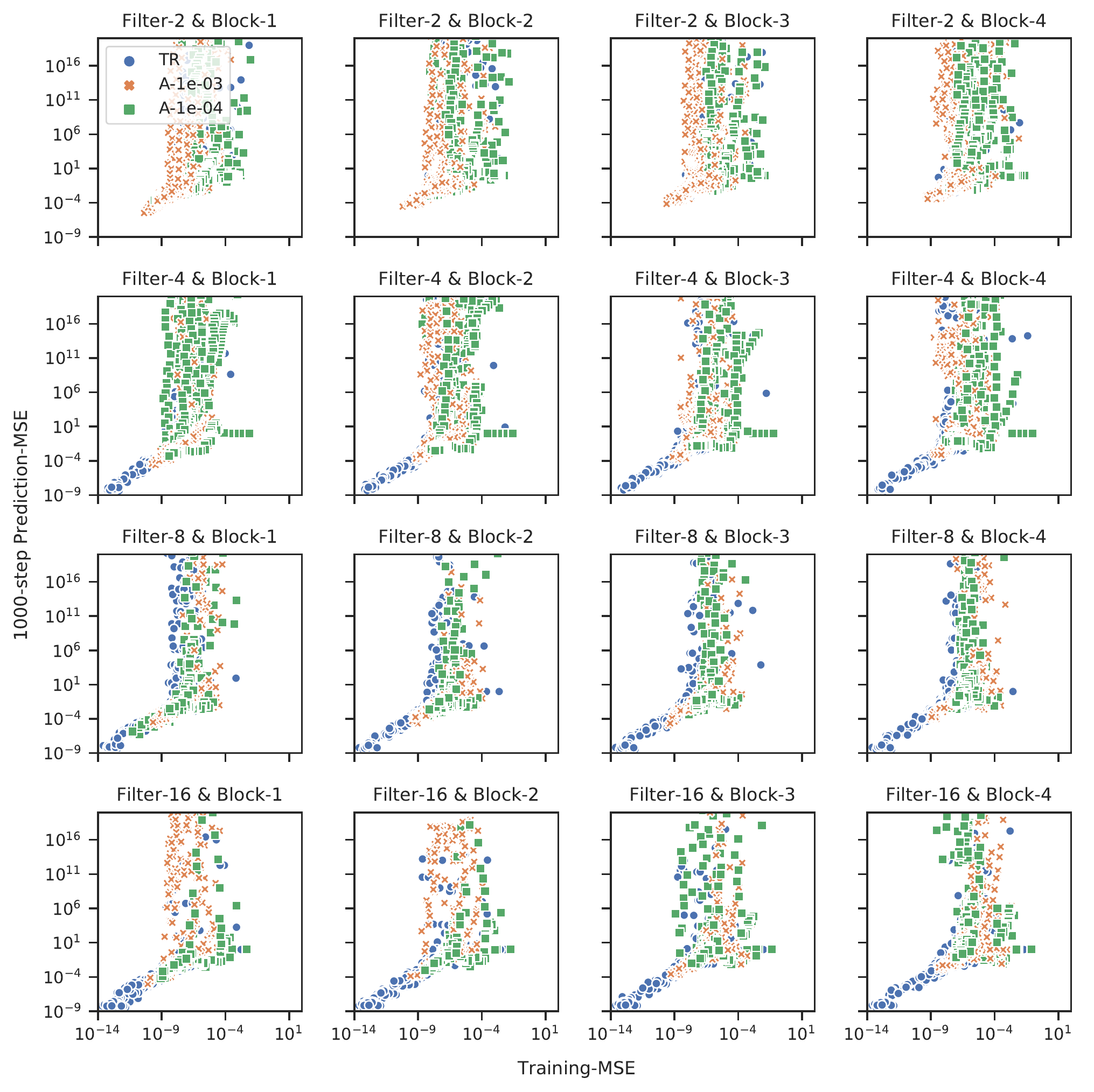}
  \caption{\textbf{Stable Case:} Training vs. 1000-step prediction errors (over the training process) by configuration.}
    \vspace*{3in}
  \label{fig.stable_scatter}
\end{figure}

\begin{figure}
  \centering
  \includegraphics[width=0.7\textwidth]{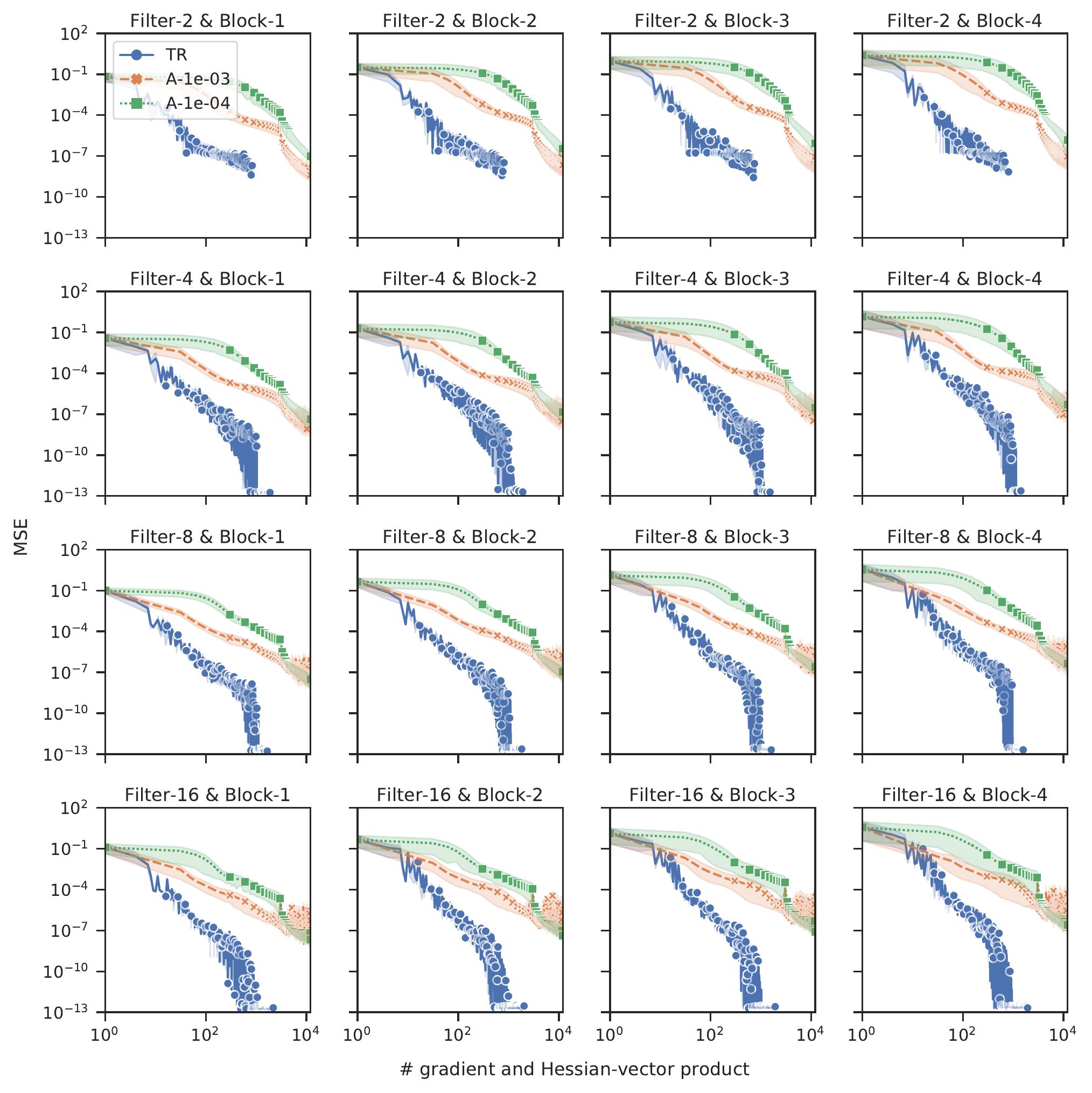}
  \caption{\textbf{Stable Case:} Evolution of the testing errors by configuration - one-step prediction.}
  \label{fig.stable_1_evol}
  \vspace{0.7in}
  \centering
  \includegraphics[width=0.7\textwidth]{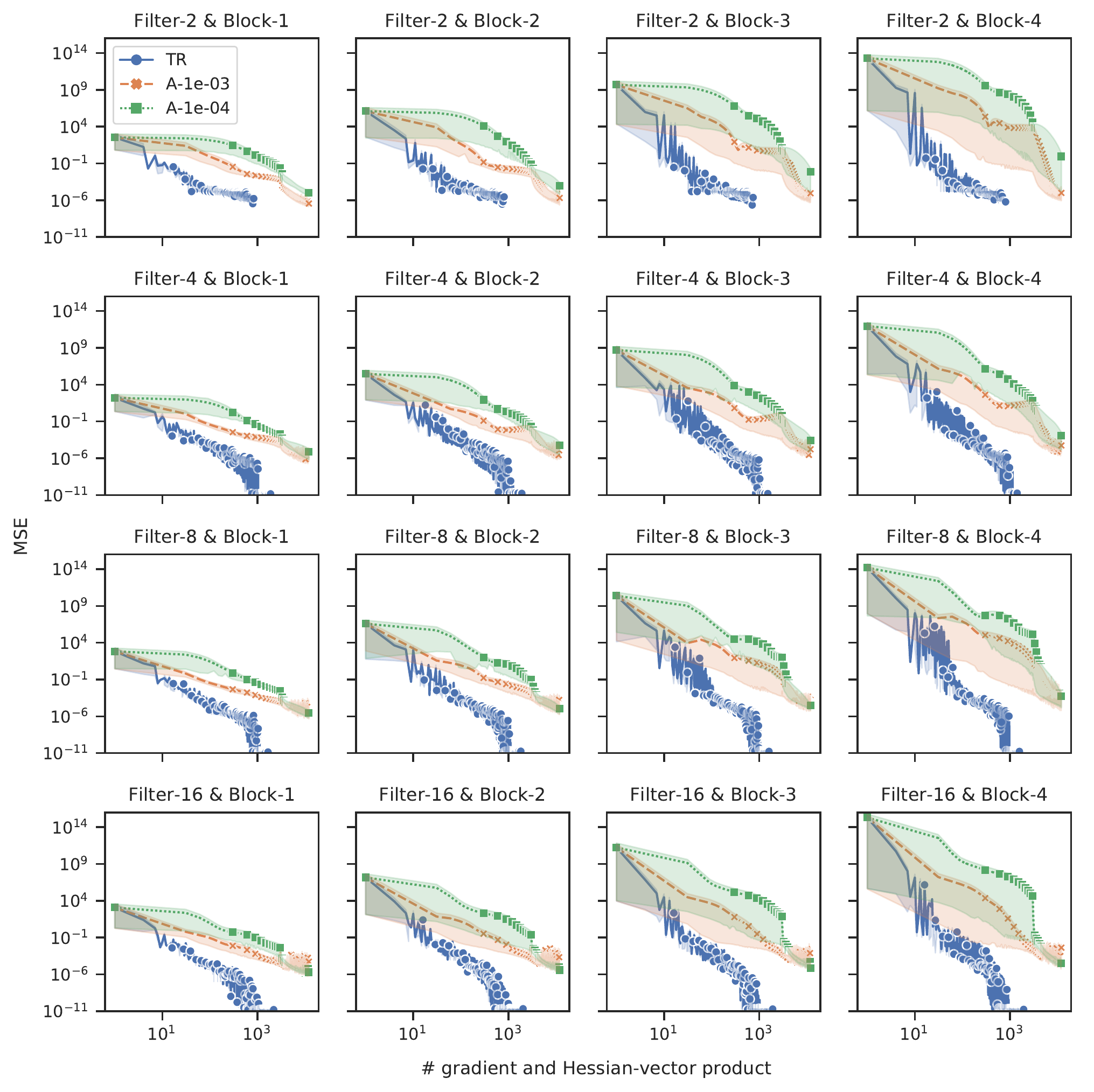}
  \caption{\textbf{Stable Case:} Evolution of the testing errors by configuration - multi-step prediction.}
    \vspace*{3in}
  \label{fig.stable_10_evol}
\end{figure}

\begin{figure}
  \centering
  \includegraphics[width=0.7\textwidth]{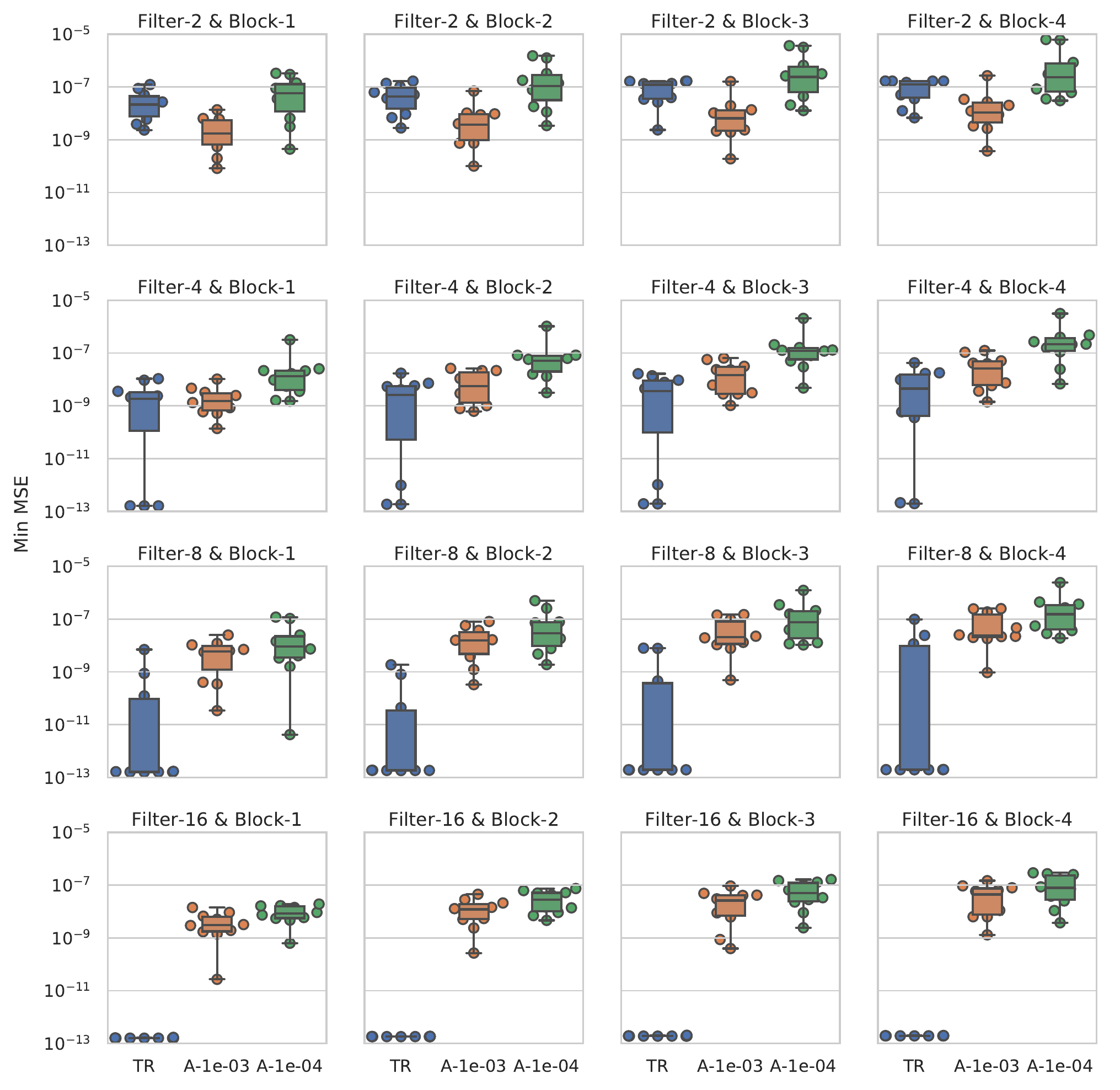}
  \caption{\textbf{Stable Case:} Minimum testing errors by configuration - one-step prediction.}
  \label{fig.stable_1}
  \vspace{0.7in}
  \centering
  \includegraphics[width=0.7\textwidth]{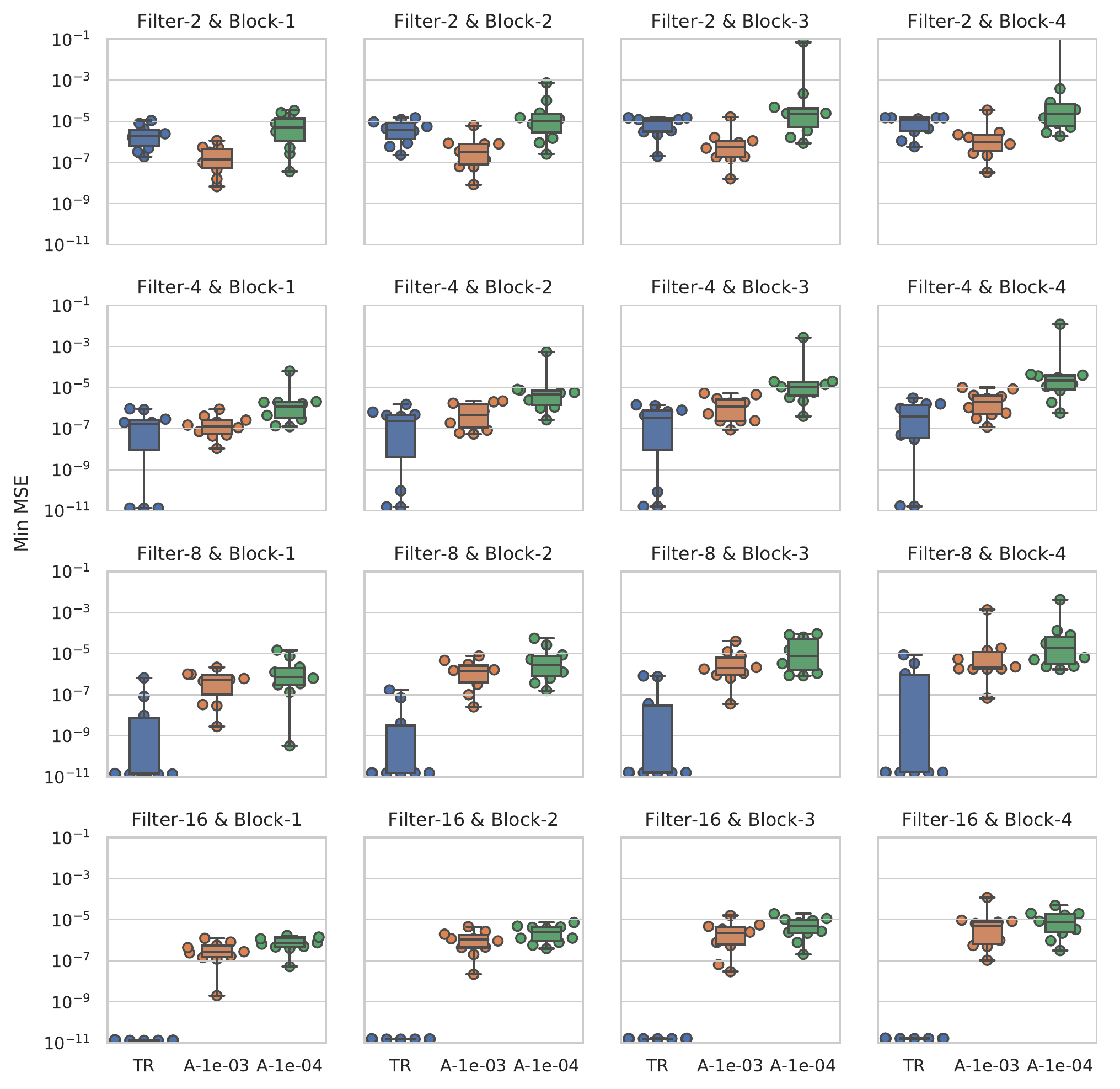}
  \caption{\textbf{Stable Case:} Minimum testing errors by configuration - multi-step prediction.}
    \vspace*{3in}
  \label{fig.stable_10}
\end{figure}

\begin{figure}[t!]
  \centering
  \includegraphics[width=0.3\textwidth]{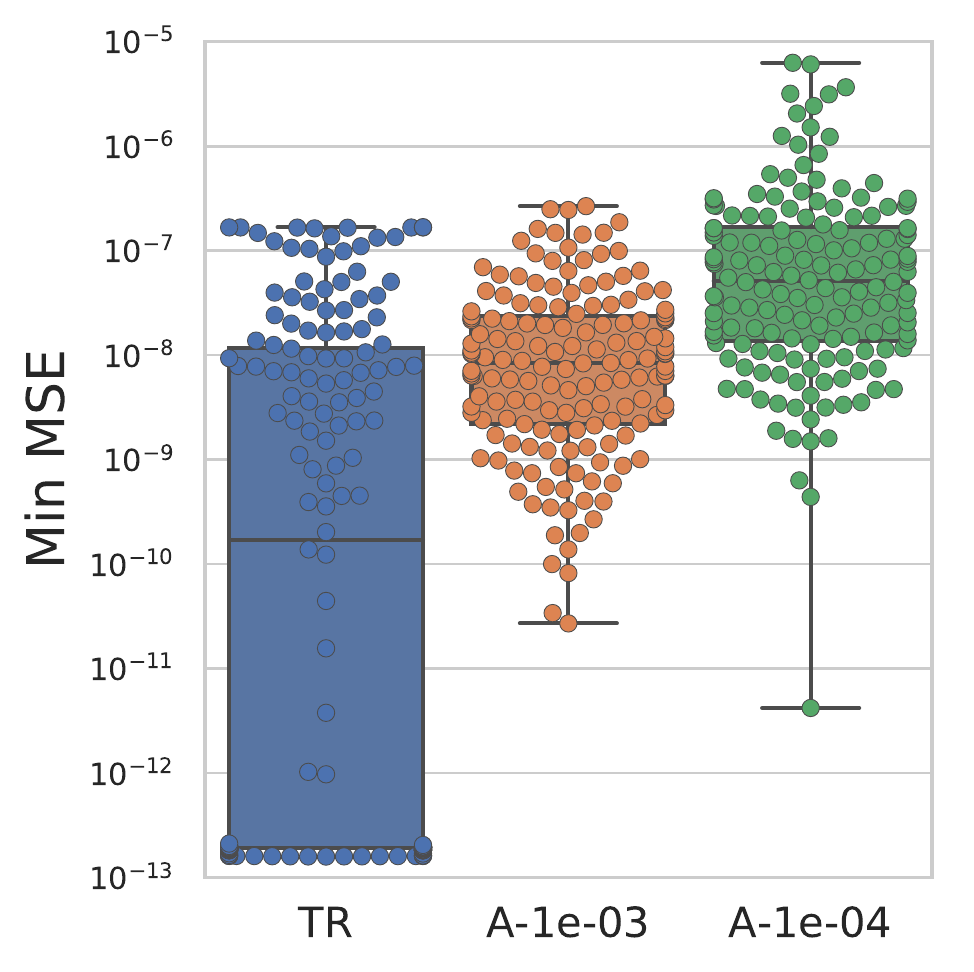}
  \includegraphics[width=0.3\textwidth]{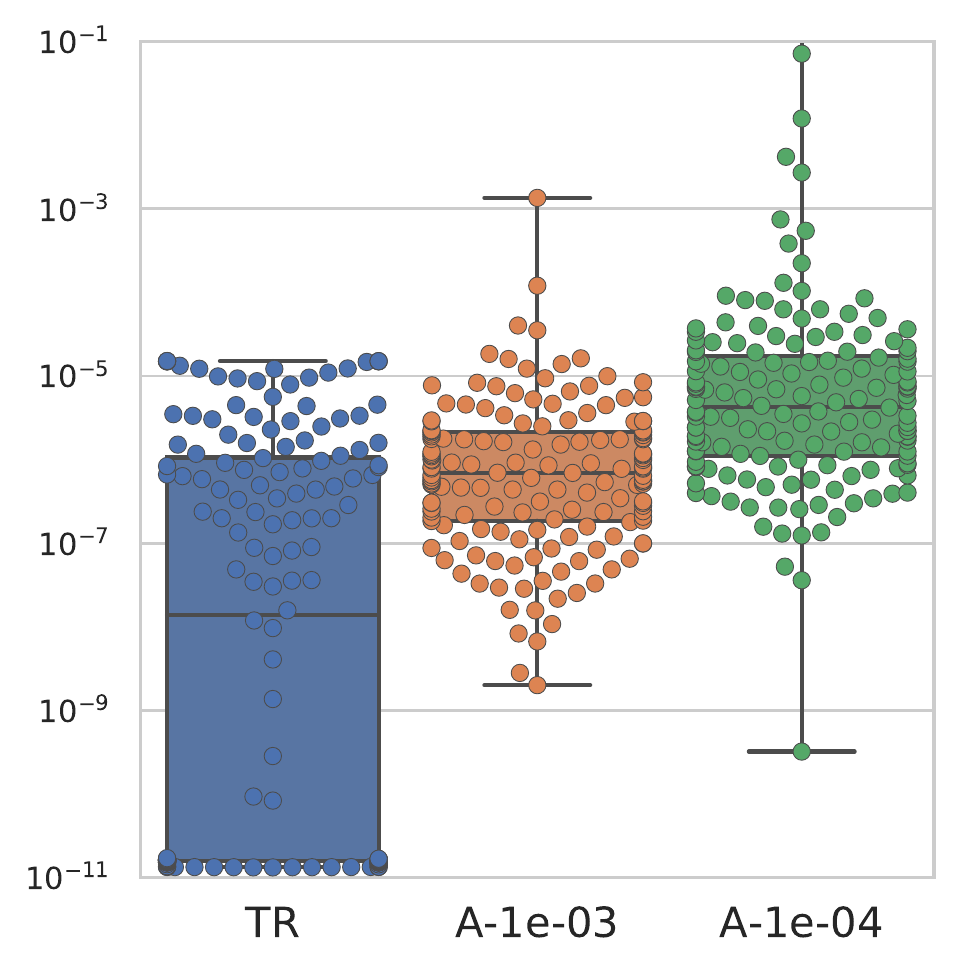}
  \includegraphics[width=0.3\textwidth]{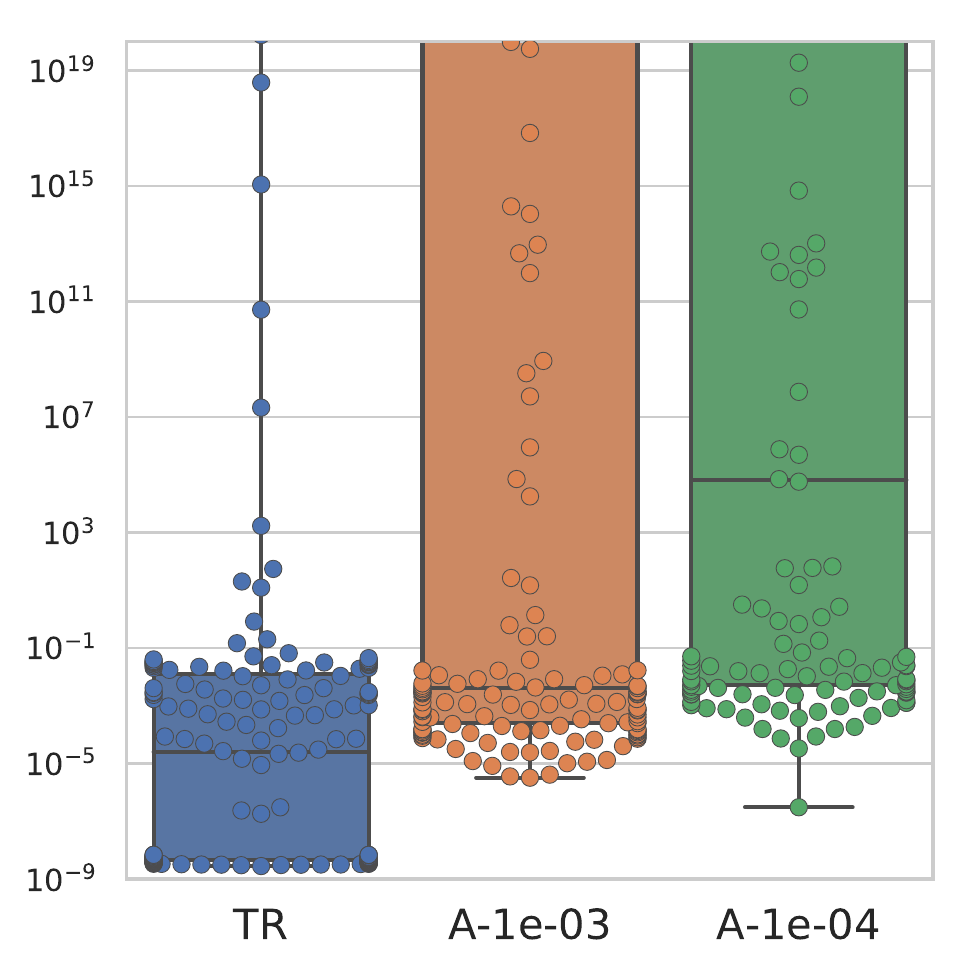}
  \caption{\textbf{Stable Case:} The minimum testing errors of the one-step (left), multi-step (middle) and 1000-step (right) predictions (aggregated over all configurations).}
  \label{fig.stable_all}
  \vspace{0.5in}
  \centering
  \includegraphics[width=0.9\textwidth]{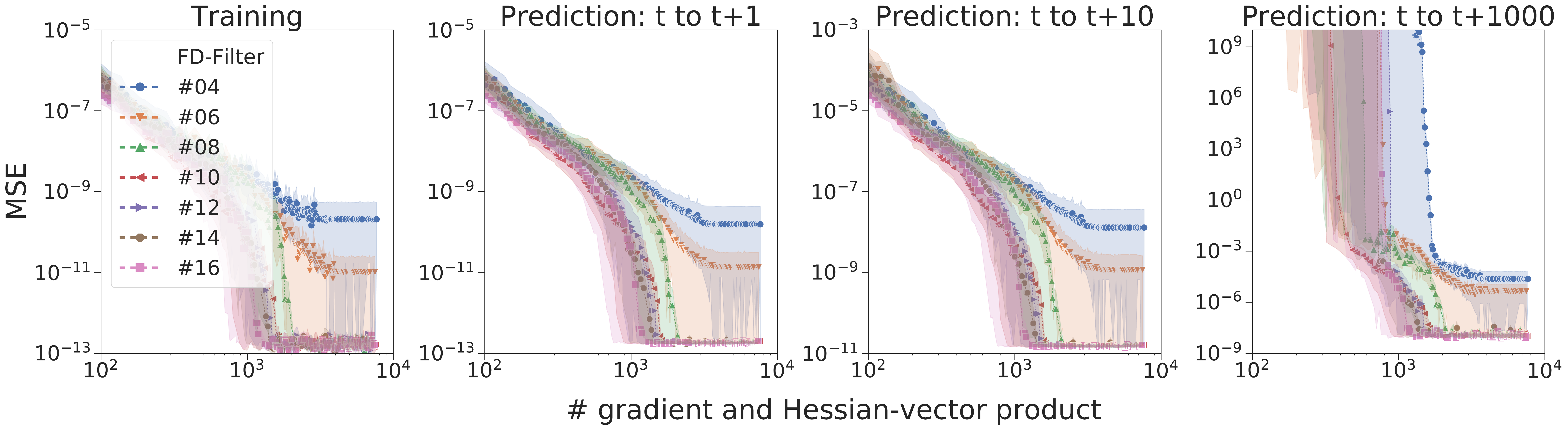}
  \vspace{0.2in}
  \centering
  \includegraphics[width=0.9\textwidth]{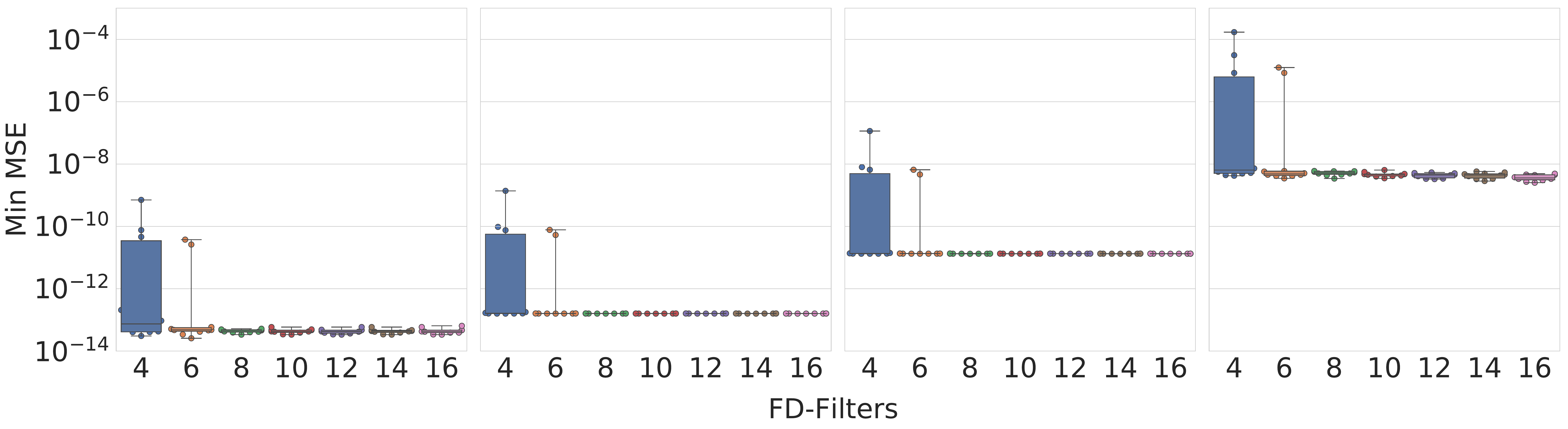}
  \caption{\textbf{Stable Case:} The evolution (first row) and the minimum (second row) of the training errors and testing errors of the one-, multi- and 1000-step predictions by configuration of \FDFilters{}.}
  \vspace*{3in}
  \label{fig.stable_filter}
\end{figure}

In this section, we present the experimental results of the stable case. Figure \ref{fig.stable_train} shows the evolution of the training errors of different configurations. Figure \ref{fig.stable_u} shows the sequential predictions and the squared errors, and Figure \ref{fig.stable_1000} shows the minimum testing errors (over the training process) of the 1000-step predictions by configuration. In addition, Figure \ref{fig.stable_scatter} shows the relationship between the training and testing errors (note: the lower and to the left is better). For the two supplementary testing procedures described in Section \ref{app_sup_test}, given the data of solution at $t$, the one-step prediction is made at $t+\Delta t$, i.e., $t+1$, and the multi-step prediction is made at $t + 10\Delta t$, i.e., $t+10$. Figures \ref{fig.stable_1_evol} \& \ref{fig.stable_10_evol} show the evolutions of the testing errors \eqref{eq12} and Figures \ref{fig.stable_1} \& \ref{fig.stable_10} show the minimum testing errors of the one- and multi-step predictions. To summarize the testing performance, we put the minimum testing errors aggregated over all configurations in Figure \ref{fig.stable_all}. Besides, the results of the sensitivity analysis on different numbers of \FDFilters{} (and $1$ \FDBlock{}) are shown in Figure \ref{fig.stable_filter}, which shows the evolution and the minimum (over the training process) of the training and testing errors.

\clearpage

\subsection{Unstable Case}

\begin{figure}[ht!]
  \centering
  \includegraphics[width=1.0\textwidth]{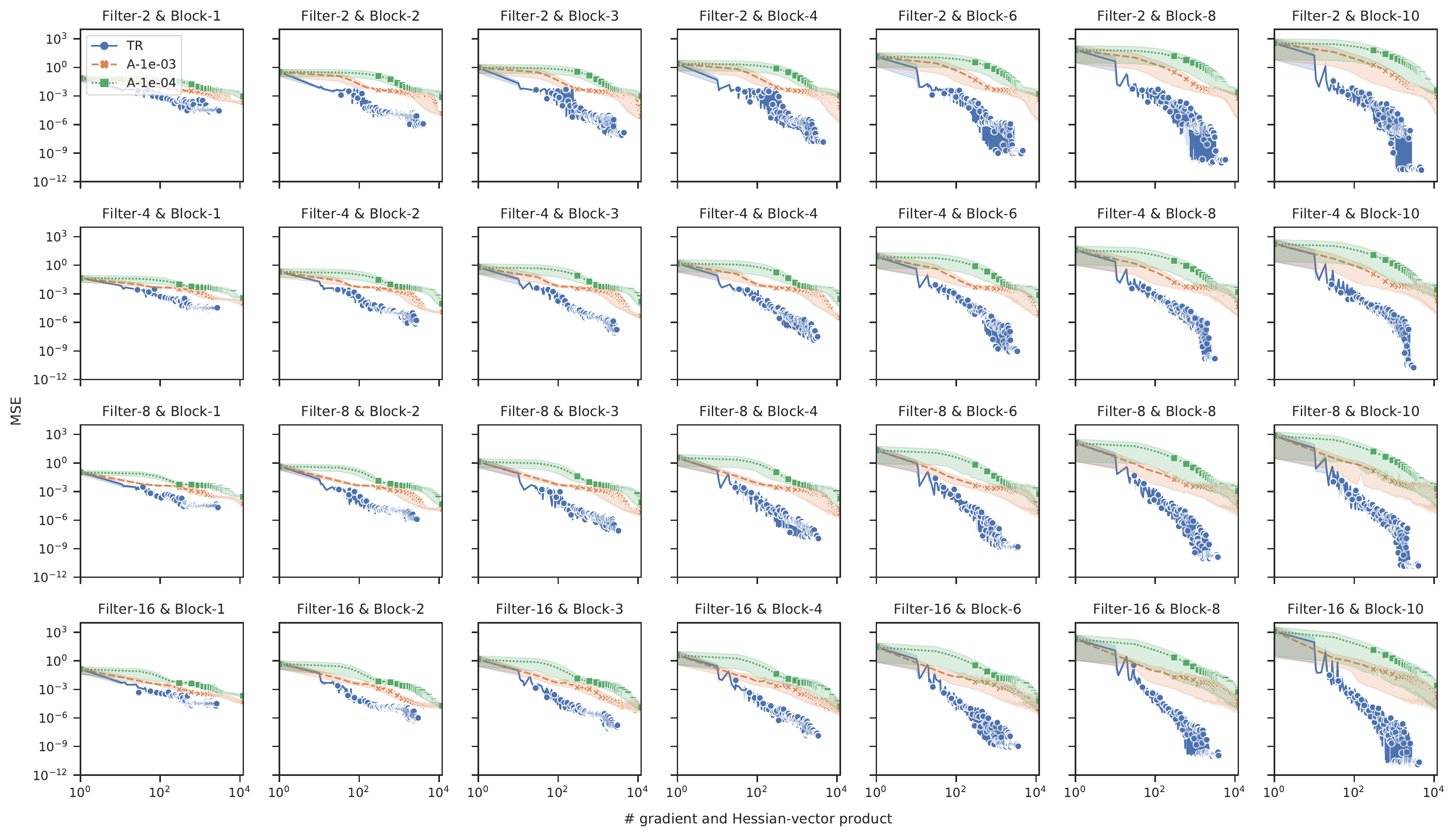}
  \caption{\textbf{Unstable Case:} Evolution of the MSE loss of stochastic mini-batch by configuration.}
  \label{fig.unstable_train}
  \vspace{0.5in}
  \centering
  \includegraphics[width=0.9\textwidth]{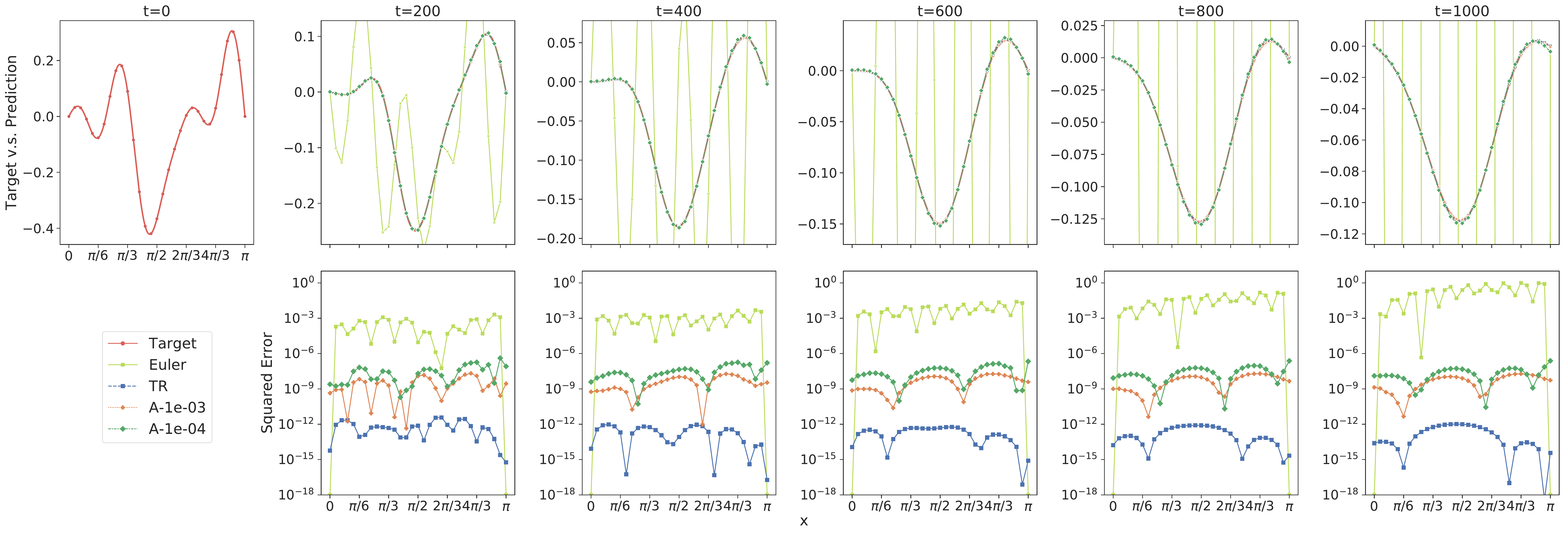}
  \caption{\textbf{Unstable Case:} Sequence of predictions.}
  \label{fig.unstable_u}
\end{figure}

\begin{figure}
  \centering
  \includegraphics[width=1.0\textwidth]{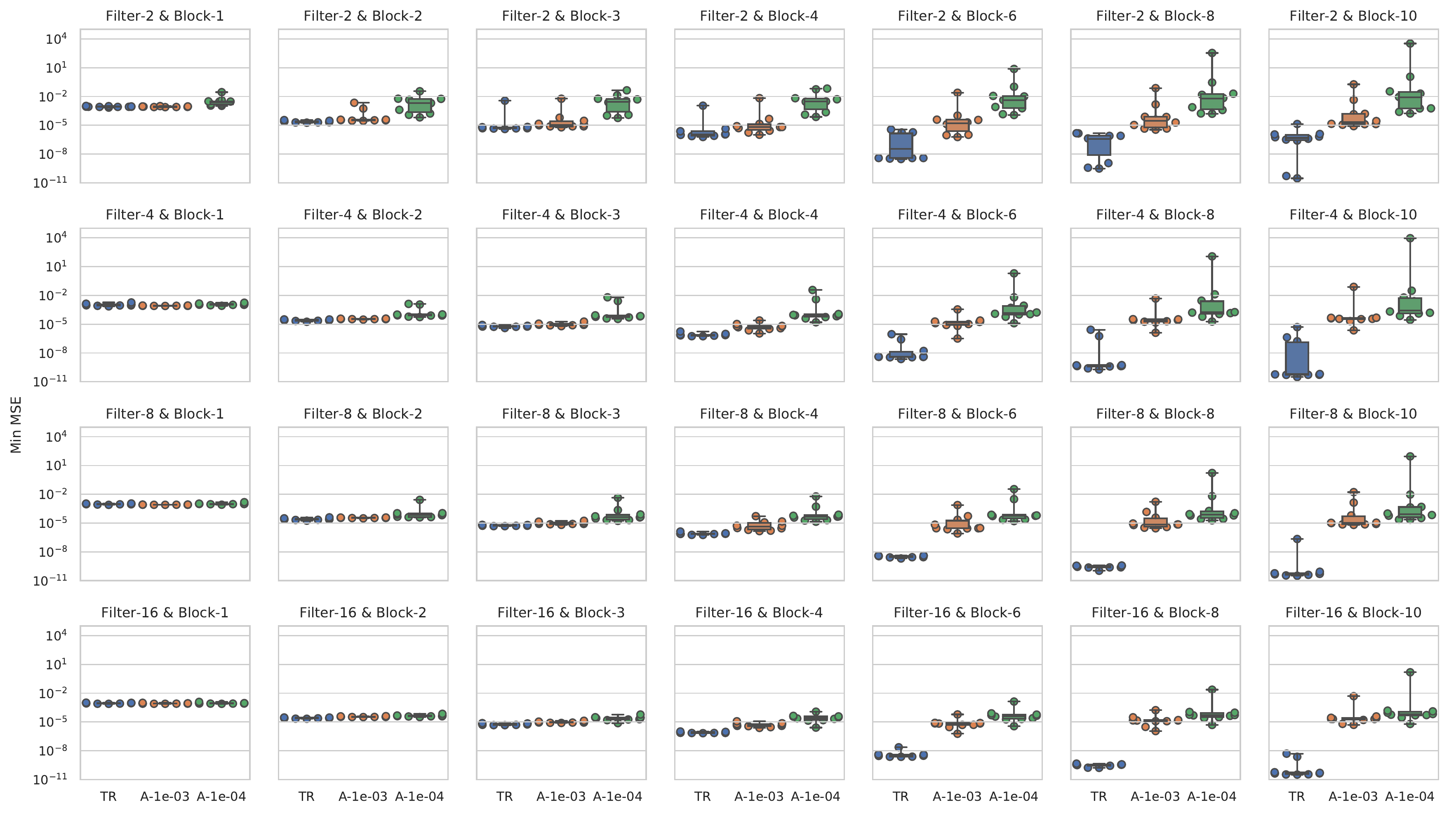}
  \caption{\textbf{Unstable Case:} Minimum testing errors by configuration - 1000-step prediction.}
  \label{fig.unstable_1000}
  \vspace{0.7in}
  \centering
  \includegraphics[width=1.0\textwidth]{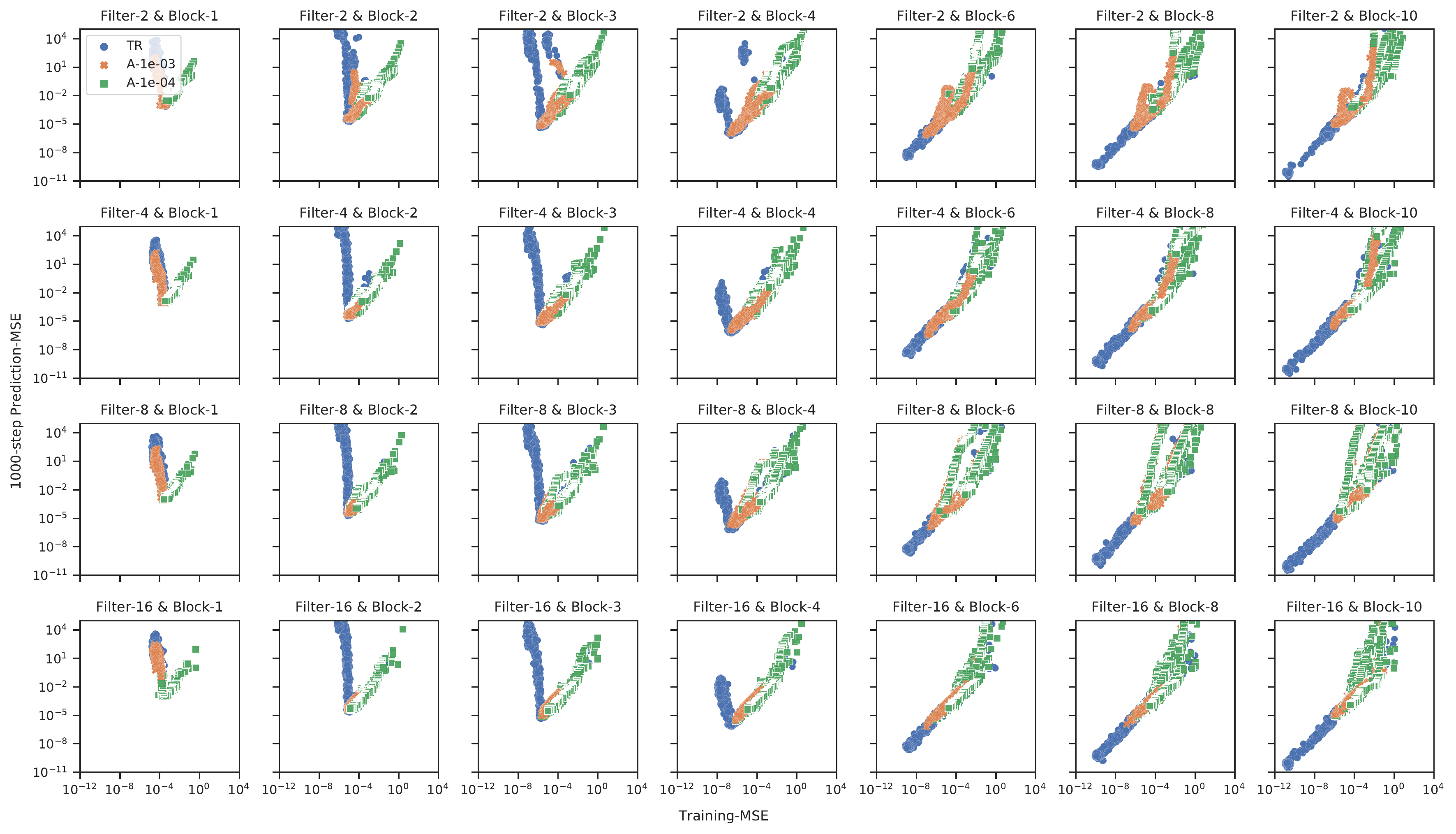}
  \caption{\textbf{Unstable Case:} Training vs. 1000-step prediction errors (over the training process) by configuration.}
  \vspace*{3in}
  \label{fig.unstable_scatter}
\end{figure}

\begin{figure}
  \centering
  \includegraphics[width=1.0\textwidth]{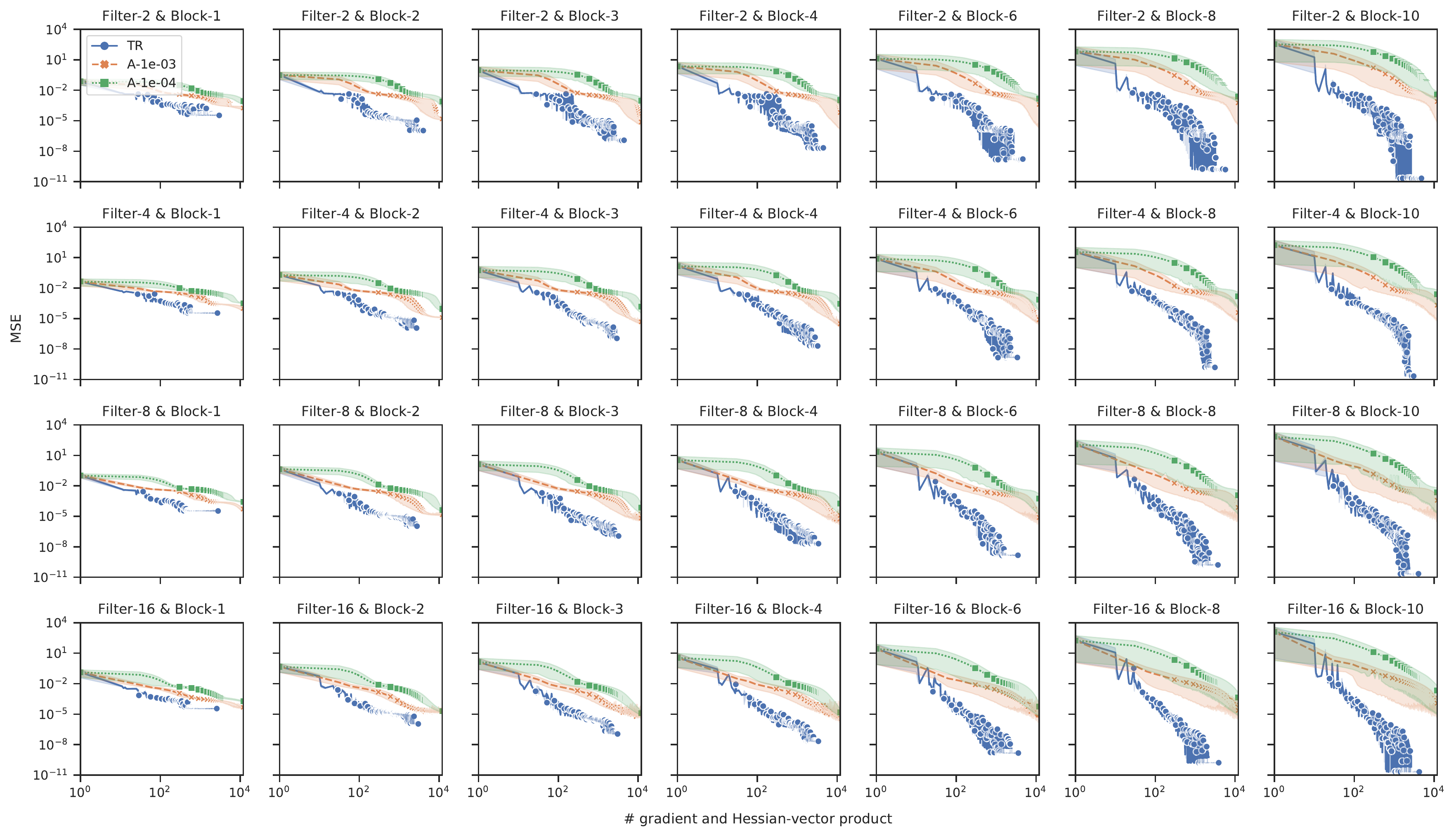}
  \caption{\textbf{Unstable Case:} Evolution of the testing errors by configuration - one-step prediction.}
  \label{fig.unstable_1_evol}
  \vspace{0.7in}
  \centering
  \includegraphics[width=1.0\textwidth]{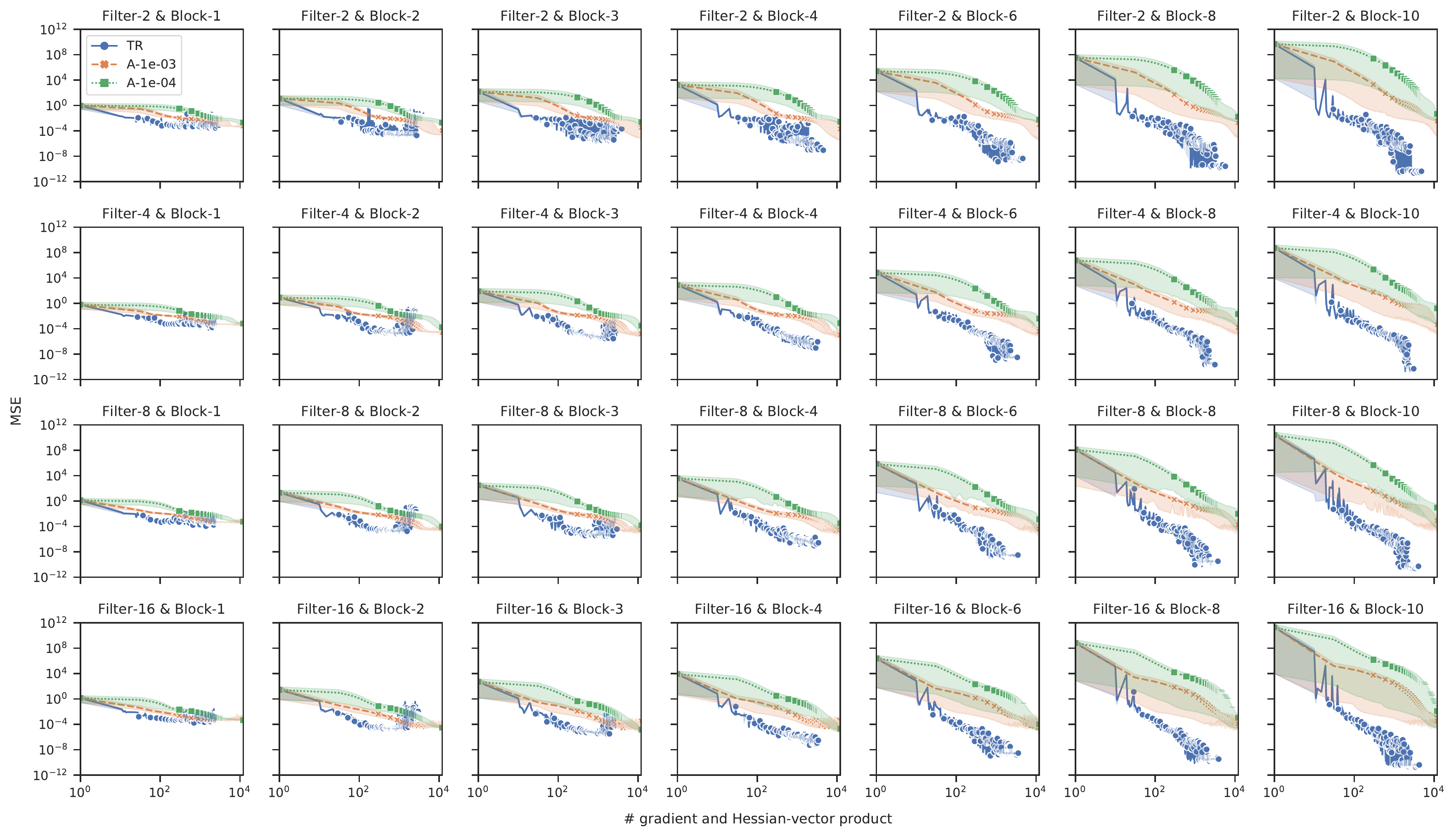}
  \caption{\textbf{Unstable Case:} Evolution of the testing errors by configuration - multi-step prediction.}
    \vspace*{3in}
  \label{fig.unstable_10_evol}
\end{figure}

\begin{figure}
  \centering
  \includegraphics[width=1.0\textwidth]{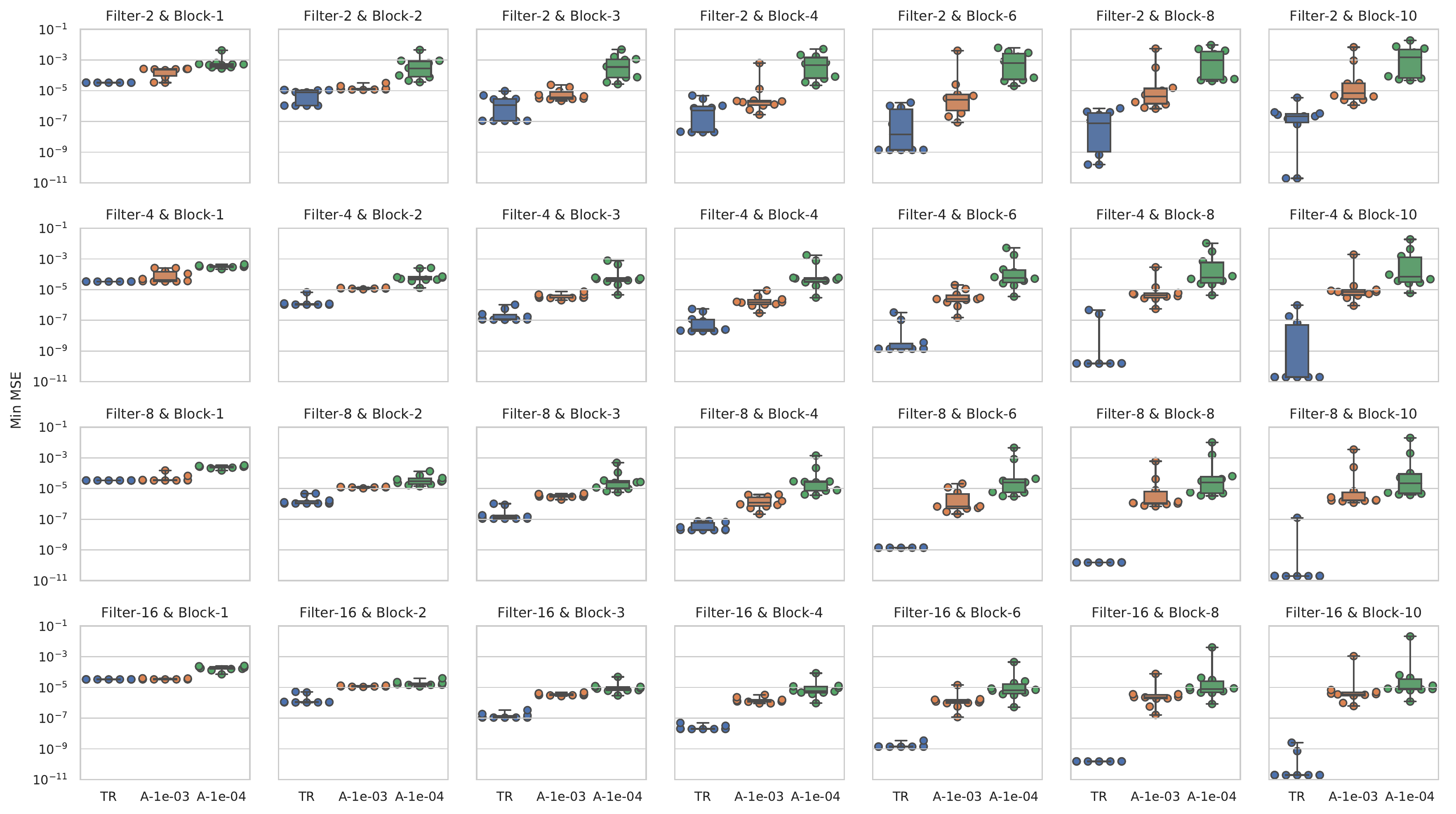}
  \caption{\textbf{Unstable Case:} Minimum testing errors by configuration - one-step prediction.}
  \label{fig.unstable_1}
  \vspace{0.7in}
  \centering
  \includegraphics[width=1.0\textwidth]{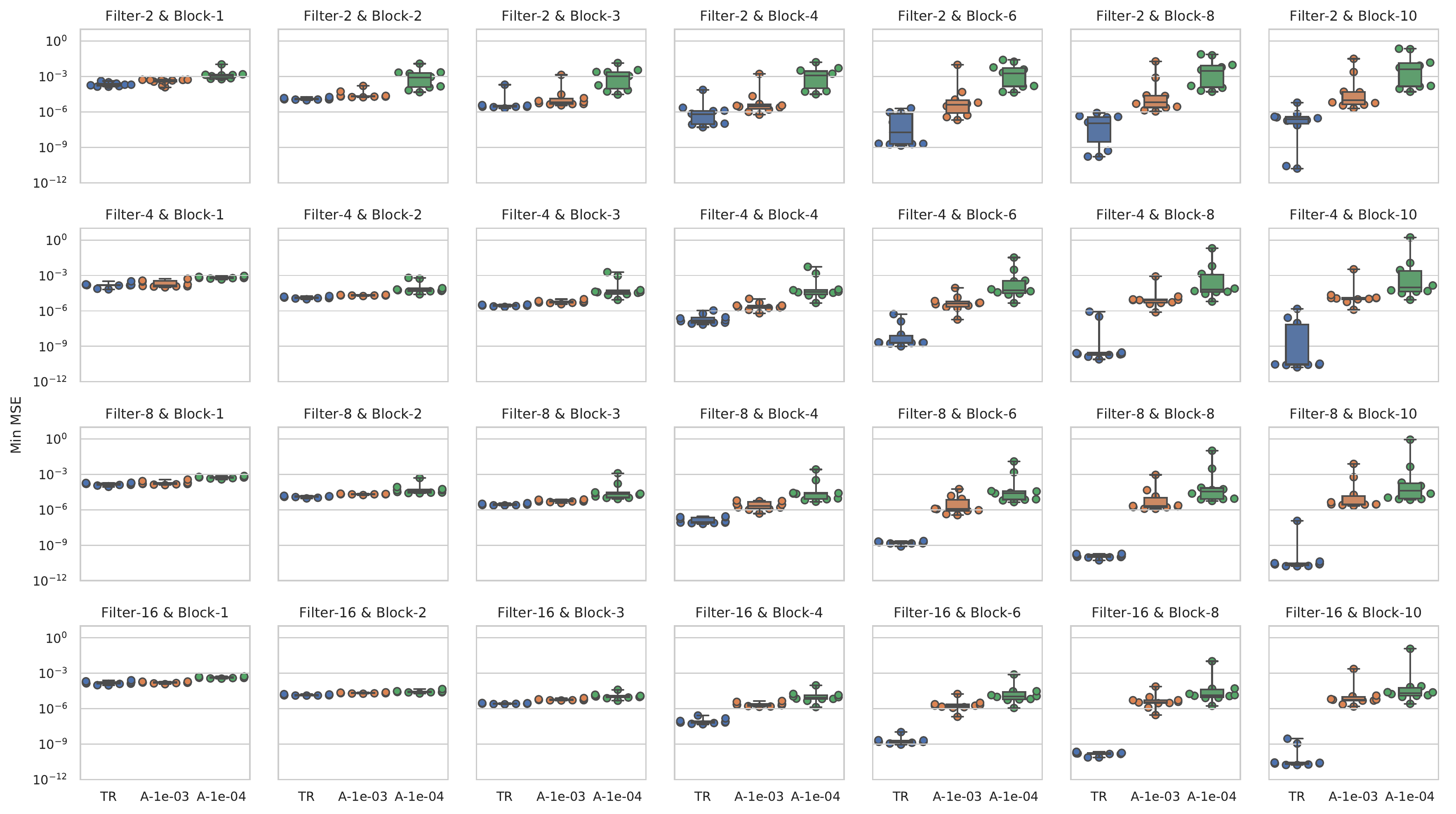}
  \caption{\textbf{Unstable Case:} Minimum testing errors by configuration - multi-step prediction.}
  \vspace*{3in}
  \label{fig.unstable_10}
\end{figure}

\begin{figure}[t!]
  \centering
  \includegraphics[width=0.3\textwidth]{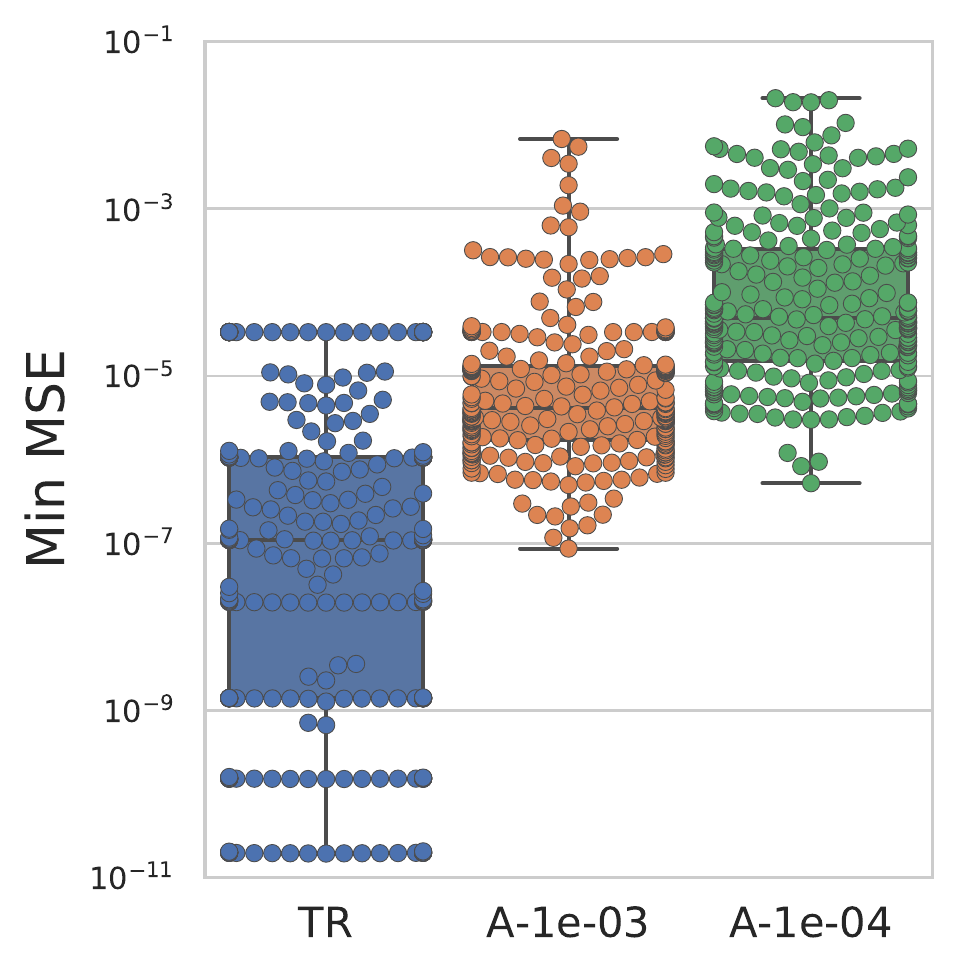}
  \includegraphics[width=0.3\textwidth]{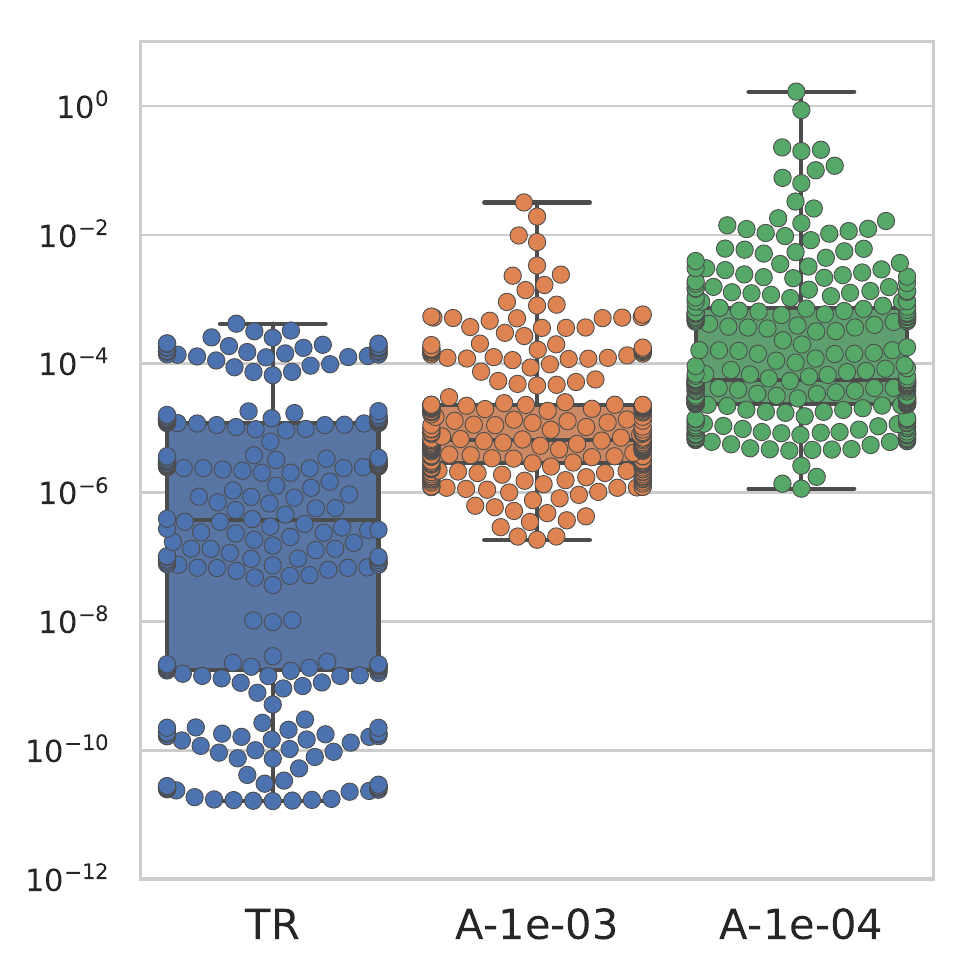}
  \includegraphics[width=0.3\textwidth]{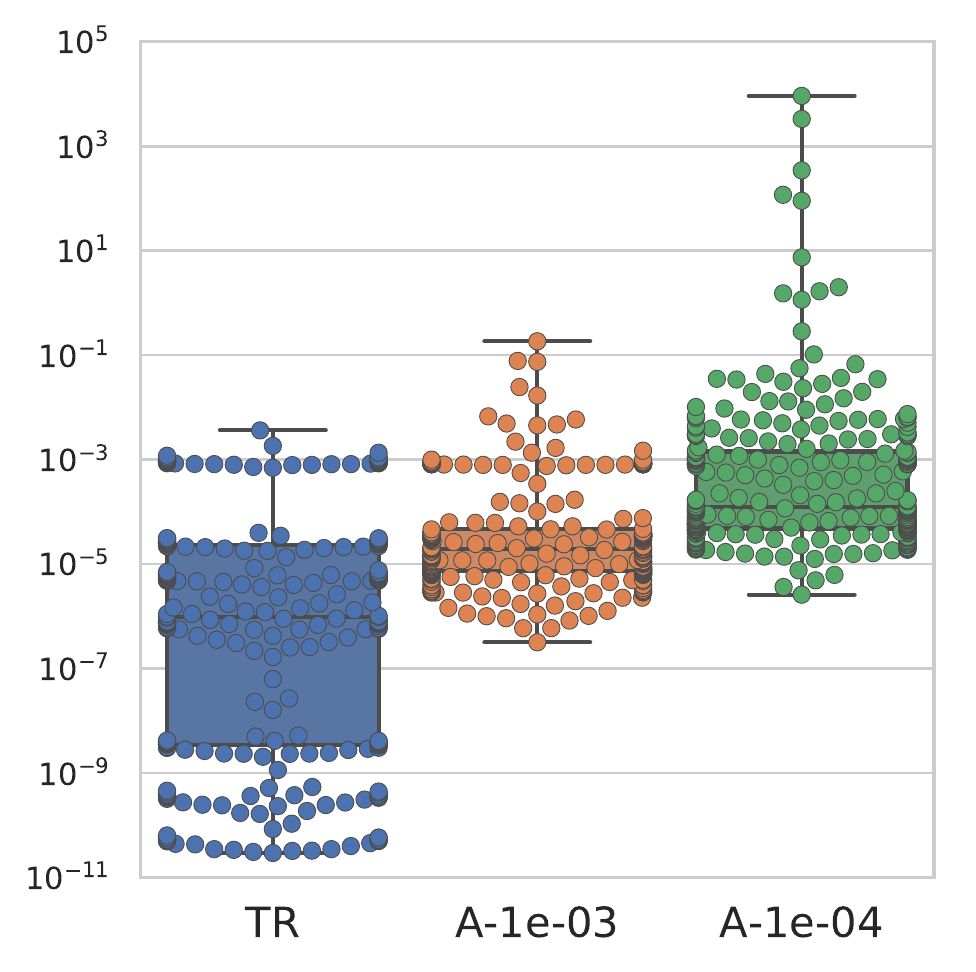}
  \caption{\textbf{Unstable Case:} The minimum testing errors of the one-step (left), multi-step (middle) and 1000-step (right) predictions (aggregated over all configurations).}
  \label{fig.unstable_all}
  \vspace{0.5in}
  \centering
  \includegraphics[width=0.9\textwidth]{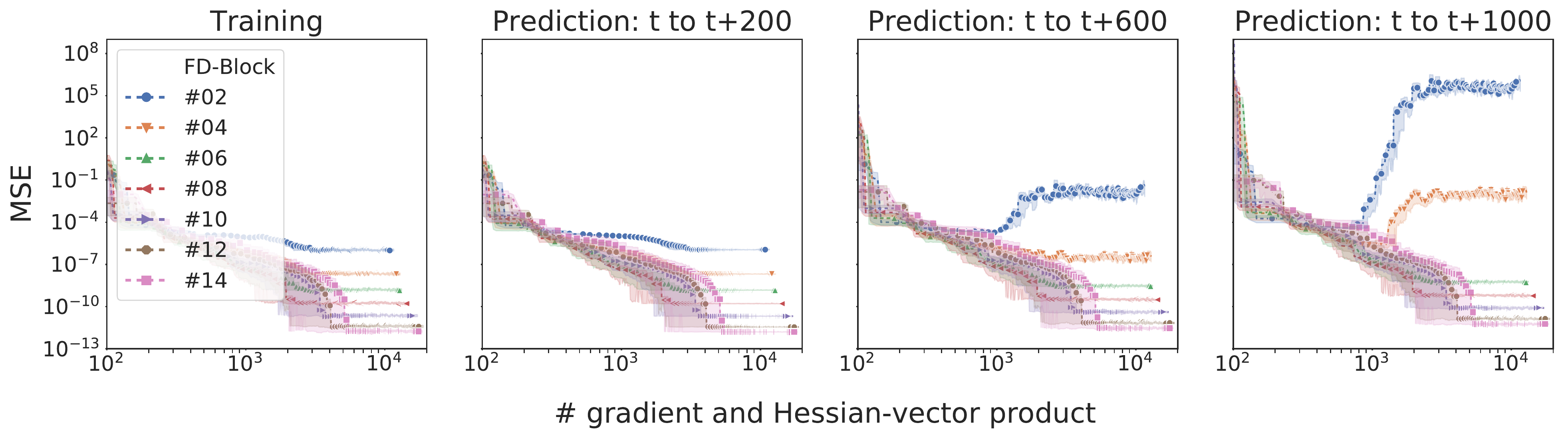}
  \vspace{0.2in}
  \centering
  \includegraphics[width=0.9\textwidth]{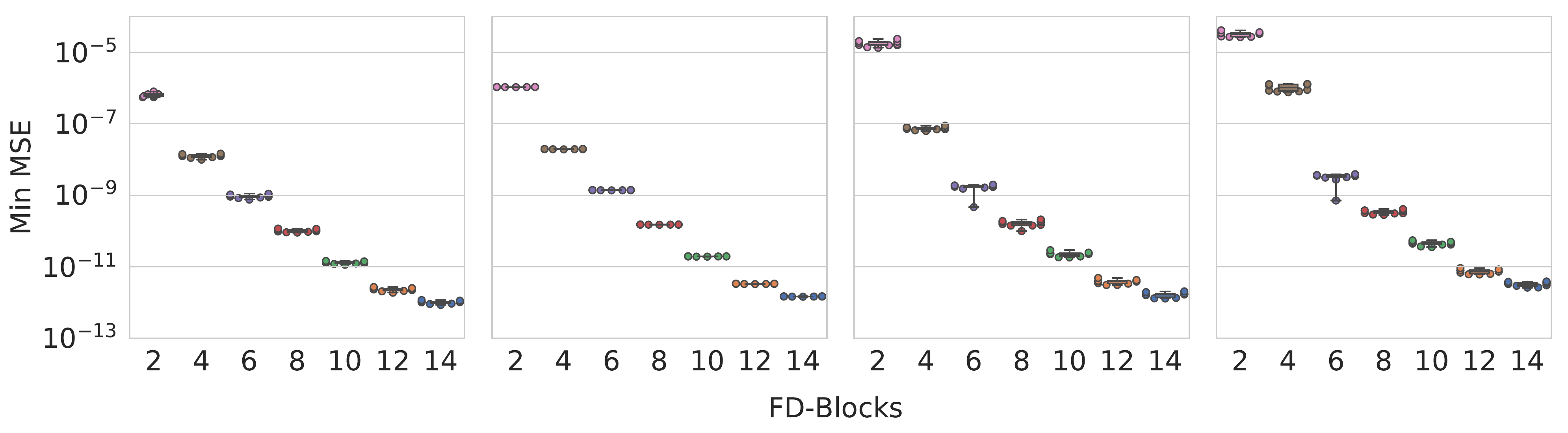}
  \caption{\textbf{Unstable Case:} The evolution (first row) and the minimum (second row) of the training errors and testing errors of the one-, multi- and 1000-step prediction by configuration of \FDBlocks{}.}
  \vspace*{3in}
  \label{fig.unstable_block}
\end{figure}

In this section, we present the experimental results of the unstable case. Figure \ref{fig.unstable_train} shows the evolution of the training errors of different configurations. Figure \ref{fig.unstable_u} shows the sequential predictions and the squared errors, and Figure \ref{fig.unstable_1000} shows the minimum testing errors (over the training process) of the 1000-step predictions by configuration. In addition, Figure \ref{fig.unstable_scatter} shows the relationship between the training and testing errors (note: the lower and to the left is better). For the two supplementary testing procedures described in Section \ref{app_sup_test}, given the data of solution at $t$, the one-step prediction is made at $t+\Delta t$, i.e., $t+200$, and the multi-step prediction is made at $t + 3\Delta t$, i.e., $t+600$. Figures \ref{fig.unstable_1_evol} \& \ref{fig.unstable_10_evol} show the evolutions of the testing errors \eqref{eq12} and Figures \ref{fig.unstable_1} \& \ref{fig.unstable_10} show the minimum testing errors of the one- and multi-step predictions. To summarize the testing performance, we put the minimum testing errors aggregated over all configurations in Figure \ref{fig.unstable_all}. Besides, the results of the sensitivity analysis on different numbers of \FDBlocks{} (and $16$ \FDFilters{}) are shown in Figure \ref{fig.unstable_block}, which shows the evolution and the minimum (over the training process) of the training and testing errors.

\clearpage

\subsection{Forcing Case}

\begin{figure}[ht!]
  \centering
  \includegraphics[width=0.7\textwidth]{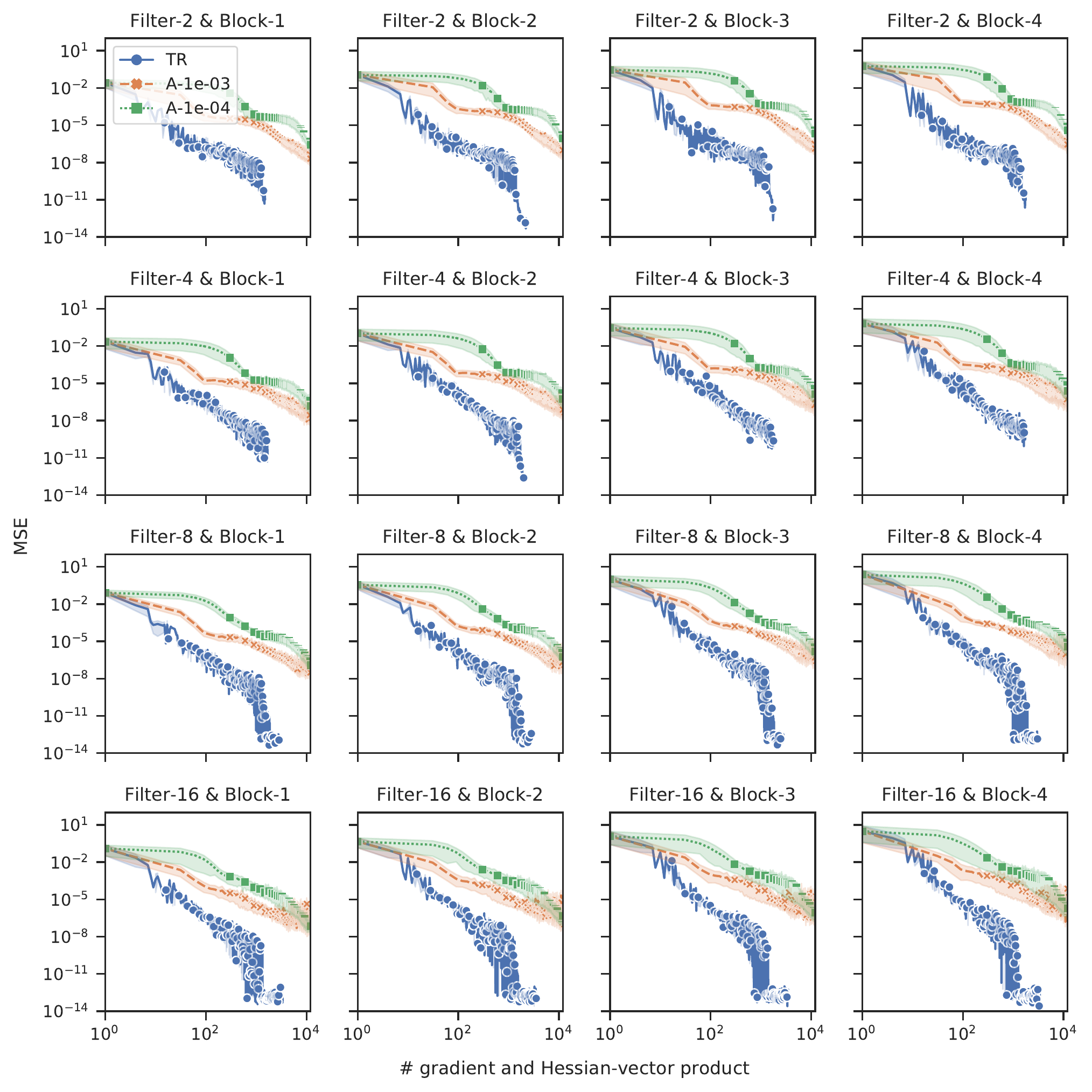}
  \caption{\textbf{Forcing Case:} Evolution of the MSE loss of stochastic mini-batch by configuration.}
  \label{fig.force_train}
  \vspace{0.5in}
  \centering
  \includegraphics[width=0.9\textwidth]{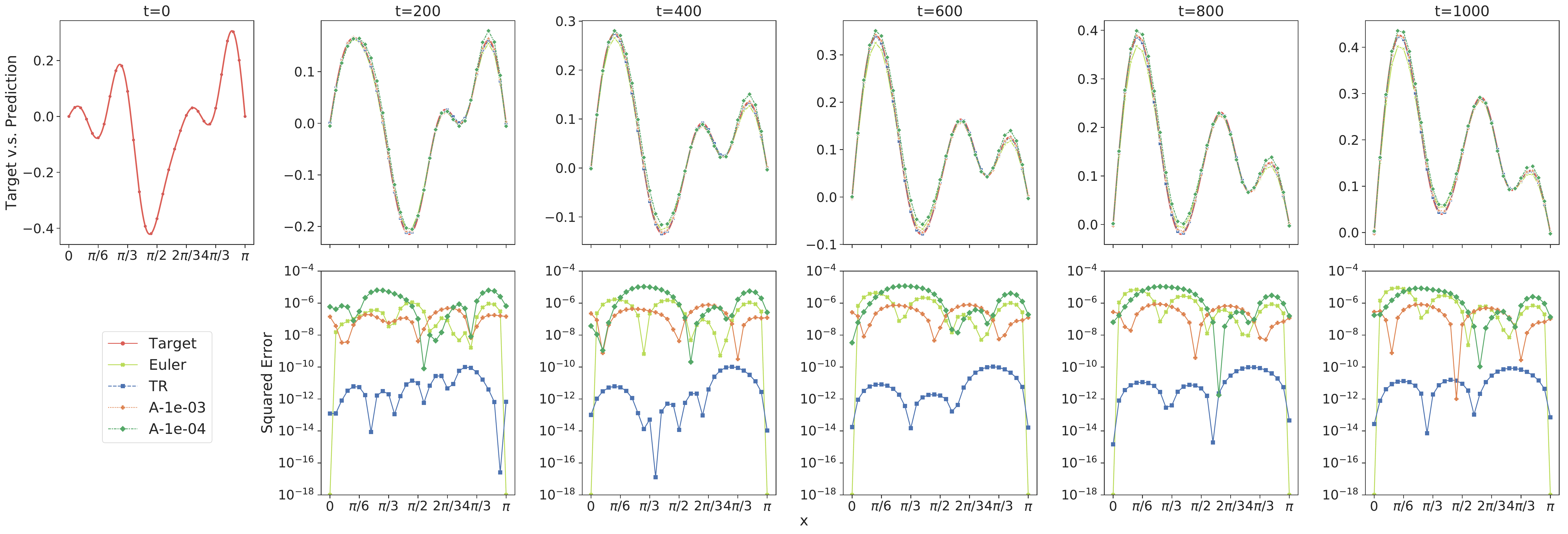}
  \caption{\textbf{Forcing Case:} Sequence of predictions.}
  \label{fig.force_u}
\end{figure}

\begin{figure}
  \centering
  \includegraphics[width=0.7\textwidth]{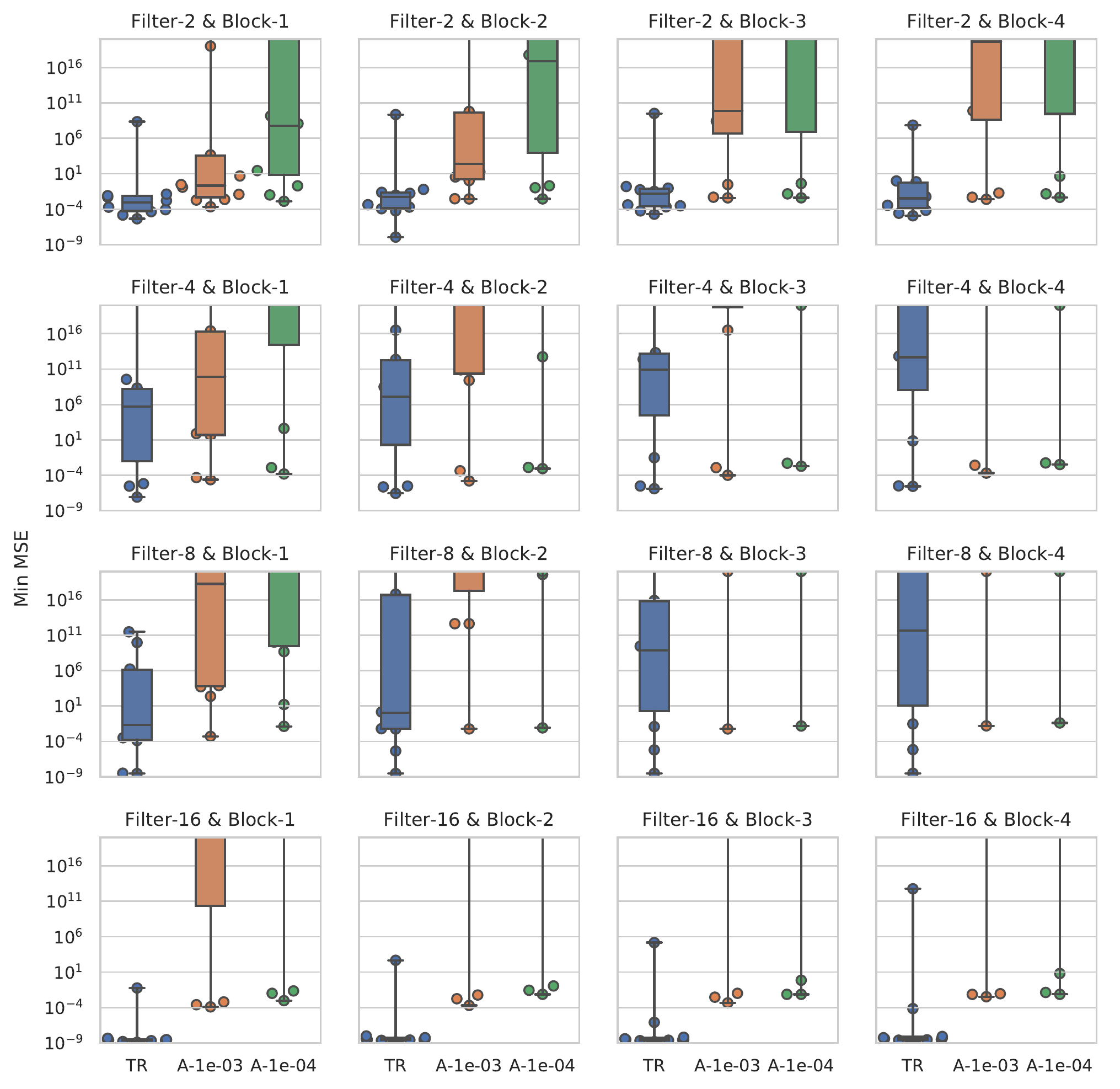}
  \caption{\textbf{Forcing Case:} Minimum testing errors by configuration - 1000-step prediction.}
  \label{fig.force_1000}
  \vspace{0.7in}
  \centering
  \includegraphics[width=0.7\textwidth]{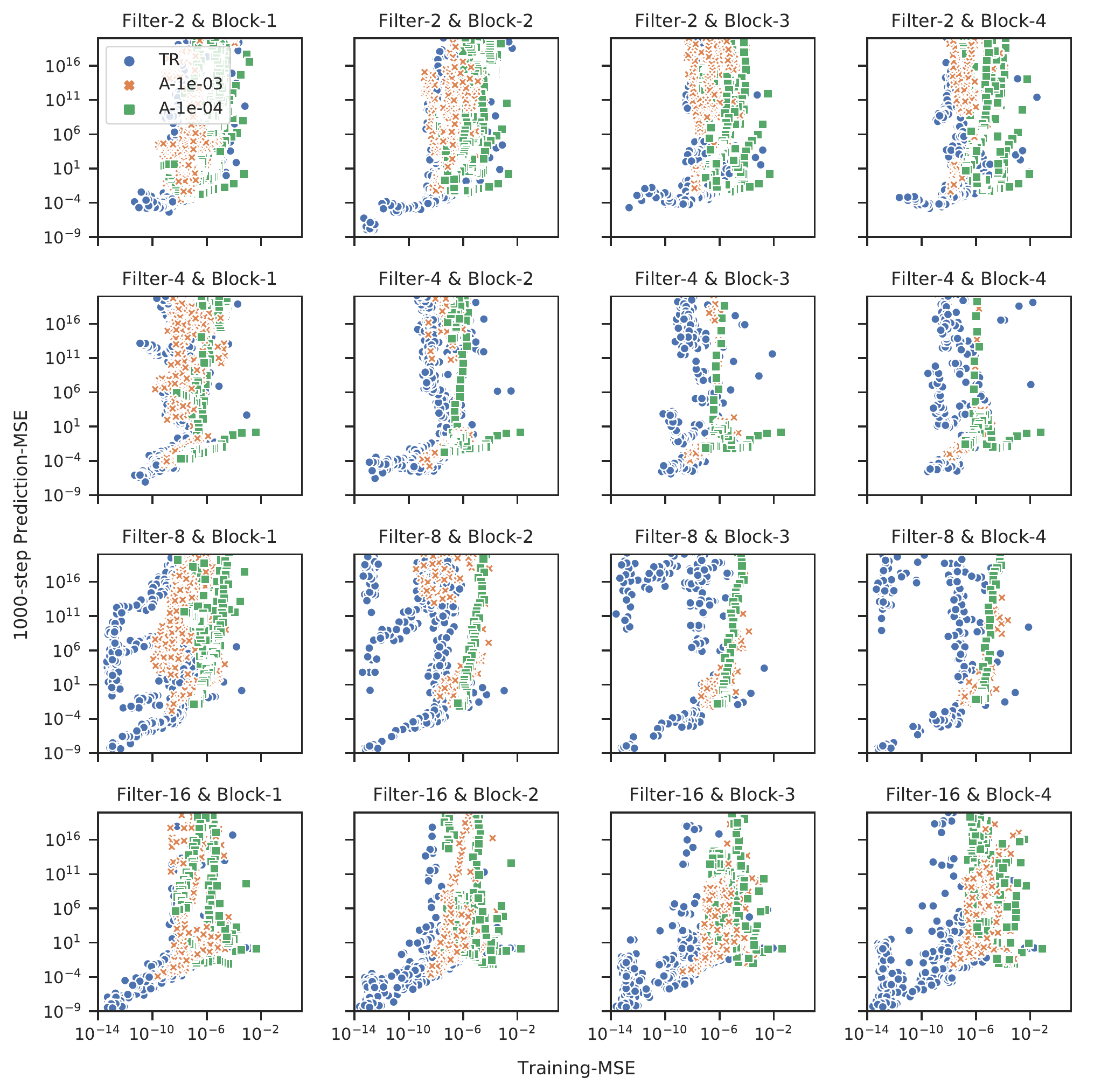}
  \caption{\textbf{Forcing Case:} Training vs. 1000-step prediction errors (over the training process) by configuration.}
   \vspace*{3in}
  \label{fig.force_scatter}
\end{figure}

\begin{figure}
  \centering
  \includegraphics[width=0.7\textwidth]{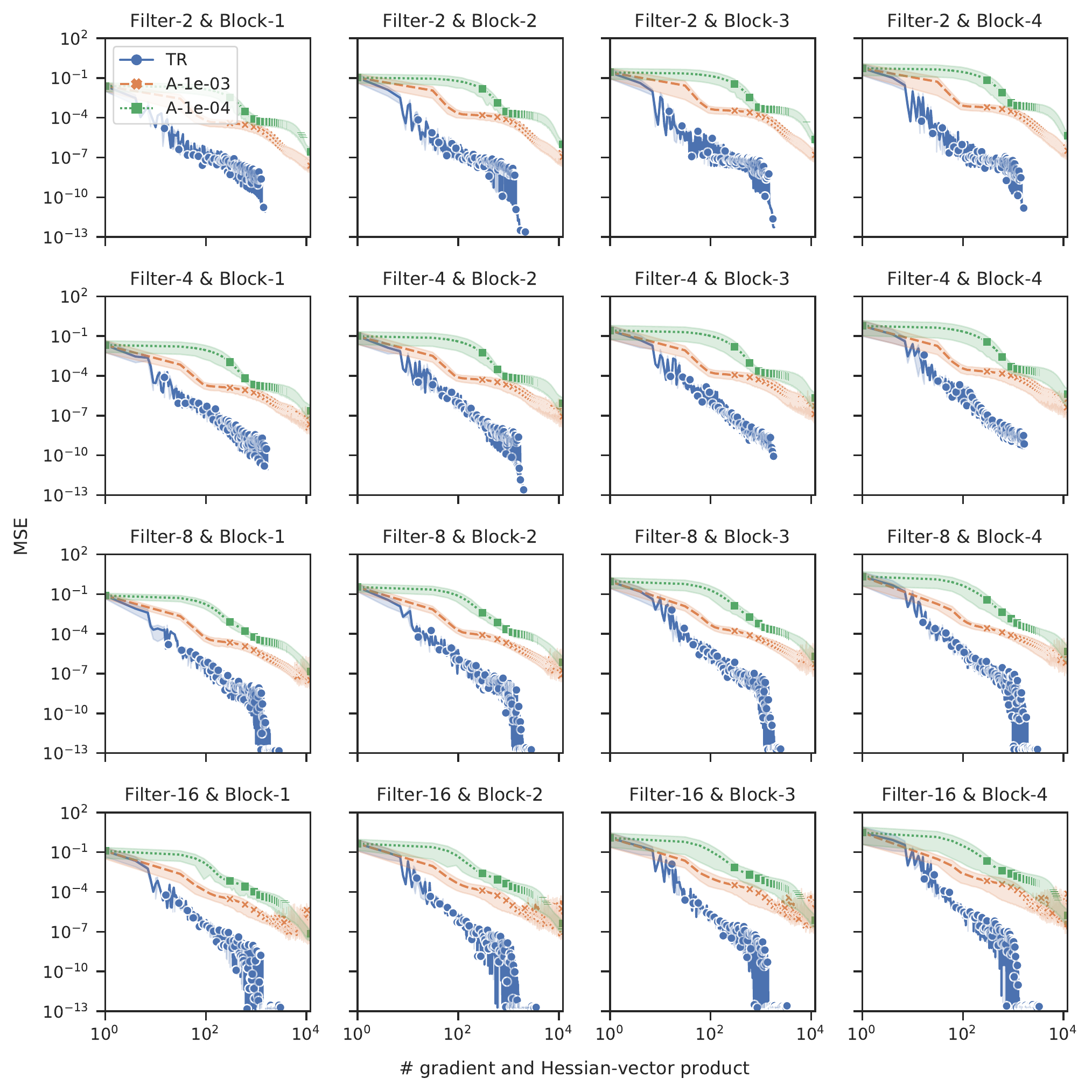}
  \caption{\textbf{Forcing Case:} Evolution of the testing errors by configuration - one-step prediction.}
  \label{fig.force_1_evol}
  \vspace{0.7in}
  \centering
  \includegraphics[width=0.7\textwidth]{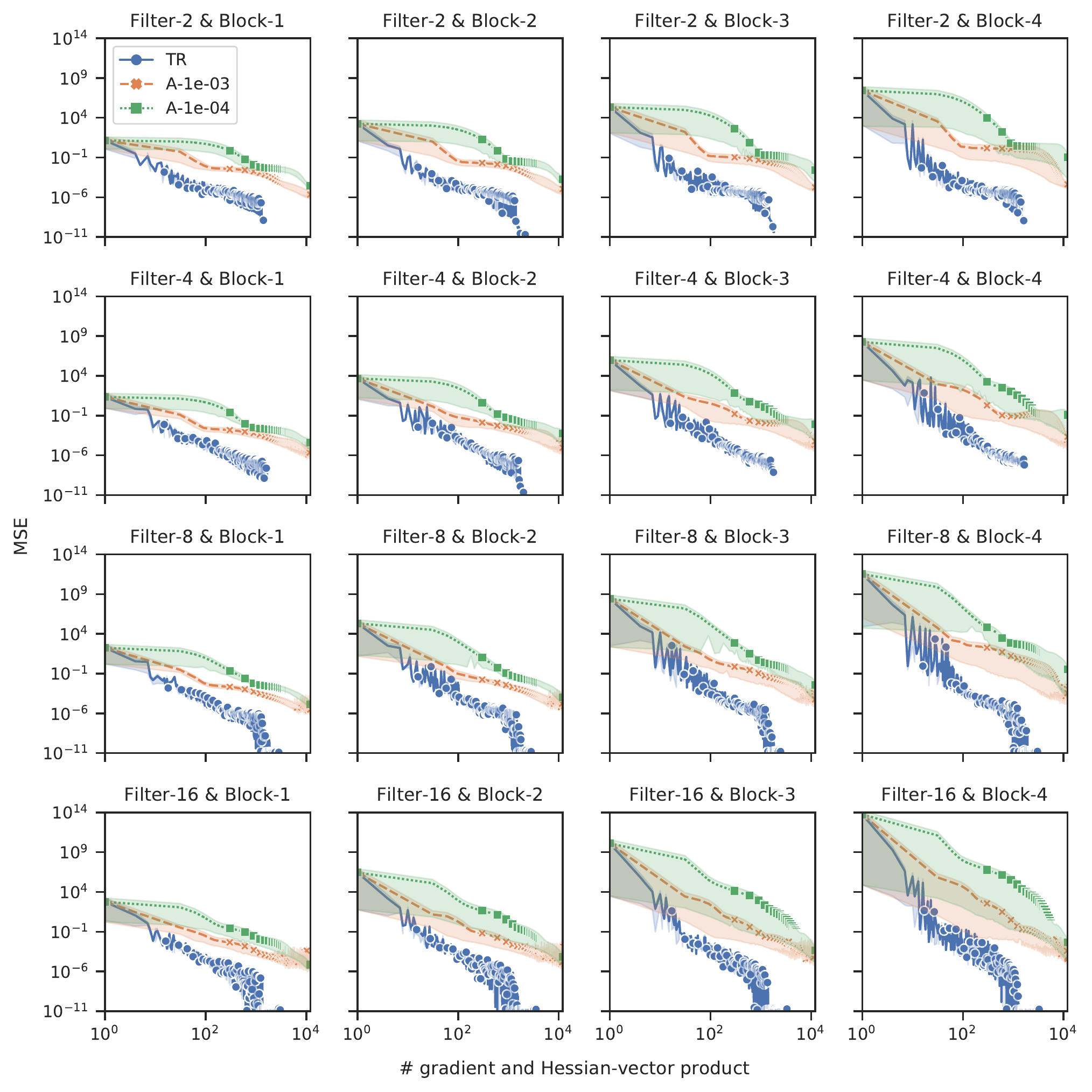}
  \caption{\textbf{Forcing Case:} Evolution of the testing errors by configuration - multi-step prediction.}
  \vspace*{3in}
  \label{fig.force_10_evol}
\end{figure}

\begin{figure}
  \centering
  \includegraphics[width=0.7\textwidth]{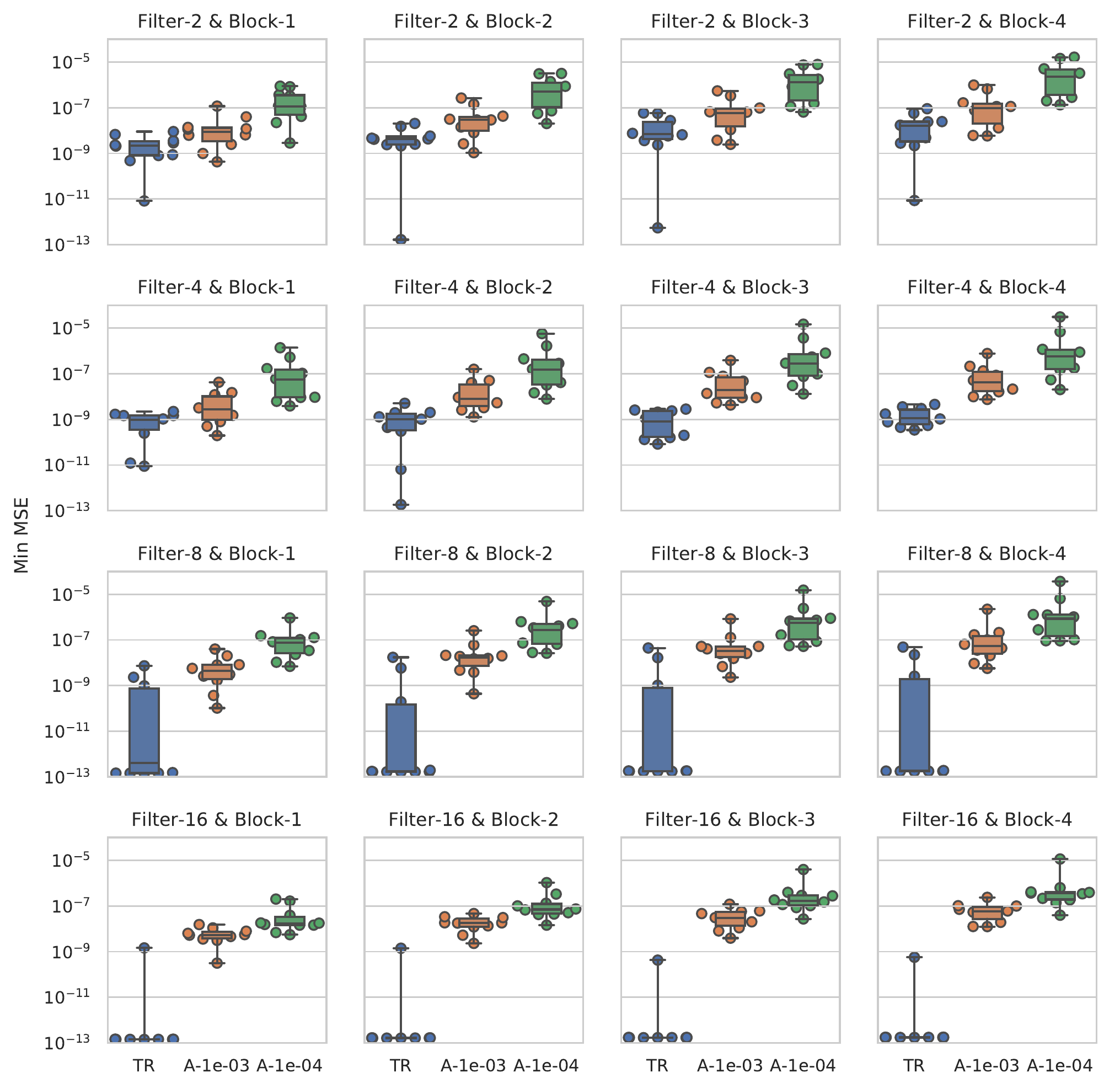}
  \caption{\textbf{Forcing Case:} Minimum testing errors by configuration - one-step prediction.}
  \label{fig.force_1}
  \vspace{0.7in}
  \centering
  \includegraphics[width=0.7\textwidth]{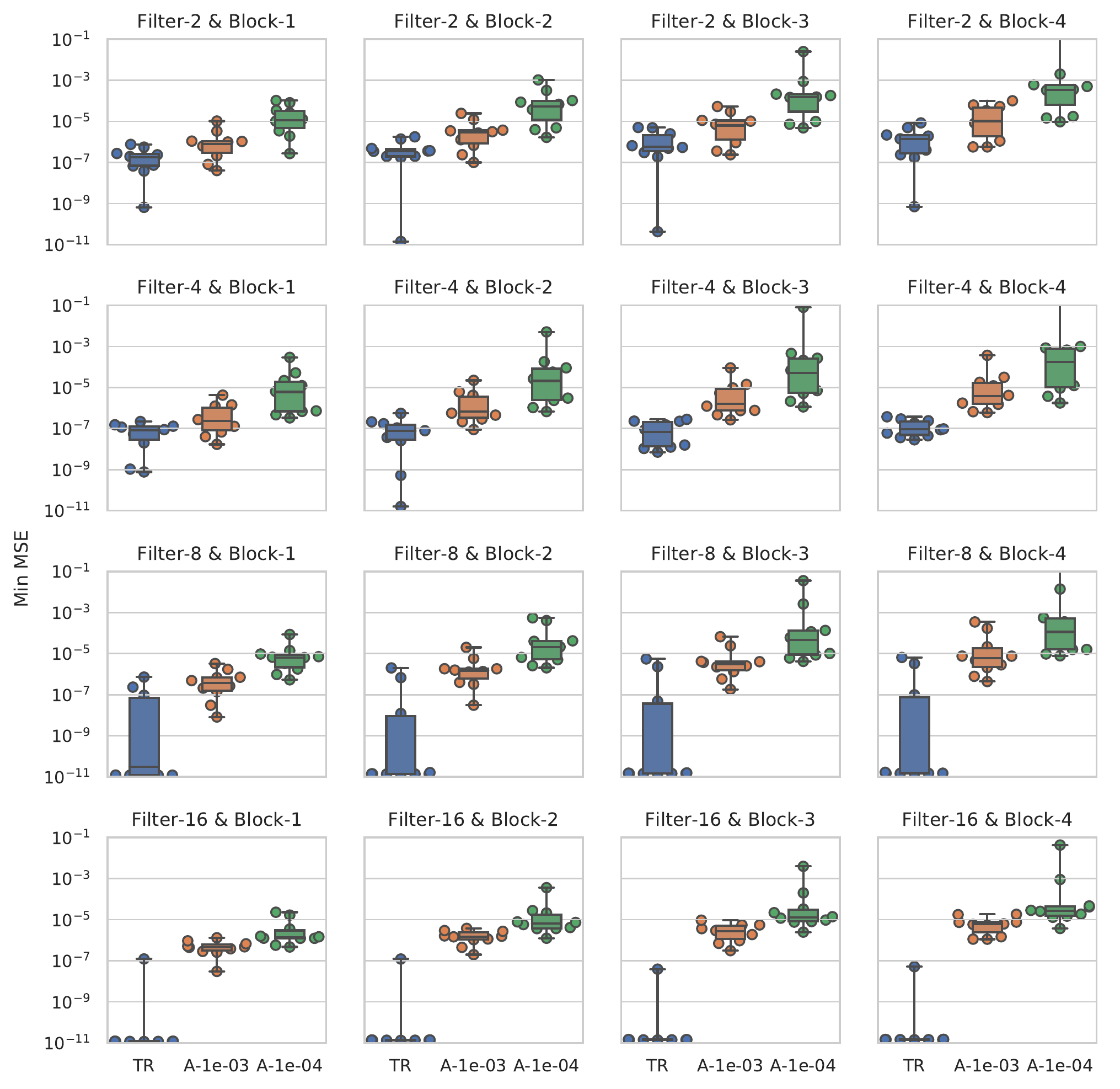}
  \caption{\textbf{Forcing Case:} Minimum testing errors by configuration - multi-step prediction.}
  \vspace*{3in}
  \label{fig.force_10}
\end{figure}

\begin{figure}[t!]
  \centering
  \includegraphics[width=0.3\textwidth]{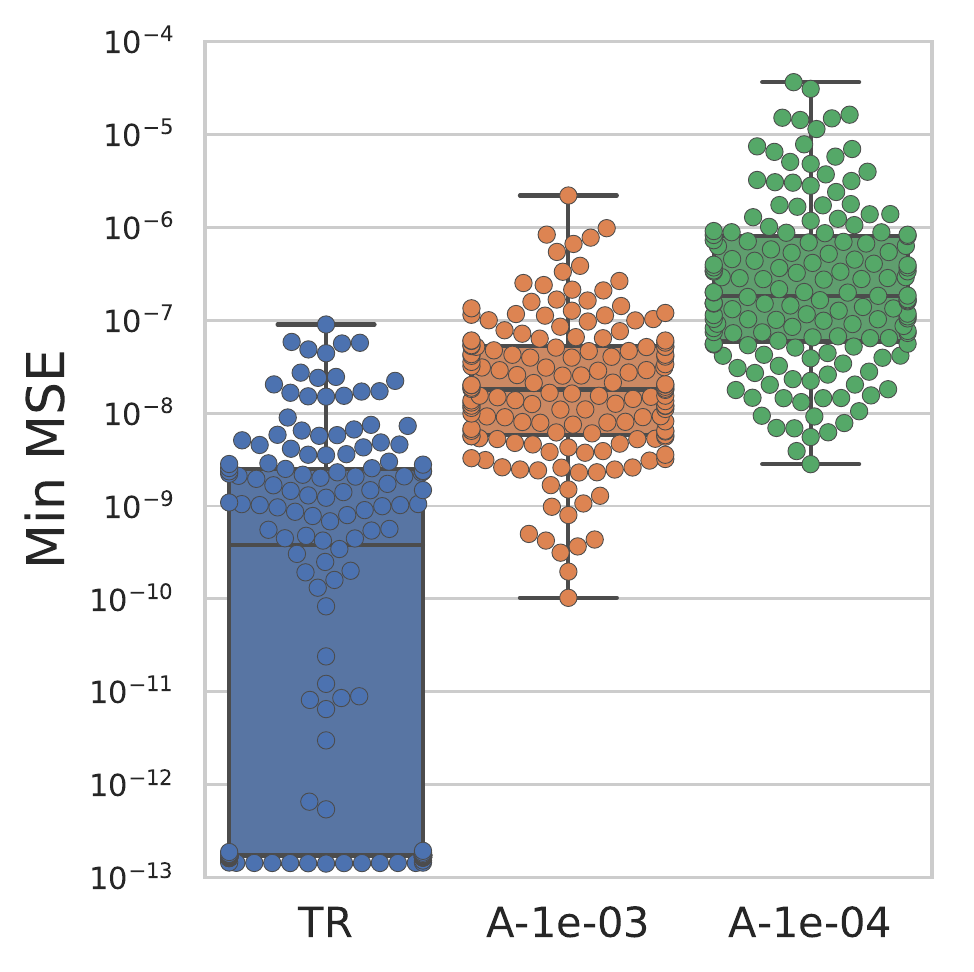}
  \includegraphics[width=0.3\textwidth]{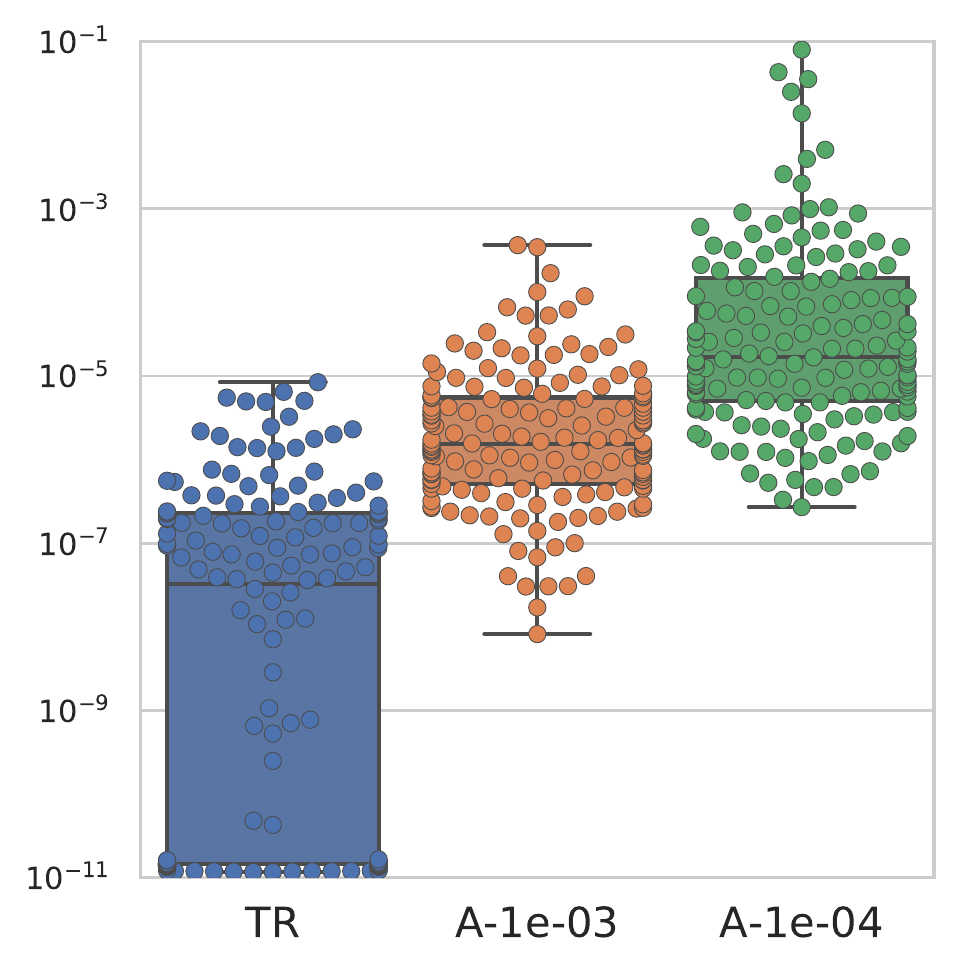}
  \includegraphics[width=0.3\textwidth]{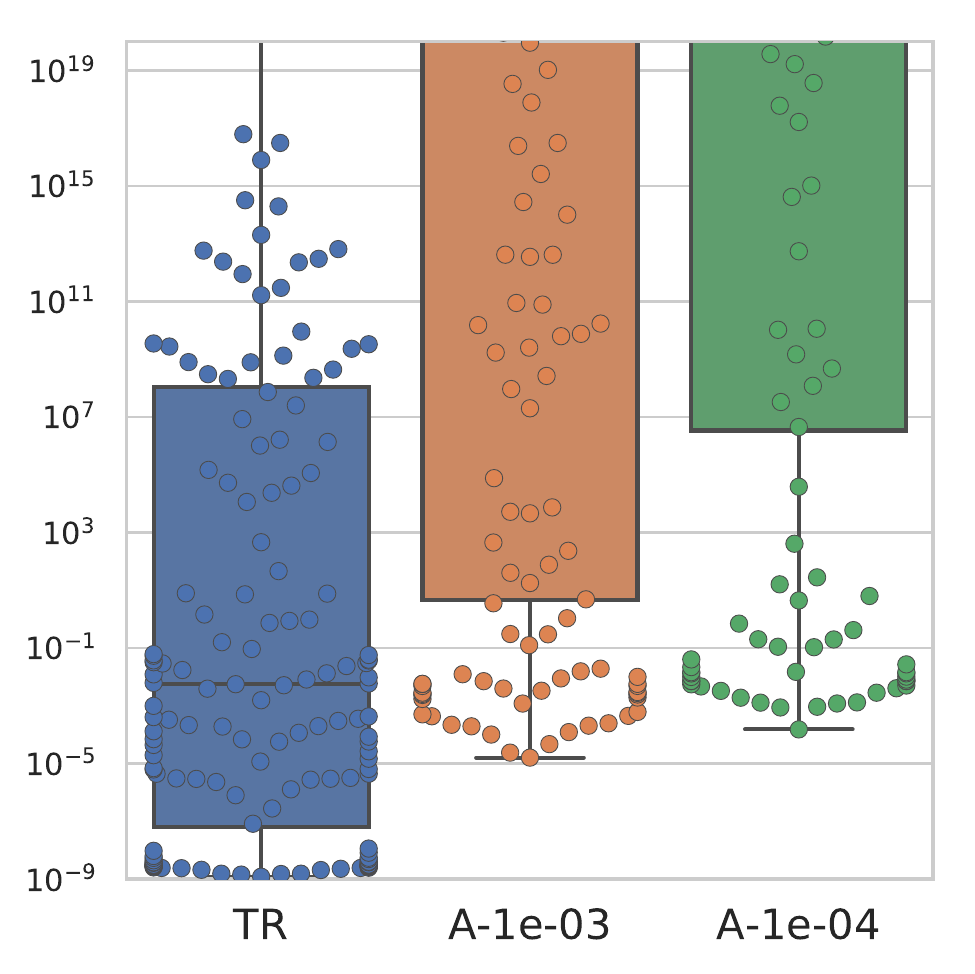}
  \caption{\textbf{Forcing Case:} The minimum testing errors of the one-step (left), multi-step (middle) and 1000-step (right) predictions (aggregated over all configurations).}
  \vspace*{6in}
  \label{fig.force_all} 
\end{figure}

In this section, we present the experimental results of the forcing case. Figure \ref{fig.force_train} shows the evolution of the training errors of different configurations. Figure \ref{fig.force_u} shows the sequential predictions and the squared errors, and Figure \ref{fig.force_1000} shows the minimum testing errors (over the training process) of the 1000-step predictions by configuration. In addition, Figure \ref{fig.force_scatter} shows the relationship between the training and testing errors (note: the lower and to the left is better). For the two supplementary testing procedures described in Section \ref{app_sup_test}, given the data of solution at $t$, the one-step prediction is made at $t+\Delta t$, i.e., $t+1$, and the multi-step prediction is made at $t + 10\Delta t$, i.e., $t+10$. Figures \ref{fig.force_1_evol} \& \ref{fig.force_10_evol} show the evolutions of the testing errors \eqref{eq12} and Figures \ref{fig.force_1} \& \ref{fig.force_10} show the minimum testing errors of the one- and multi-step predictions. To summarize the testing performance, we put the minimum testing errors aggregated over all configurations in Figure \ref{fig.force_all}. 

\clearpage

\subsection{Noisy Case}

\begin{figure}[ht!]
    \centering
    \begin{subfigure}{0.48\textwidth}
        \includegraphics[width=0.98\textwidth]{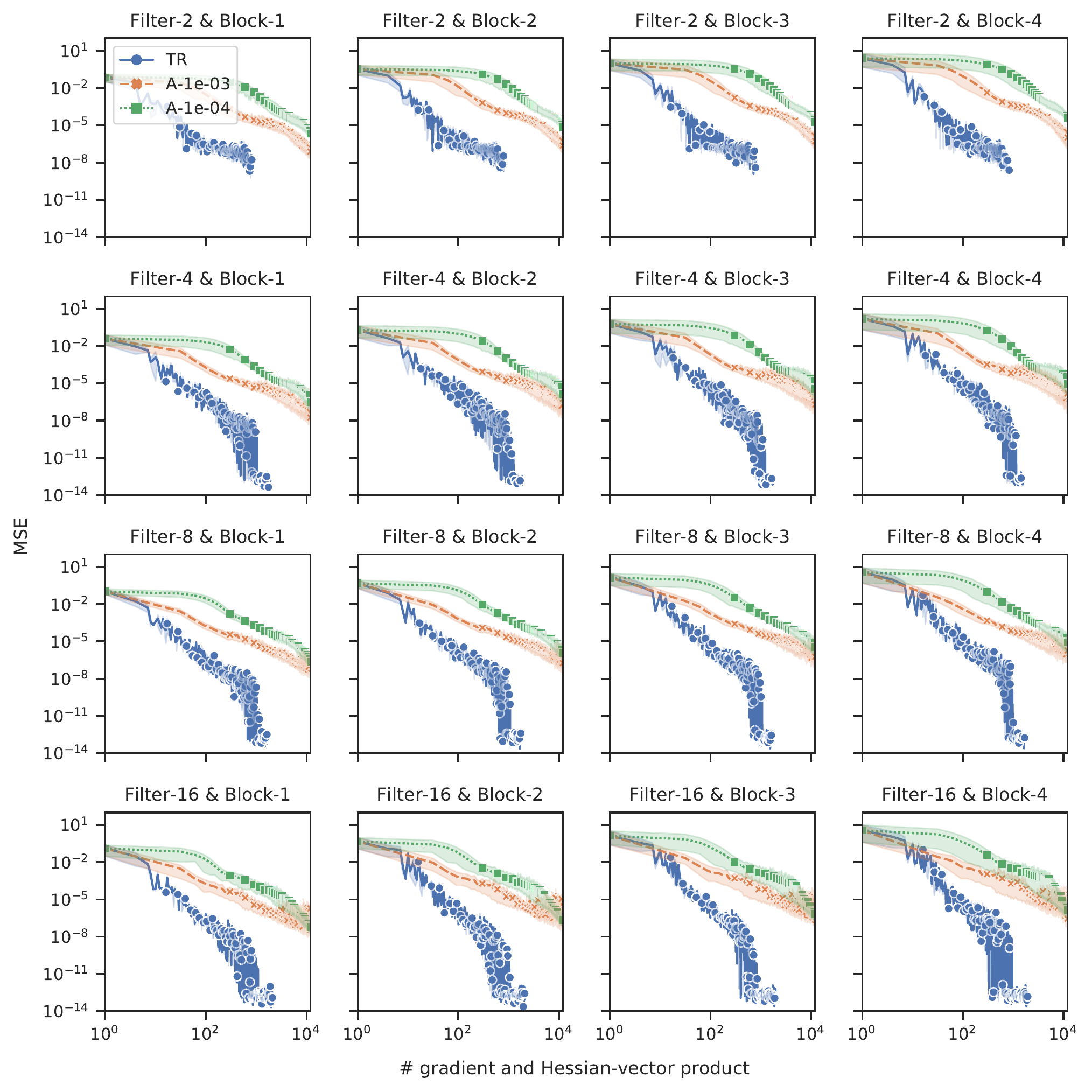}
        \caption{Low Noise}
    \end{subfigure}
    \centering
    \begin{subfigure}{0.48\textwidth}
        \includegraphics[width=0.98\textwidth]{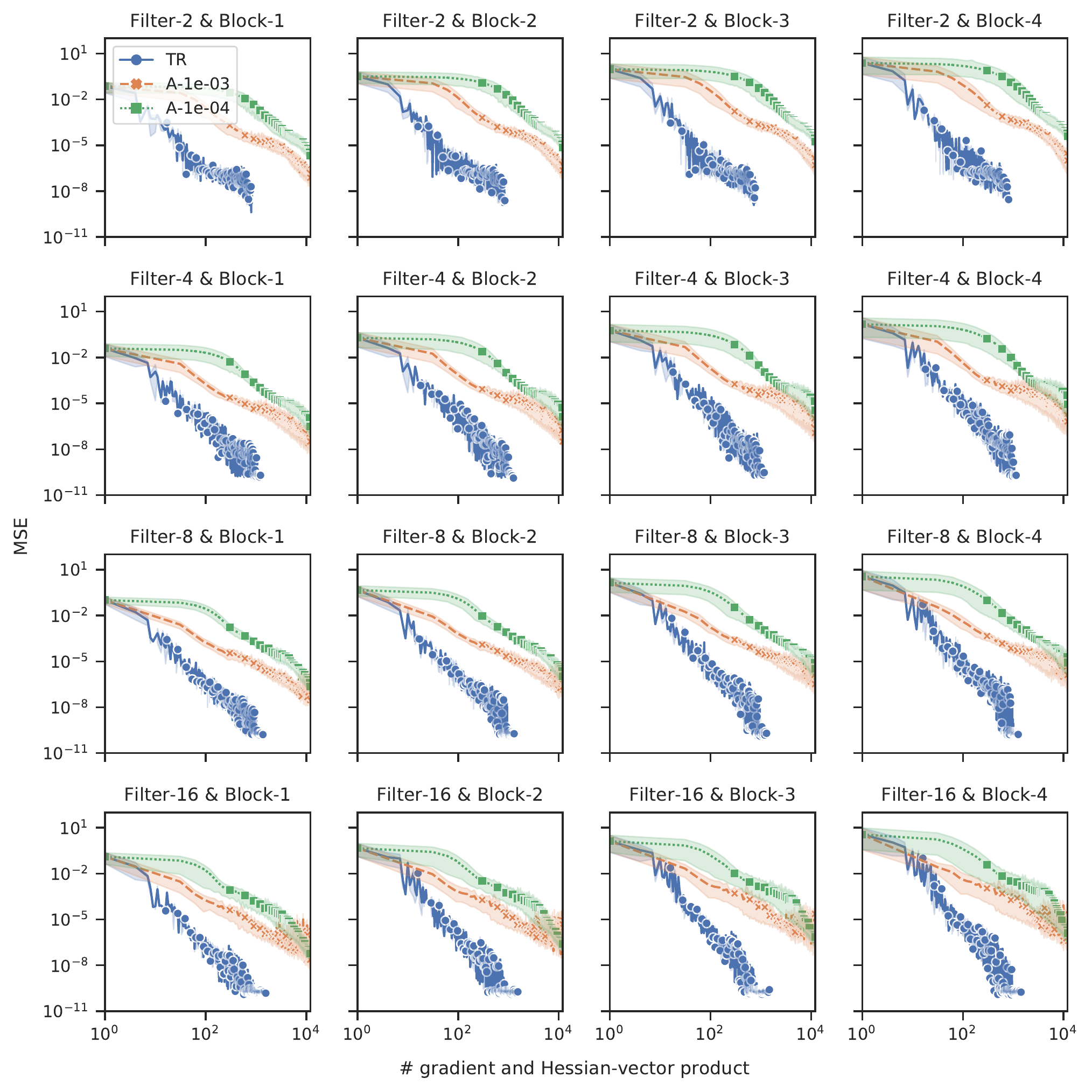}
        \caption{Medium Noise}
    \end{subfigure}\\[3ex]
    \centering
    \begin{subfigure}{0.48\textwidth}
        \includegraphics[width=0.98\textwidth]{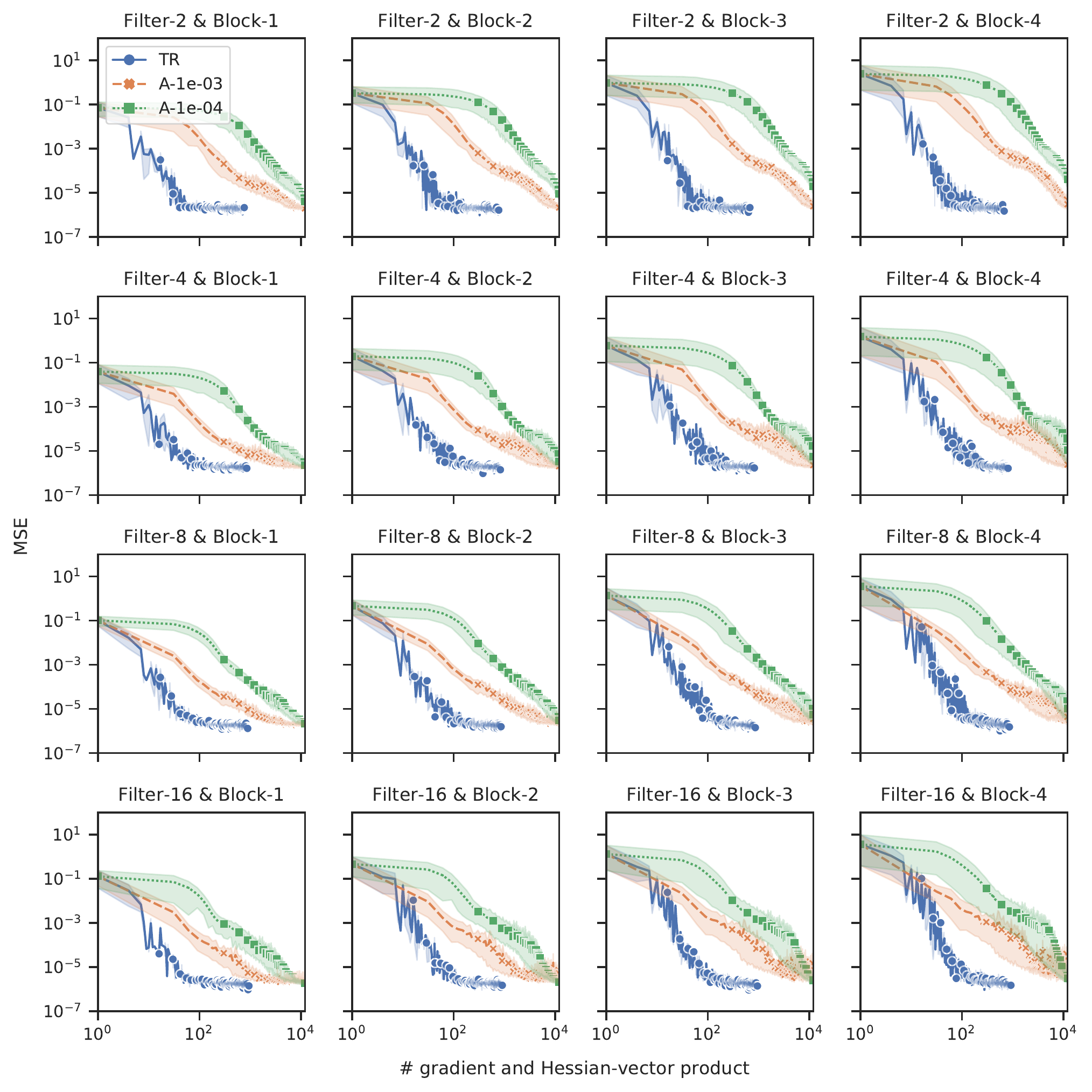}
        \caption{High Noise}
    \end{subfigure}
  \caption{\textbf{Noisy Case:} Evolution of the MSE loss of stochastic mini-batch by level of noise and configuration.}
  \label{fig.noisy_train}
\end{figure}

\begin{figure}
    \centering
    \begin{subfigure}{0.9\textwidth}
        \includegraphics[width=0.98\textwidth]{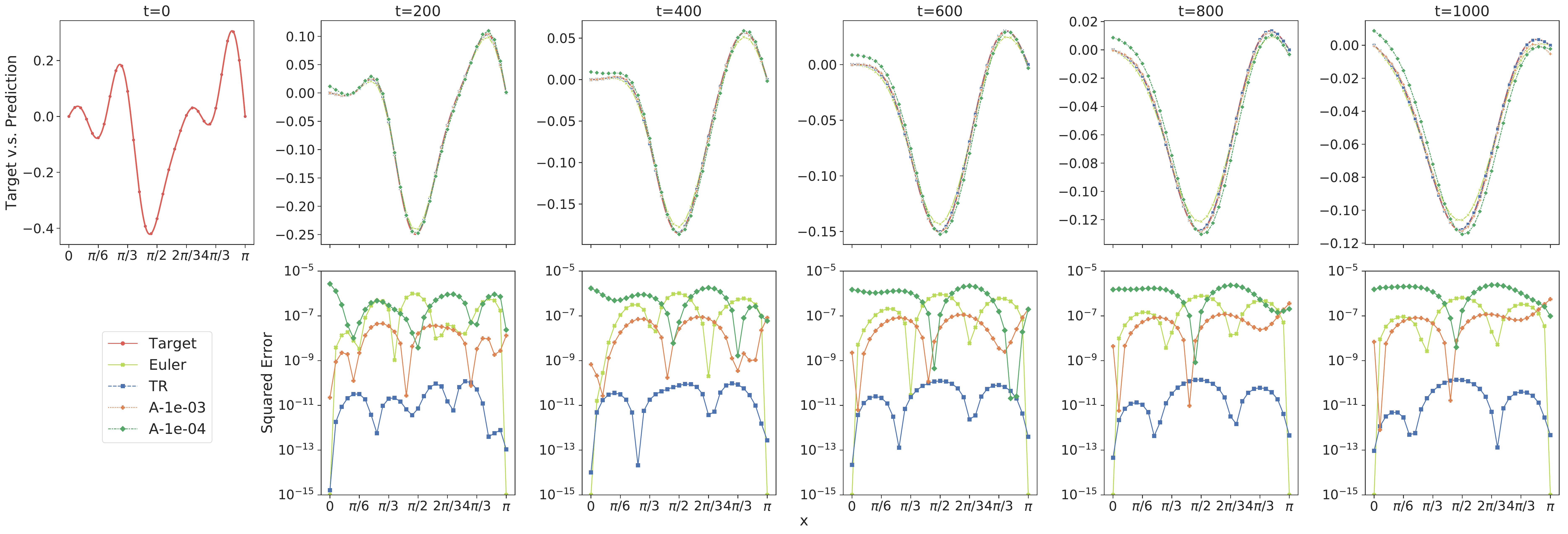}
        \caption{Low Noise}
    \end{subfigure}\\[3ex]
    \vspace{0.5in}
    \centering
    \begin{subfigure}{0.9\textwidth}
        \includegraphics[width=0.98\textwidth]{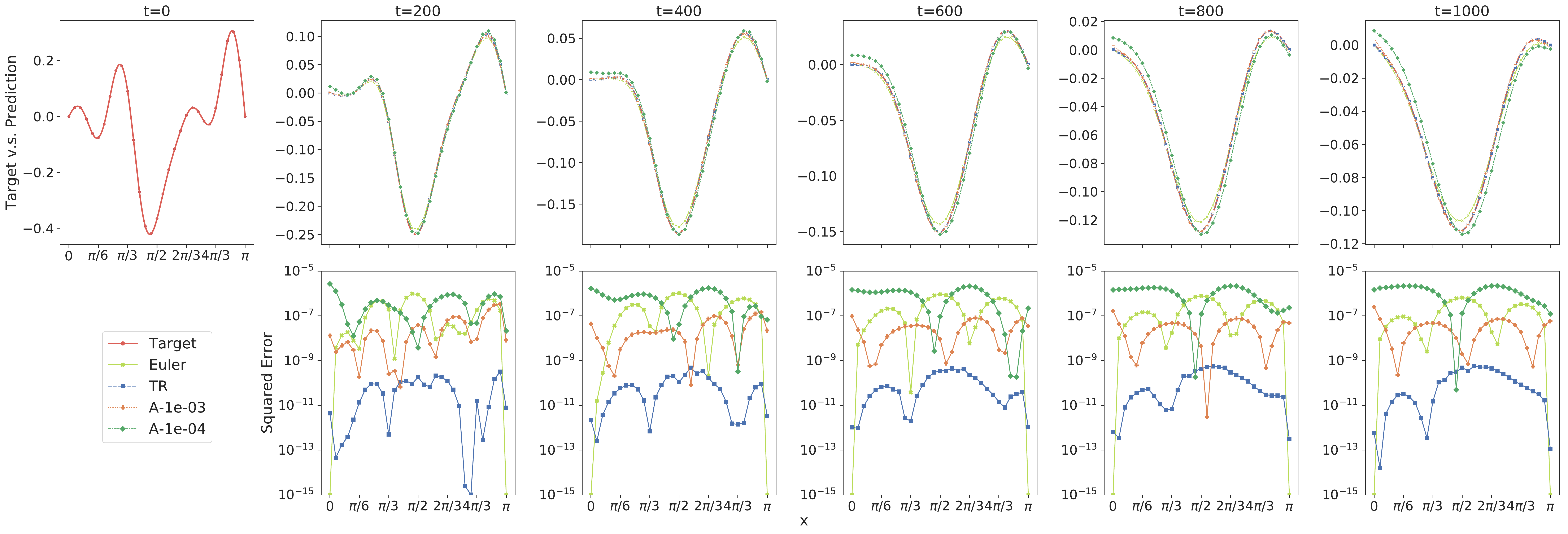}
        \caption{Medium Noise}
    \end{subfigure}\\[3ex]
    \vspace{0.5in}
    \centering
    \begin{subfigure}{0.9\textwidth}
        \includegraphics[width=0.98\textwidth]{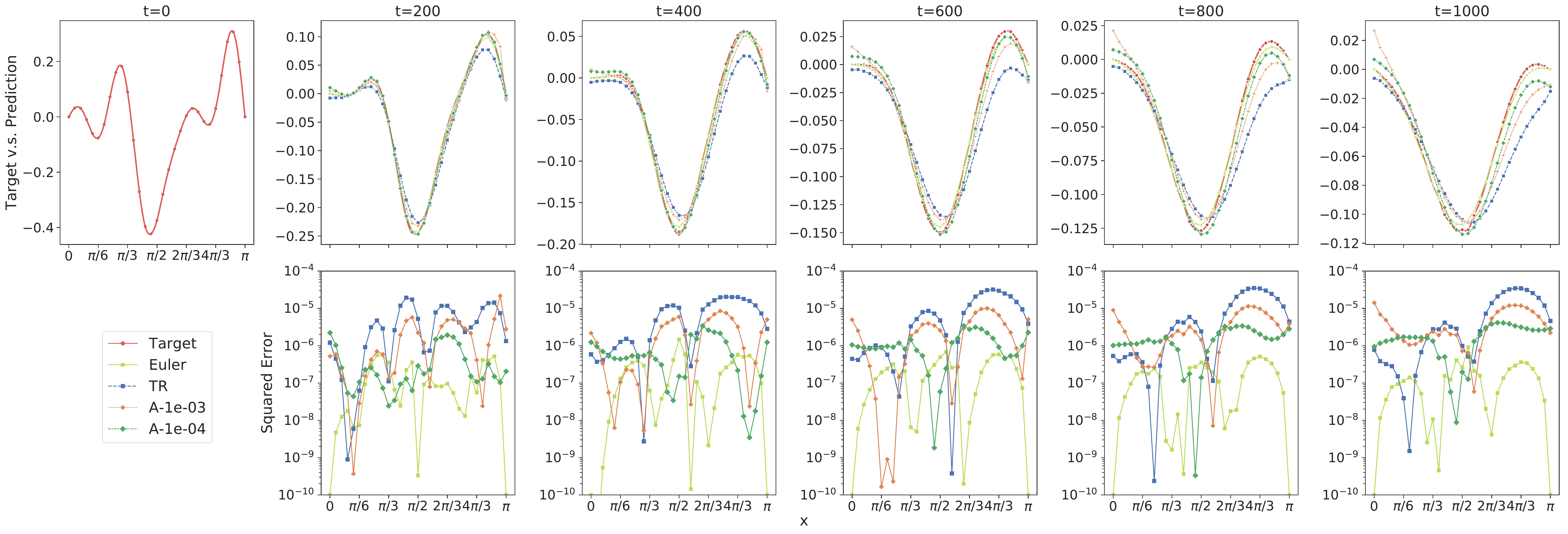}
        \caption{High Noise}
    \end{subfigure}
  \caption{\textbf{Noisy Case:} Sequence of predictions by level of noise.}
  \label{fig.noisy_u}
\end{figure}

\begin{figure}
  \centering
    \begin{subfigure}{0.48\textwidth}
        \includegraphics[width=0.98\textwidth]{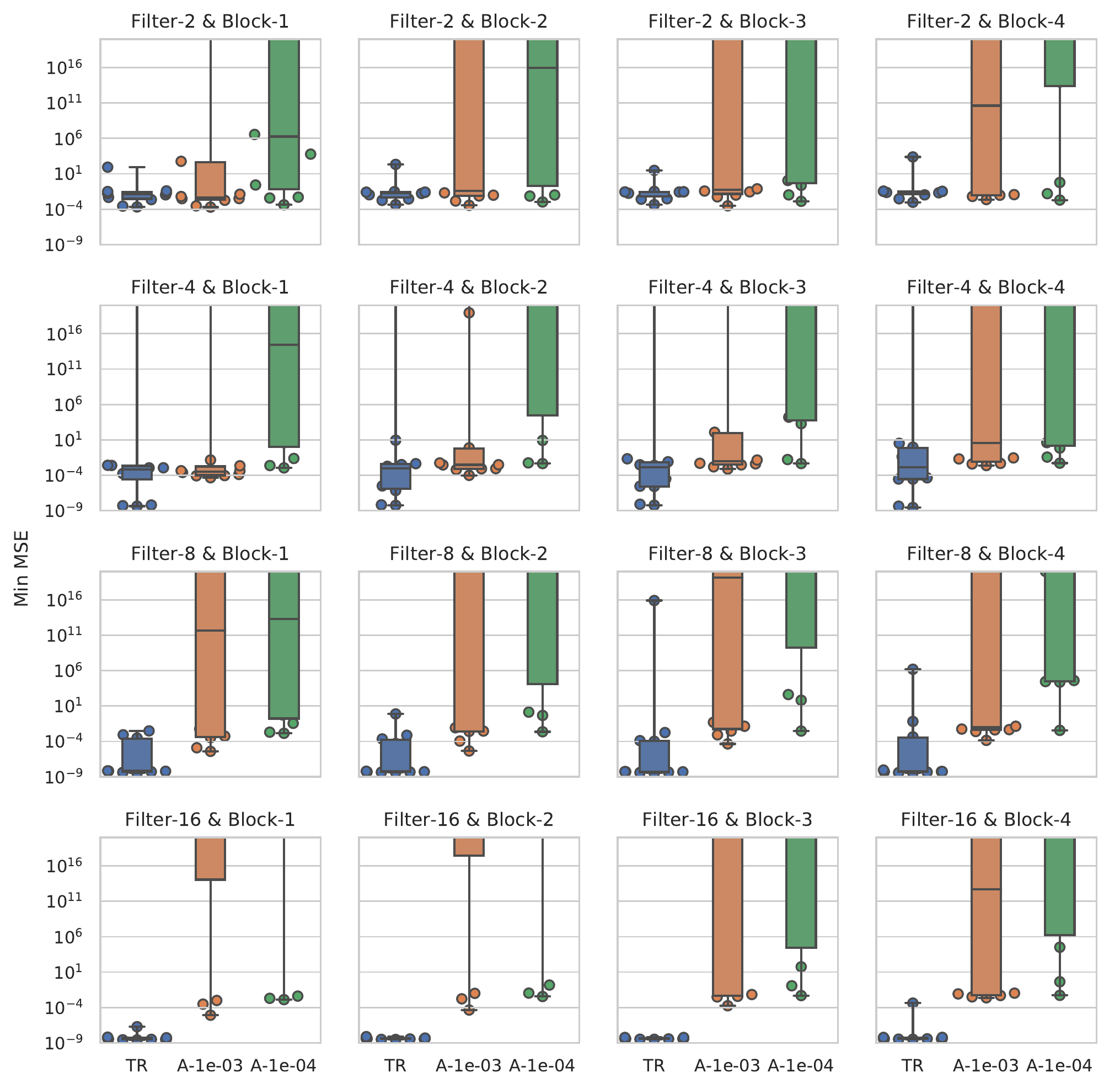}
        \caption{Low Noise}
    \end{subfigure}
    \centering
    \begin{subfigure}{0.48\textwidth}
        \includegraphics[width=0.98\textwidth]{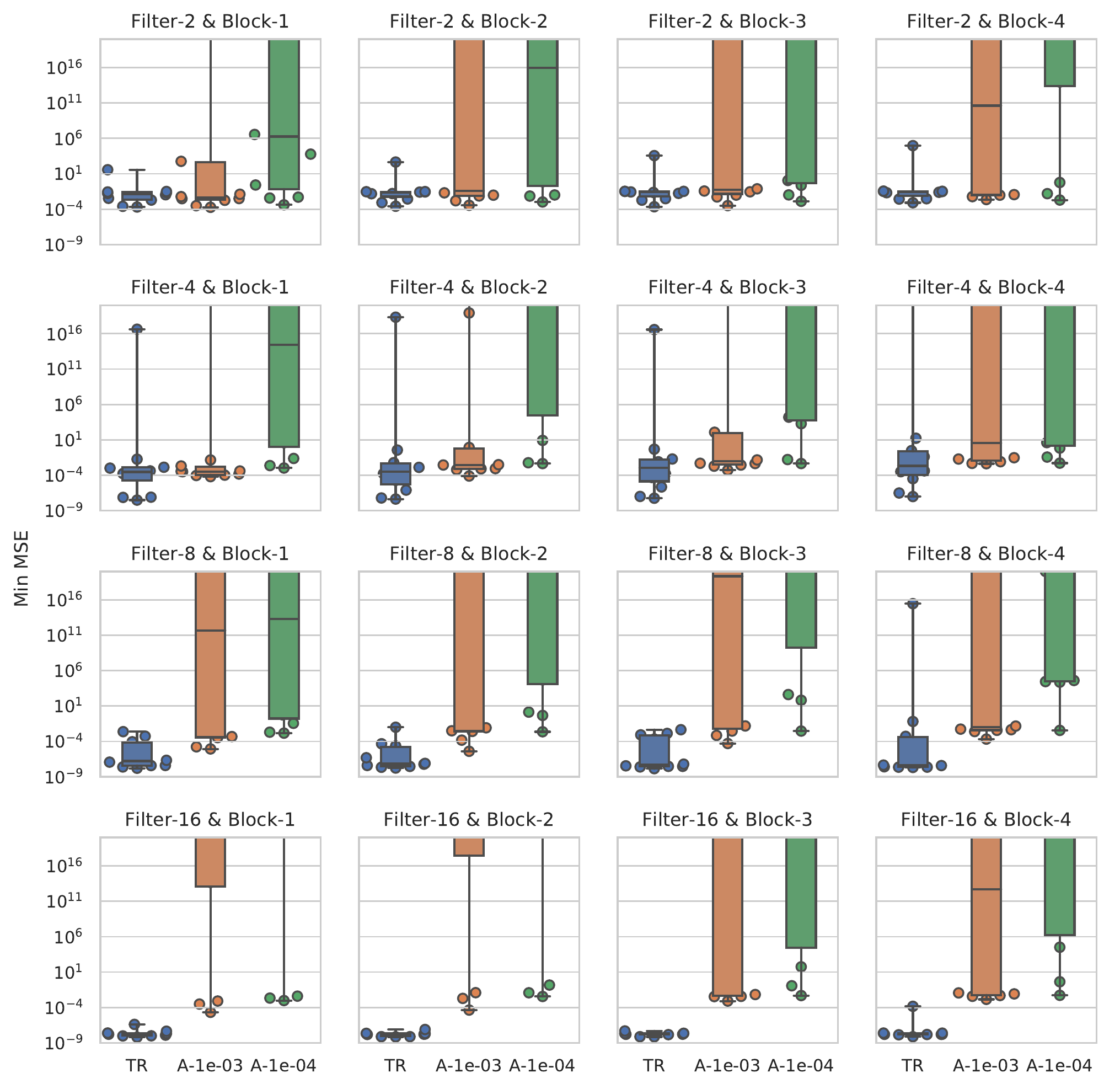}
        \caption{Medium Noise}
    \end{subfigure}\\[3ex]
    \centering
    \begin{subfigure}{0.48\textwidth}
        \includegraphics[width=0.98\textwidth]{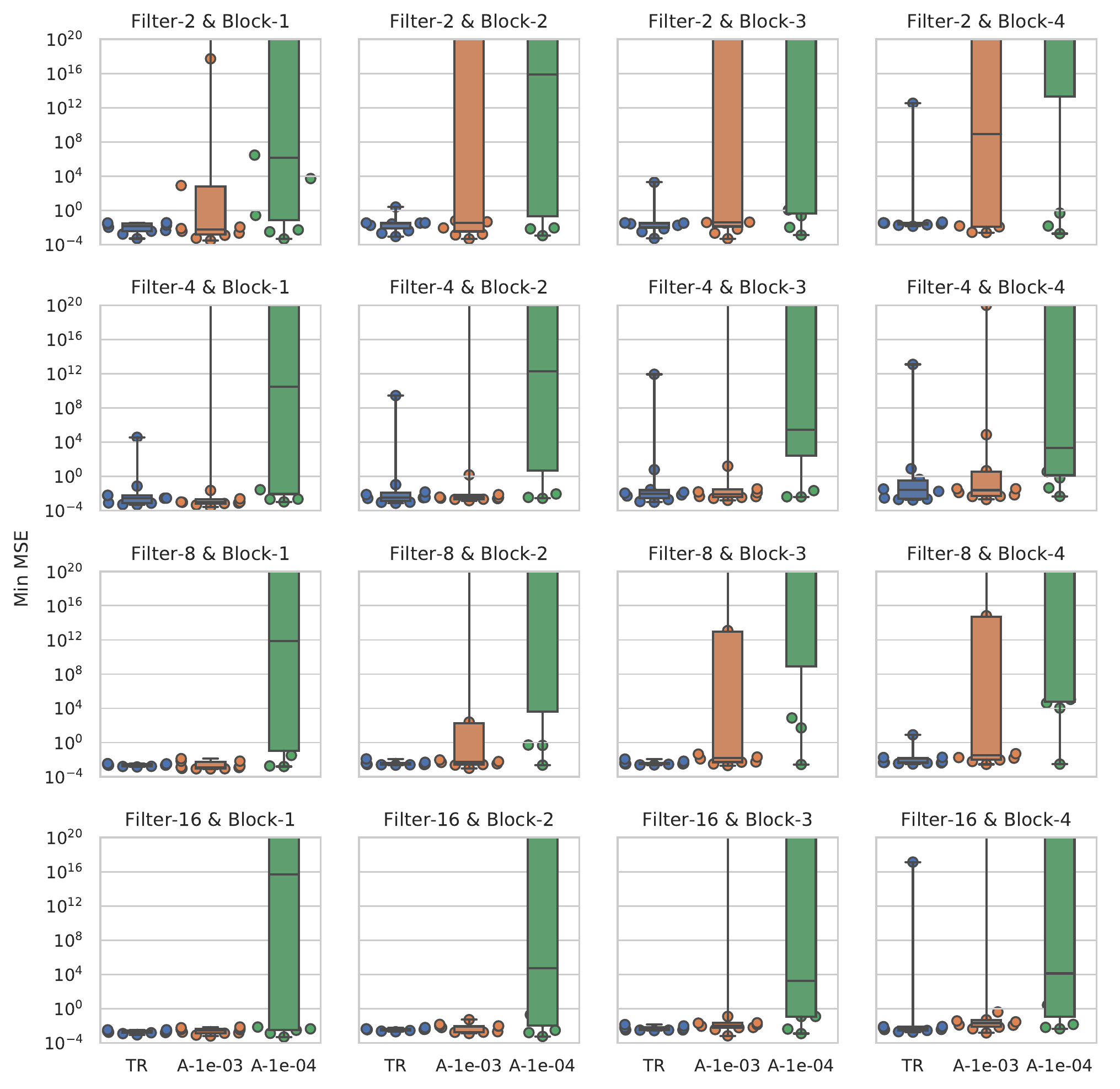}
        \caption{High Noise}
    \end{subfigure}
  \caption{\textbf{Noisy Case:} Minimum testing errors by level of noise and configuration - 1000-step prediction.}
  \label{fig.noisy_1000}
\end{figure}

\begin{figure}
  \centering
    \begin{subfigure}{0.48\textwidth}
        \includegraphics[width=0.98\textwidth]{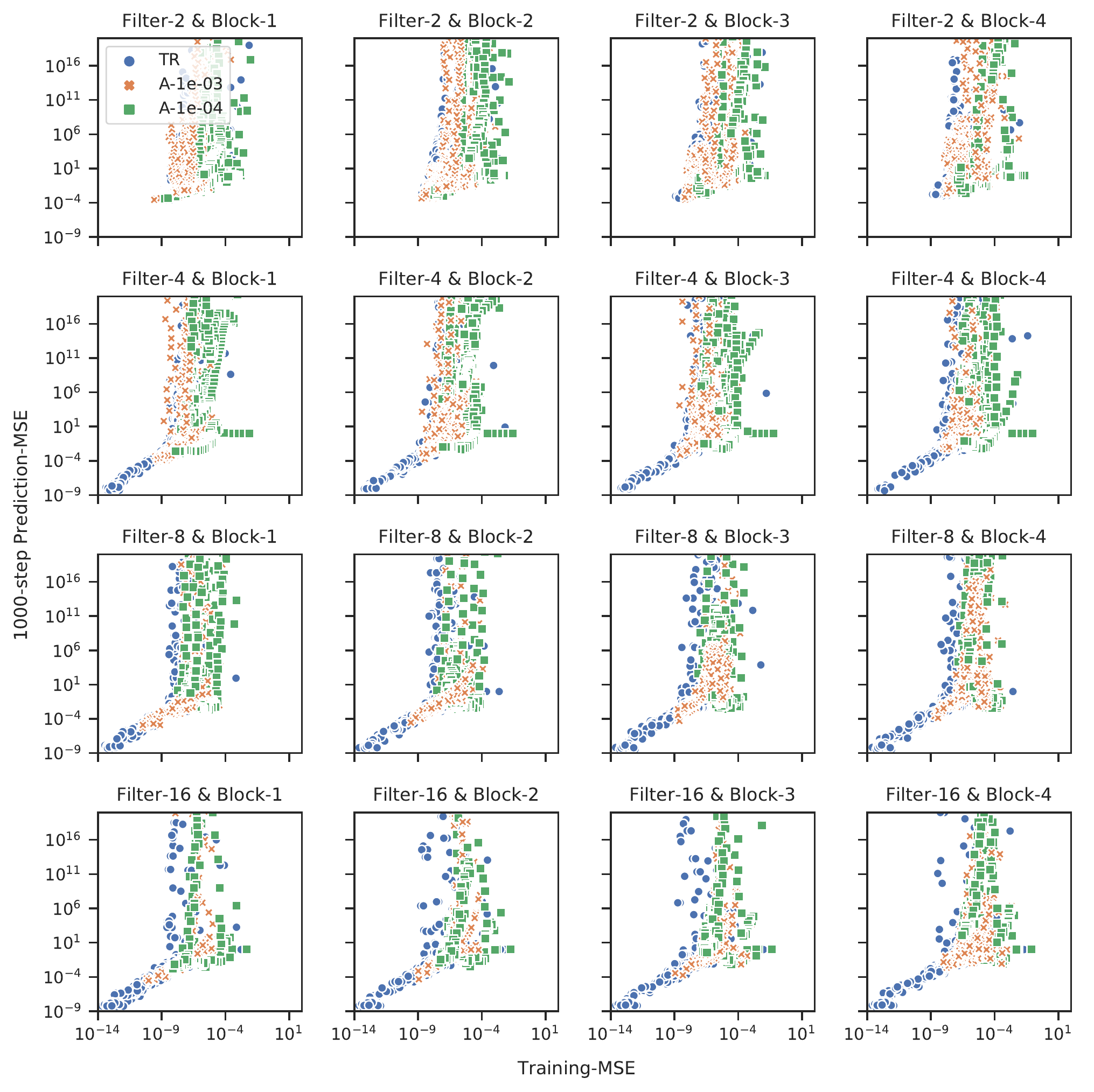}
        \caption{Low Noise}
    \end{subfigure}
    \centering
    \begin{subfigure}{0.48\textwidth}
        \includegraphics[width=0.98\textwidth]{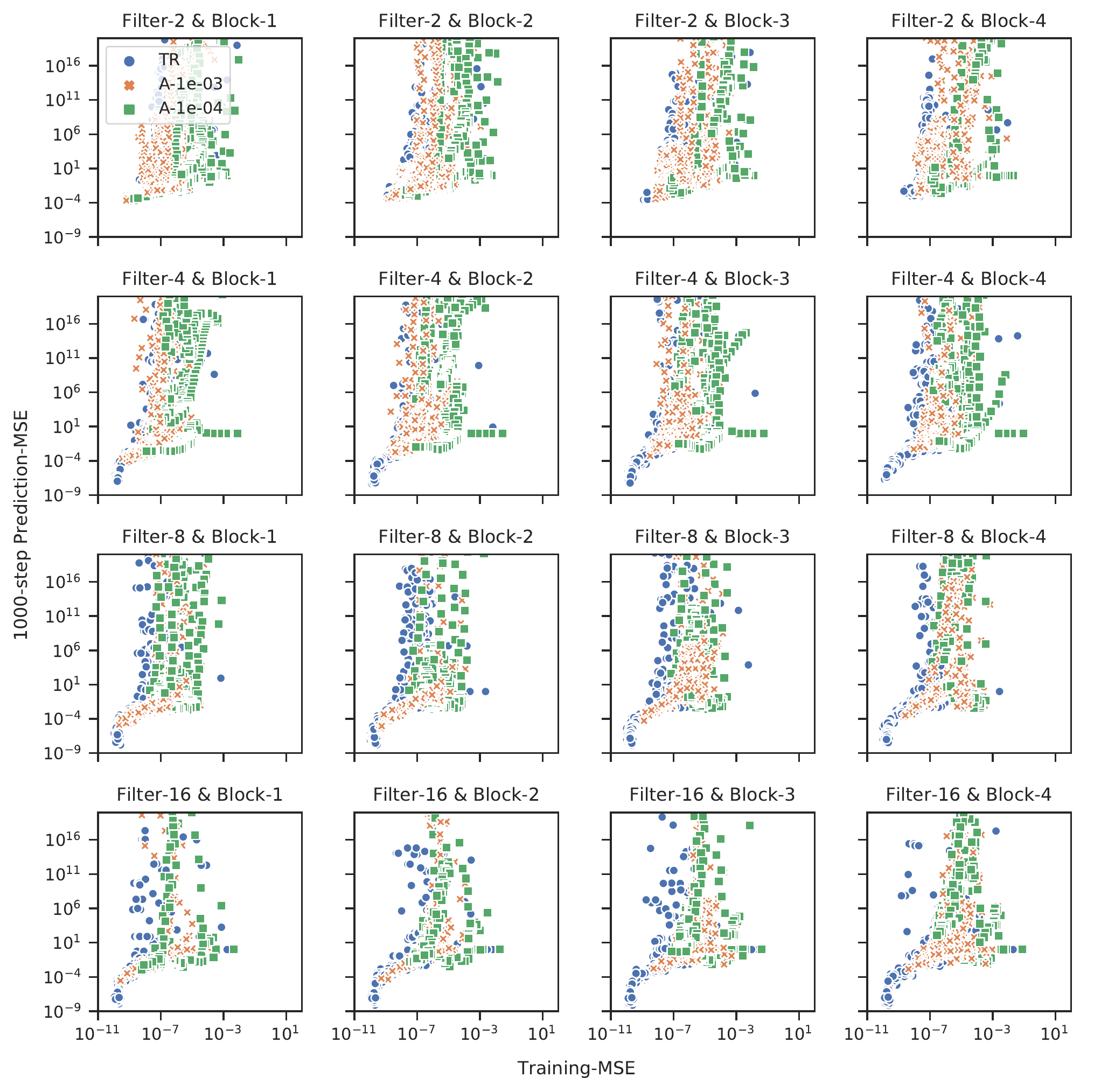} 
        \caption{Medium Noise}
    \end{subfigure}\\[3ex]
    \centering
    \begin{subfigure}{0.48\textwidth}
        \includegraphics[width=0.98\textwidth]{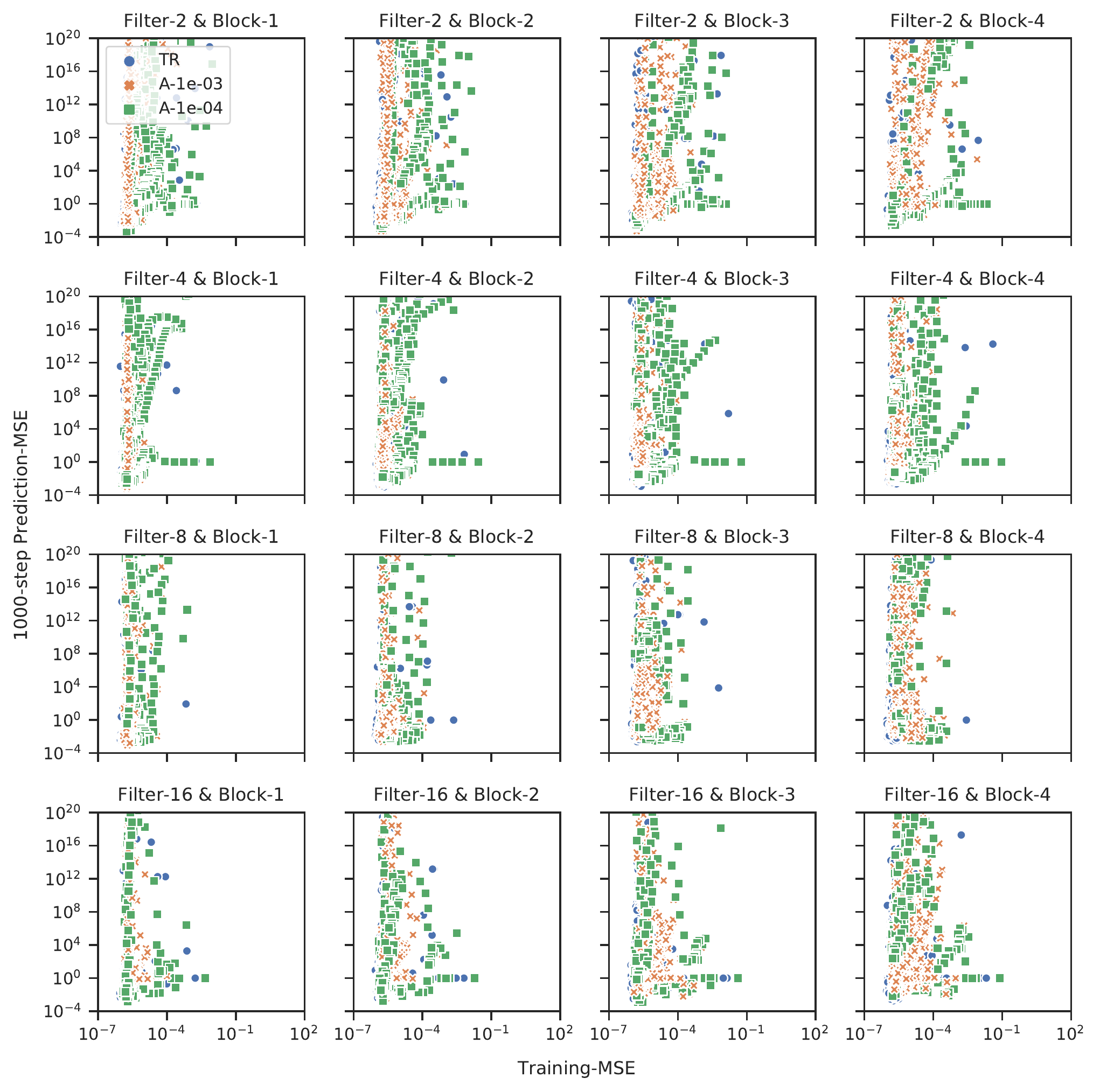}
        \caption{High Noise}
    \end{subfigure}
  \caption{\textbf{Noisy Case:} Training vs. 1000-step prediction errors (over the training process) by level of noise and configuration.}
  \label{fig.noisy_scatter}
\end{figure}

\begin{figure}
  \centering
    \begin{subfigure}{0.48\textwidth}
        \includegraphics[width=0.98\textwidth]{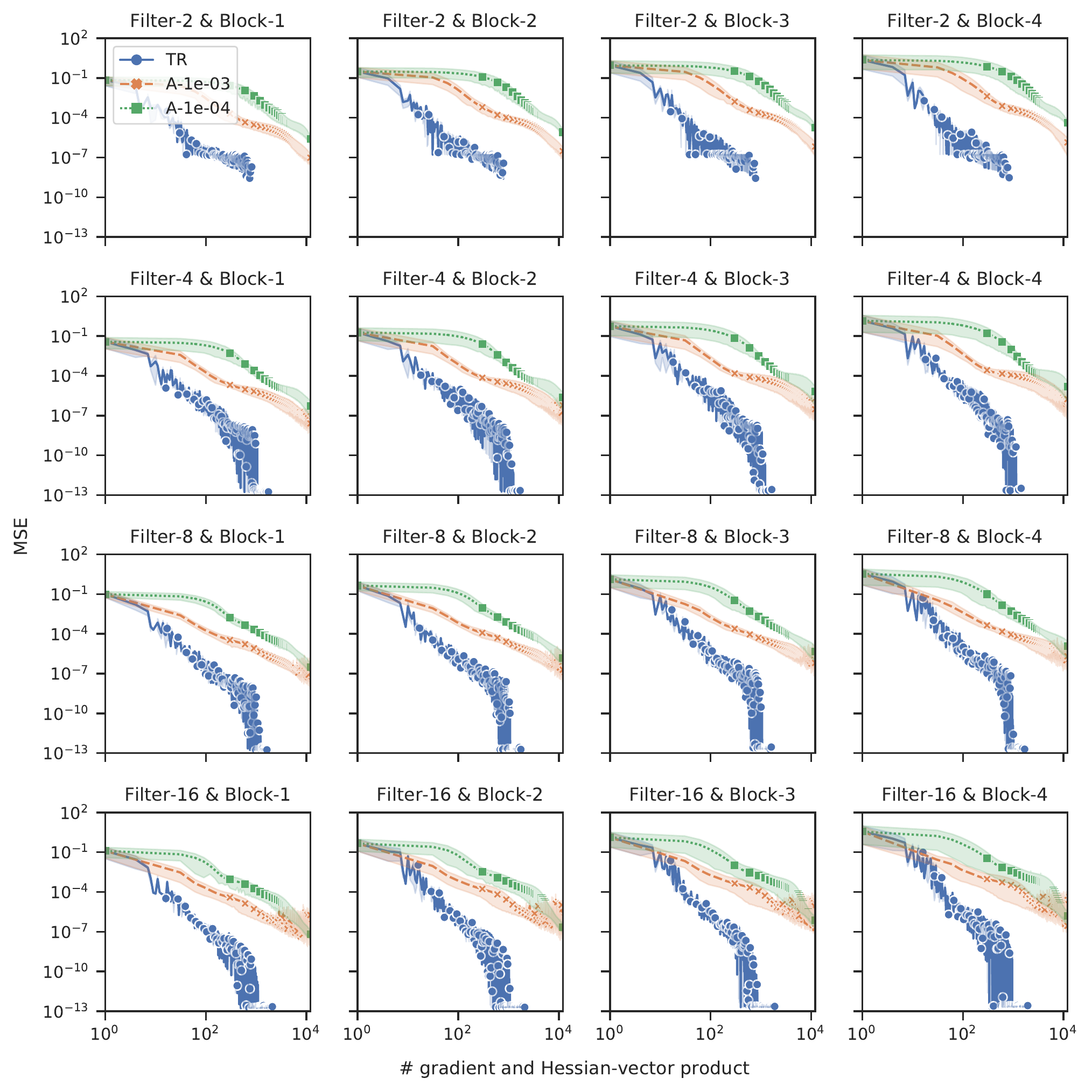}
        \caption{Low Noise}
    \end{subfigure}
    \centering
    \begin{subfigure}{0.48\textwidth}
        \includegraphics[width=0.98\textwidth]{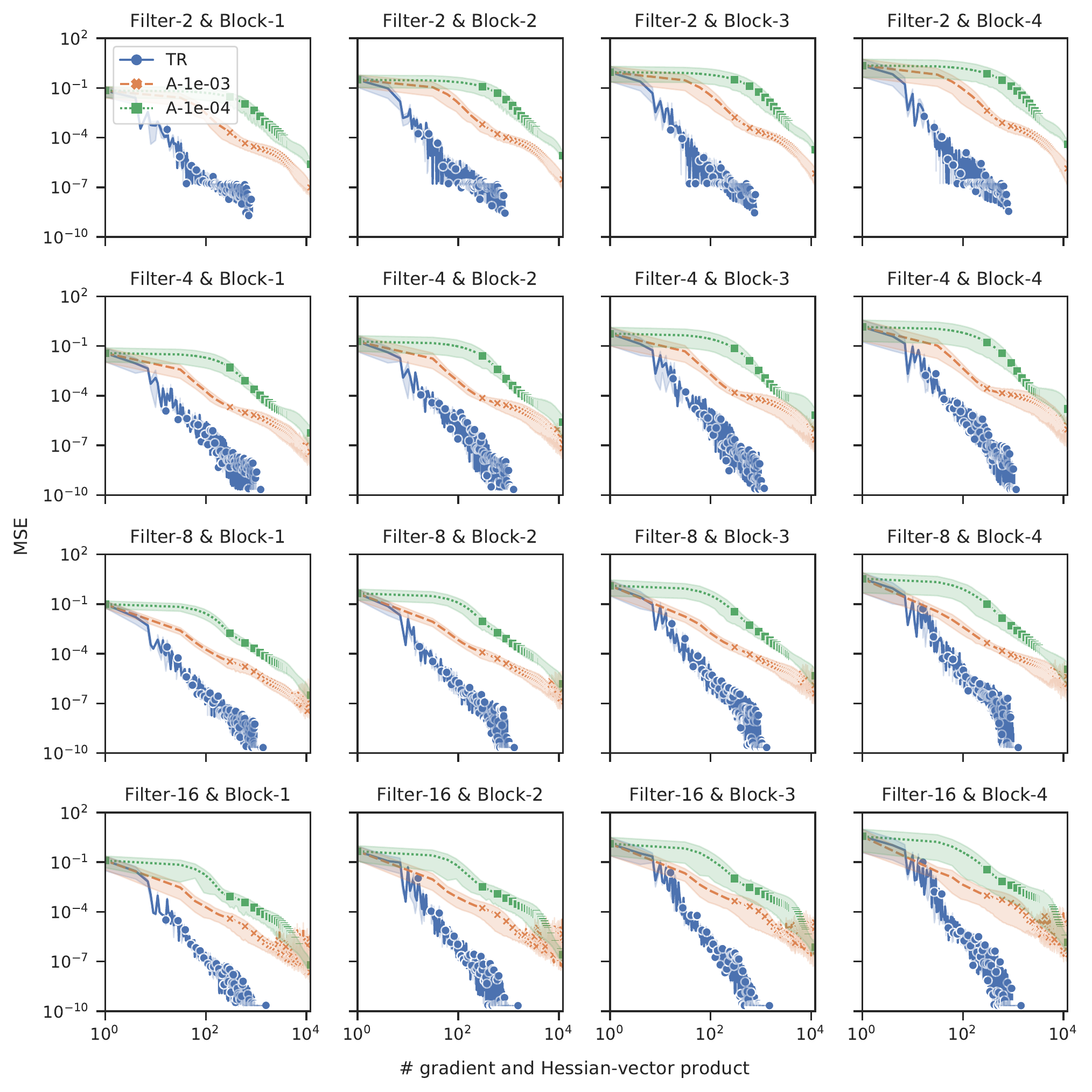}
        \caption{Medium Noise}
    \end{subfigure}\\[3ex]
    \centering
    \begin{subfigure}{0.48\textwidth}
        \includegraphics[width=0.98\textwidth]{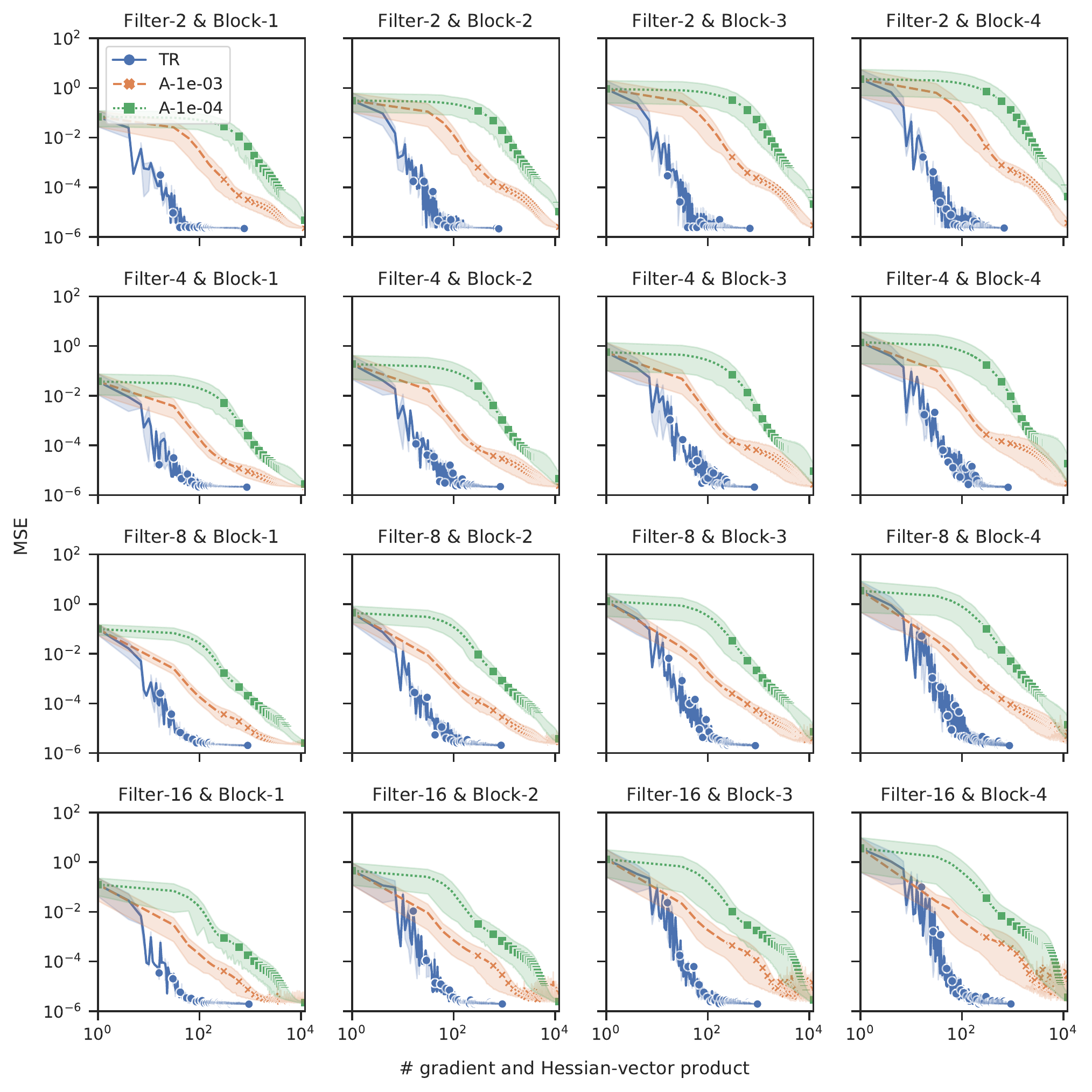}
        \caption{High Noise}
    \end{subfigure}
  \caption{\textbf{Noisy Case:} Evolution of the testing errors by level of noise and configuration - one-step prediction.}
  \label{fig.noisy_1_evol}
\end{figure}

\begin{figure}
  \centering
    \begin{subfigure}{0.48\textwidth}
        \includegraphics[width=0.98\textwidth]{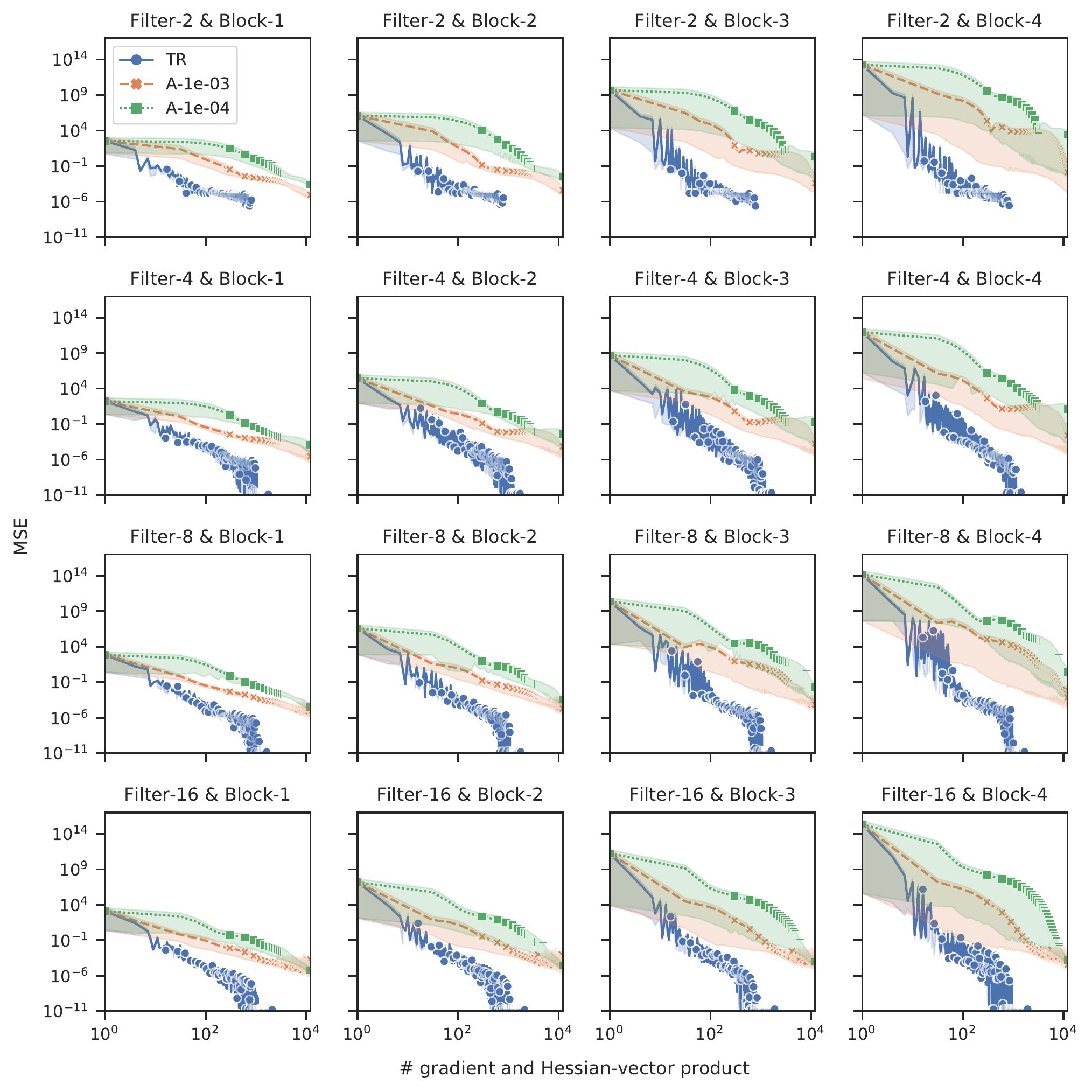}
        \caption{Low Noise}
    \end{subfigure}
    \centering
    \begin{subfigure}{0.48\textwidth}
        \includegraphics[width=0.98\textwidth]{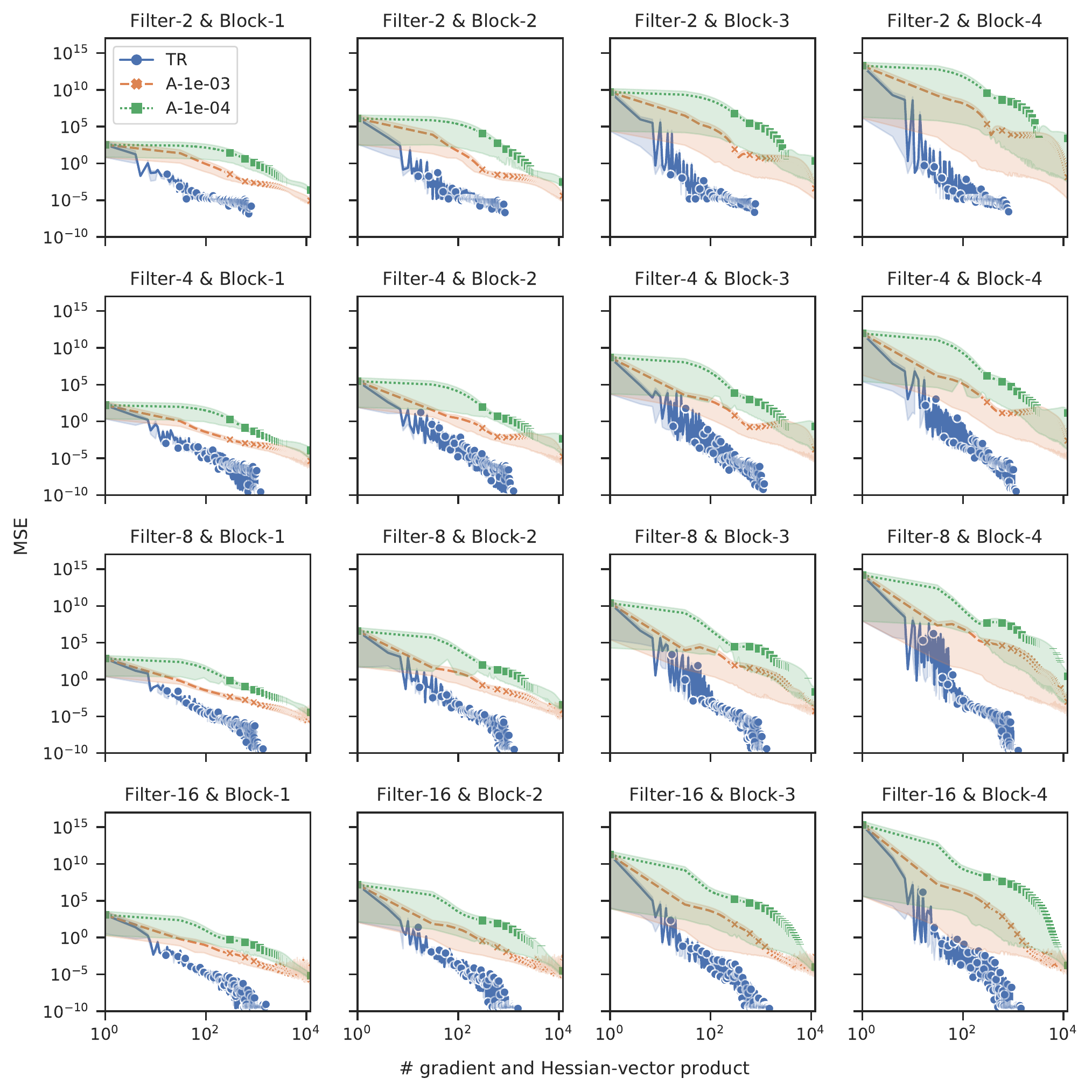}
        \caption{Medium Noise}
    \end{subfigure}\\[3ex]
    \centering
    \begin{subfigure}{0.48\textwidth}
        \includegraphics[width=0.98\textwidth]{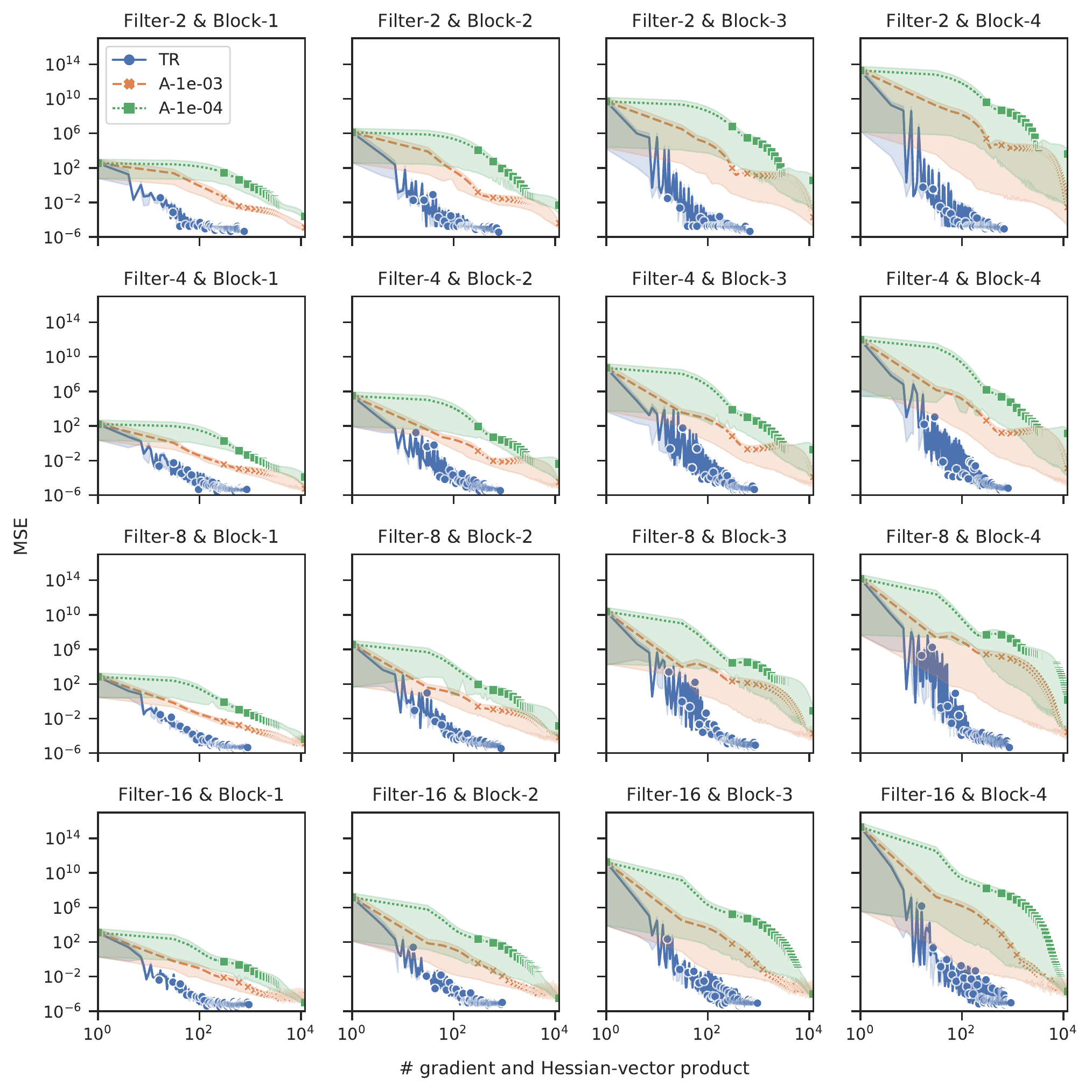}
        \caption{High Noise}
    \end{subfigure}
  \caption{\textbf{Noisy Case:} Evolution of the testing errors by level of noise and configuration - multi-step prediction.}
  \label{fig.noisy_10_evol}
\end{figure}

\begin{figure}
  \centering
    \begin{subfigure}{0.48\textwidth}
        \includegraphics[width=0.98\textwidth]{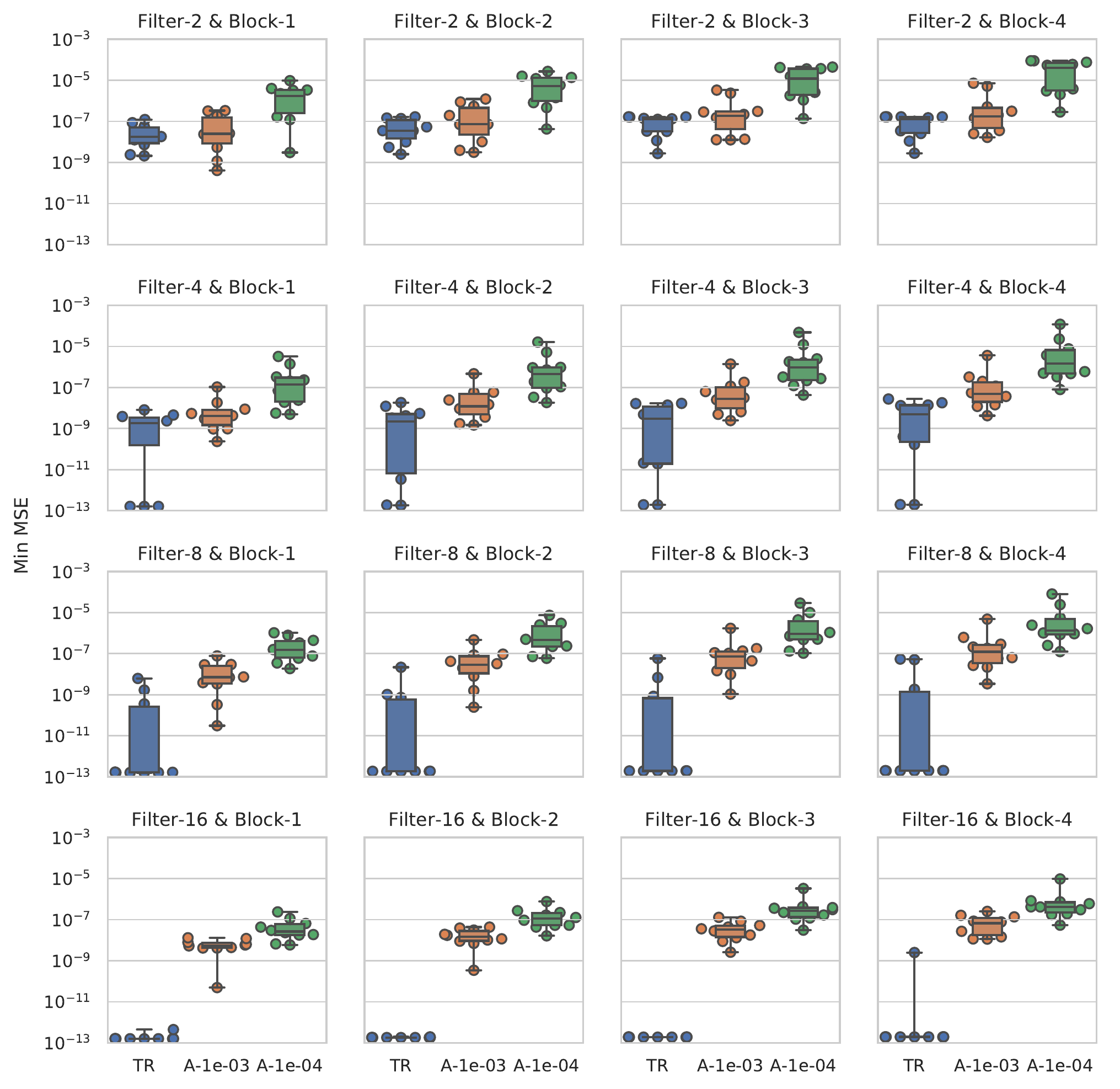}
        \caption{Low Noise}
    \end{subfigure}
    \centering
    \begin{subfigure}{0.48\textwidth}
        \includegraphics[width=0.98\textwidth]{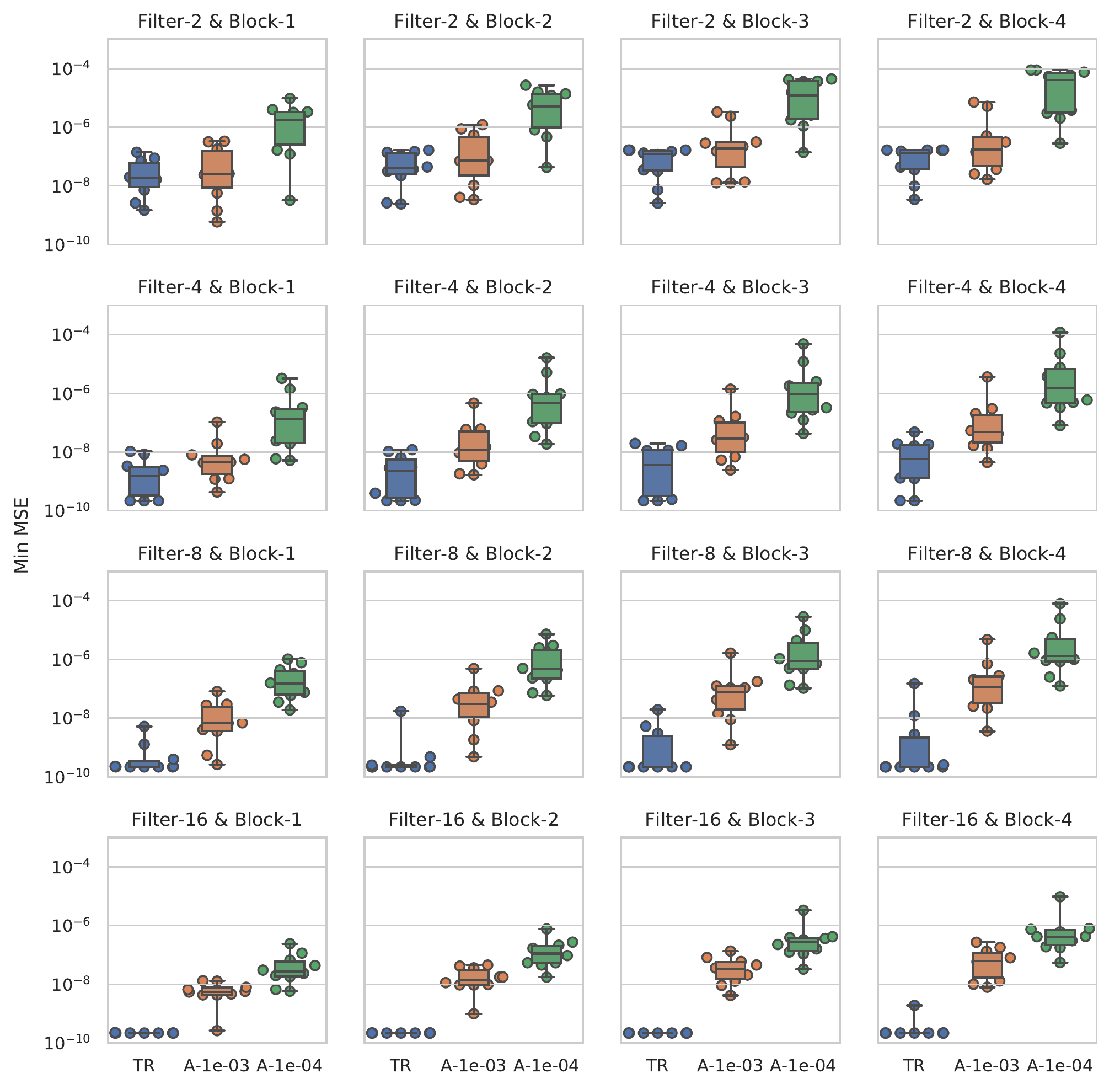}
        \caption{Medium Noise}
    \end{subfigure}\\[3ex]
    \centering
    \begin{subfigure}{0.48\textwidth}
        \includegraphics[width=0.98\textwidth]{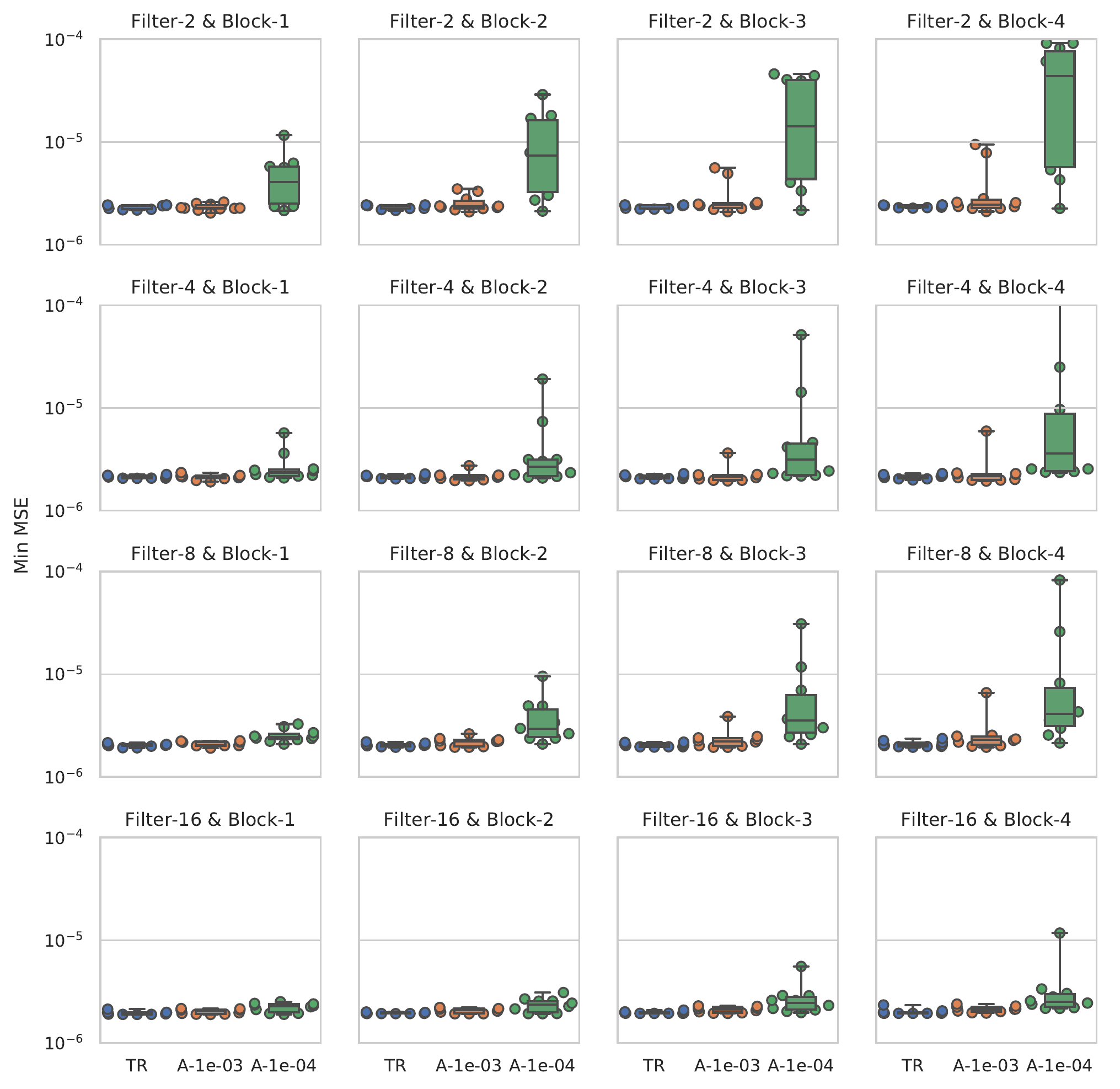}
        \caption{High Noise}
    \end{subfigure}
  \caption{\textbf{Noisy Case:} Minimum testing errors by level of noise and configuration - one-step prediction.}
  \label{fig.noisy_1}
\end{figure}

\begin{figure}
  \centering
    \begin{subfigure}{0.48\textwidth}
        \includegraphics[width=0.98\textwidth]{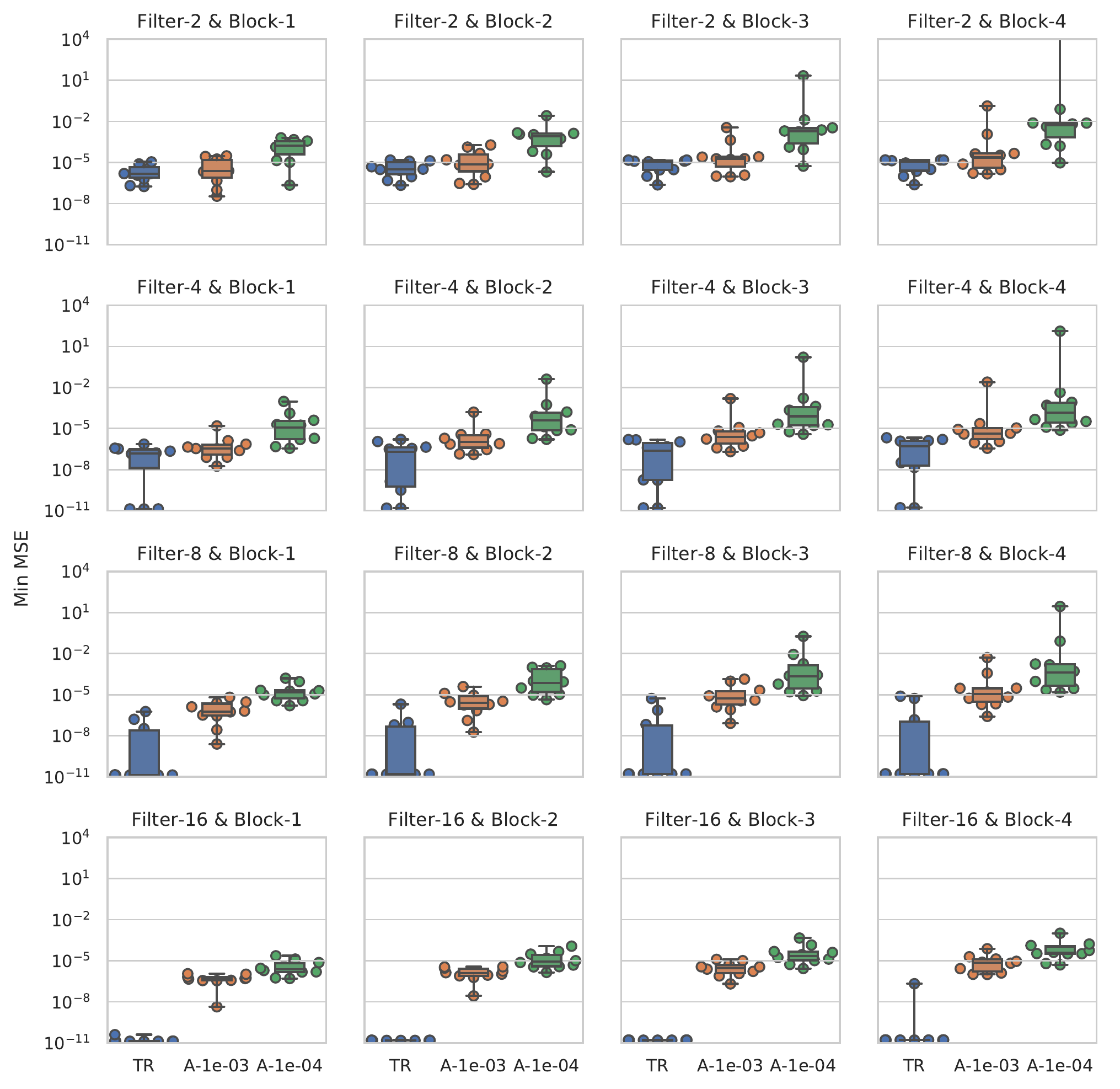}
        \caption{Low Noise}
    \end{subfigure}
    \centering
    \begin{subfigure}{0.48\textwidth}
        \includegraphics[width=0.98\textwidth]{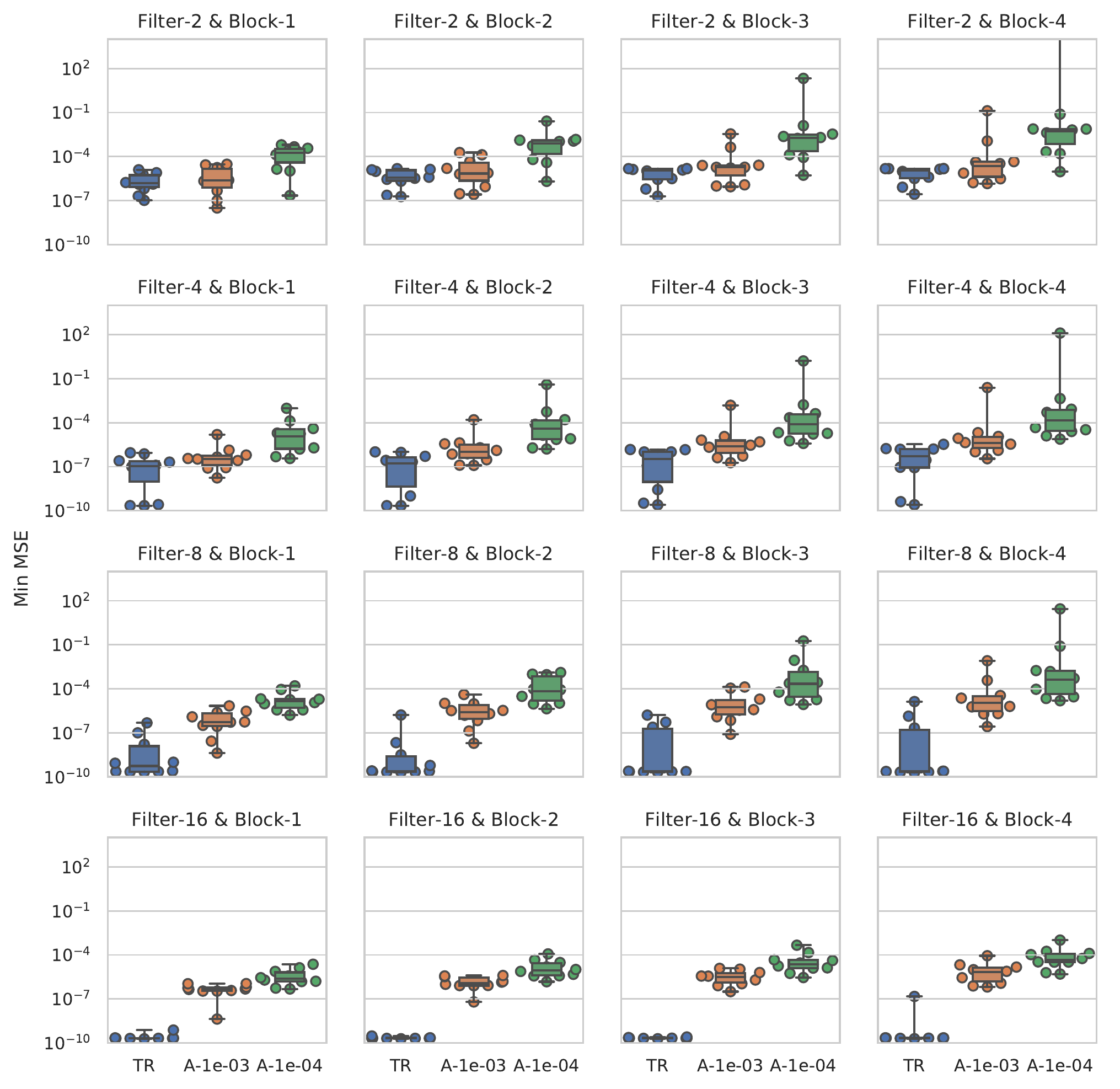}
        \caption{Medium Noise}
    \end{subfigure}\\[3ex]
    \centering
    \begin{subfigure}{0.48\textwidth}
        \includegraphics[width=0.98\textwidth]{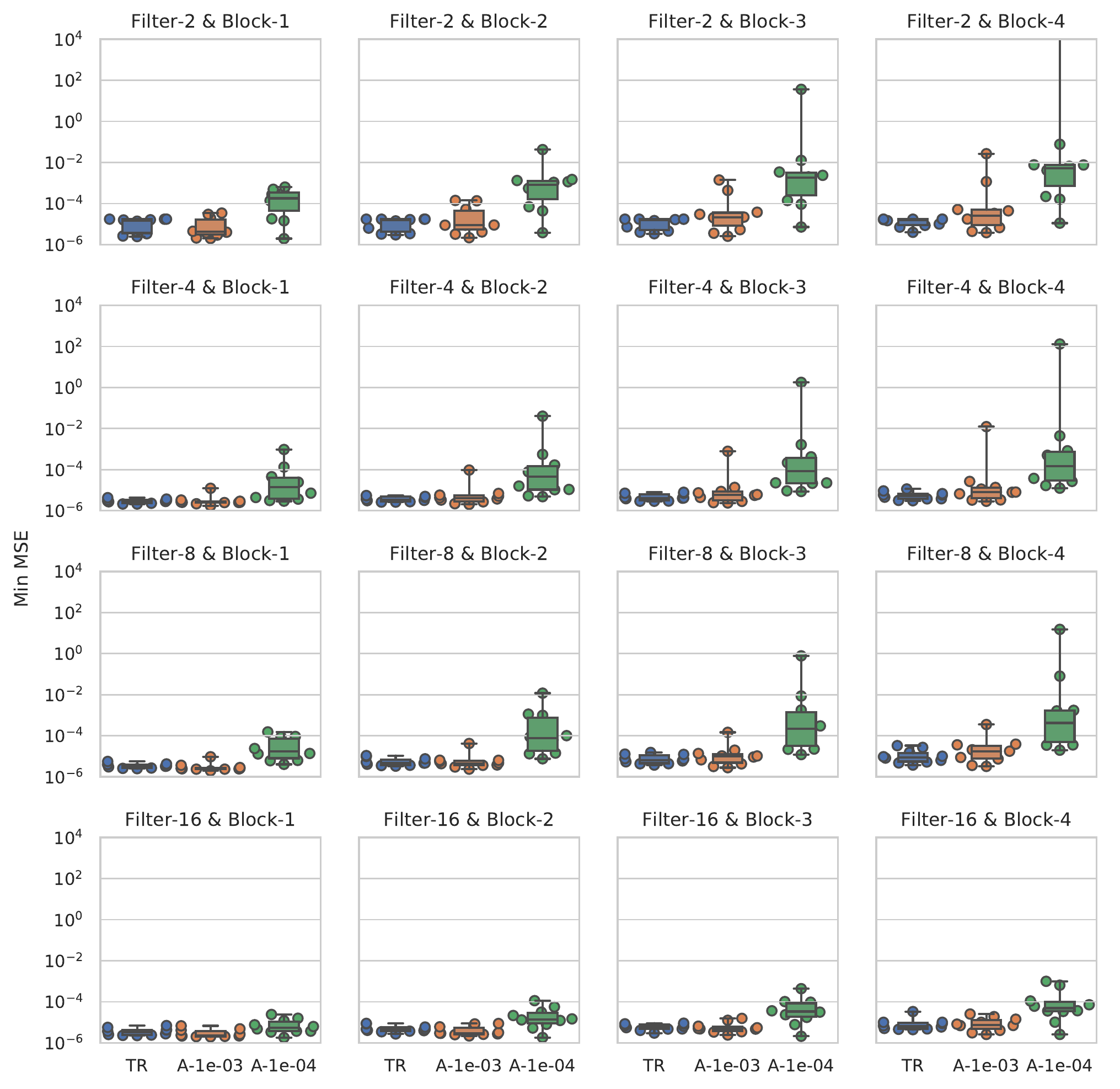}
        \caption{High Noise}
    \end{subfigure}
  \caption{\textbf{Noisy Case:} Minimum testing errors by level of noise and configuration - multi-step prediction.}
  \label{fig.noisy_10}
\end{figure}

\begin{figure}
  \centering
    \begin{subfigure}{1\textwidth}
    \centering
        \includegraphics[width=0.3\textwidth]{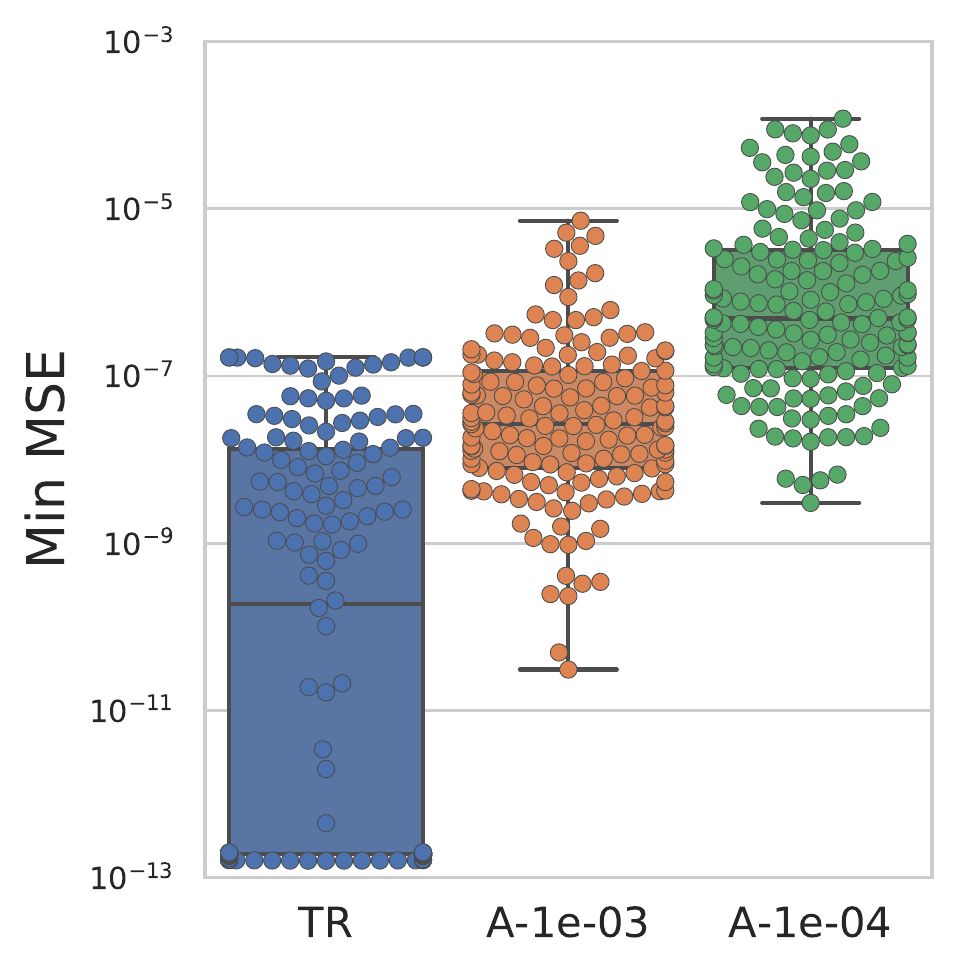}
        \includegraphics[width=0.3\textwidth]{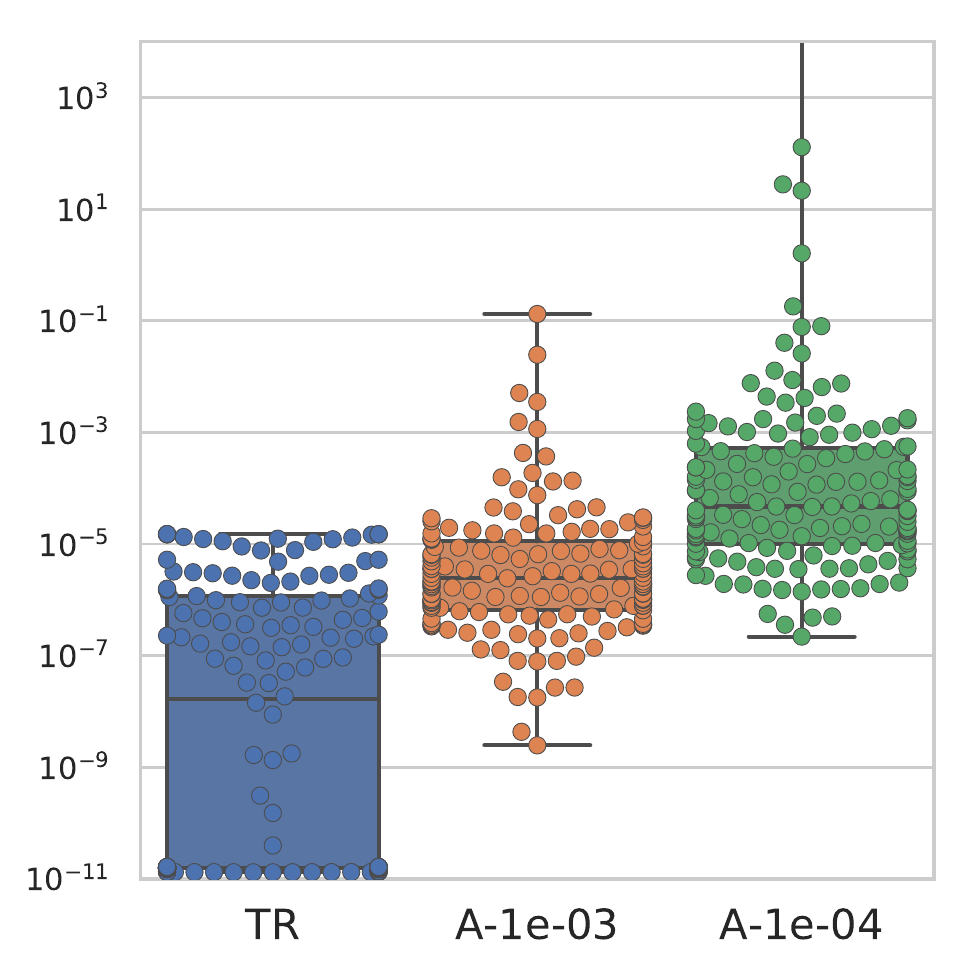}
        \includegraphics[width=0.3\textwidth]{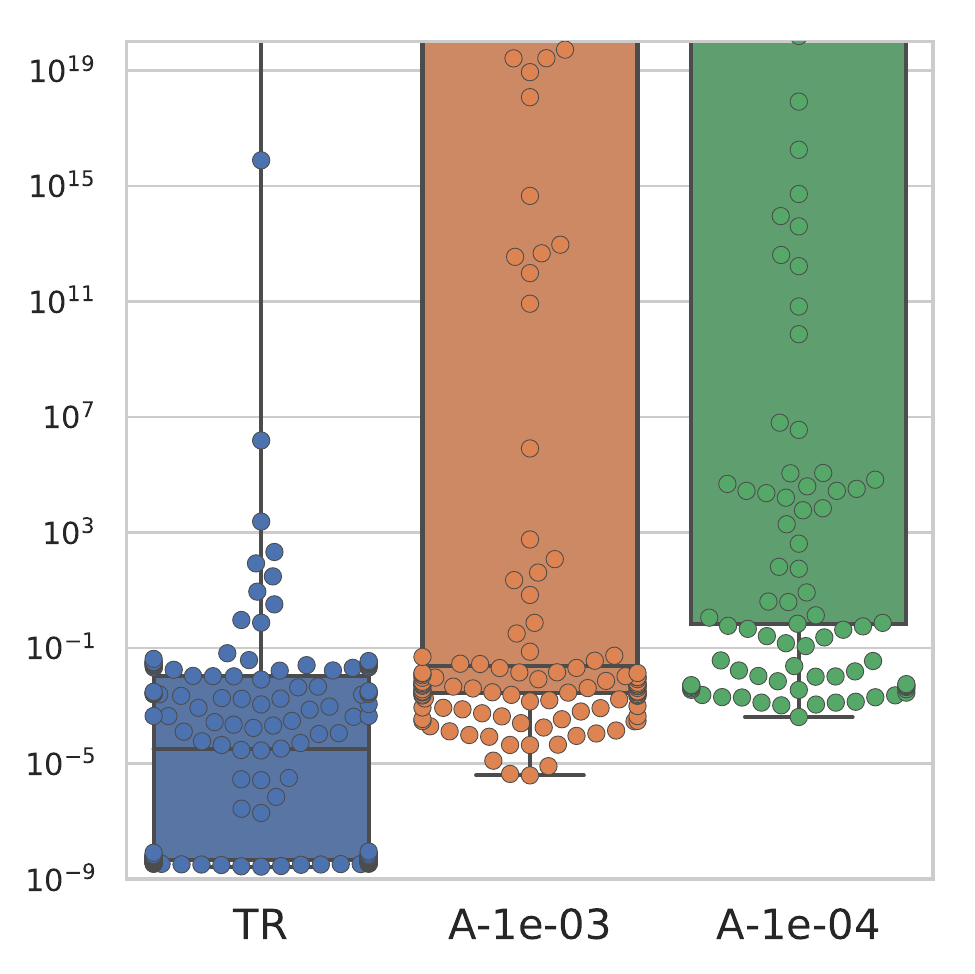}
        \caption{Low Noise}
    \end{subfigure}\\[3ex]
    \centering
    \begin{subfigure}{1\textwidth}
    \centering
        \includegraphics[width=0.3\textwidth]{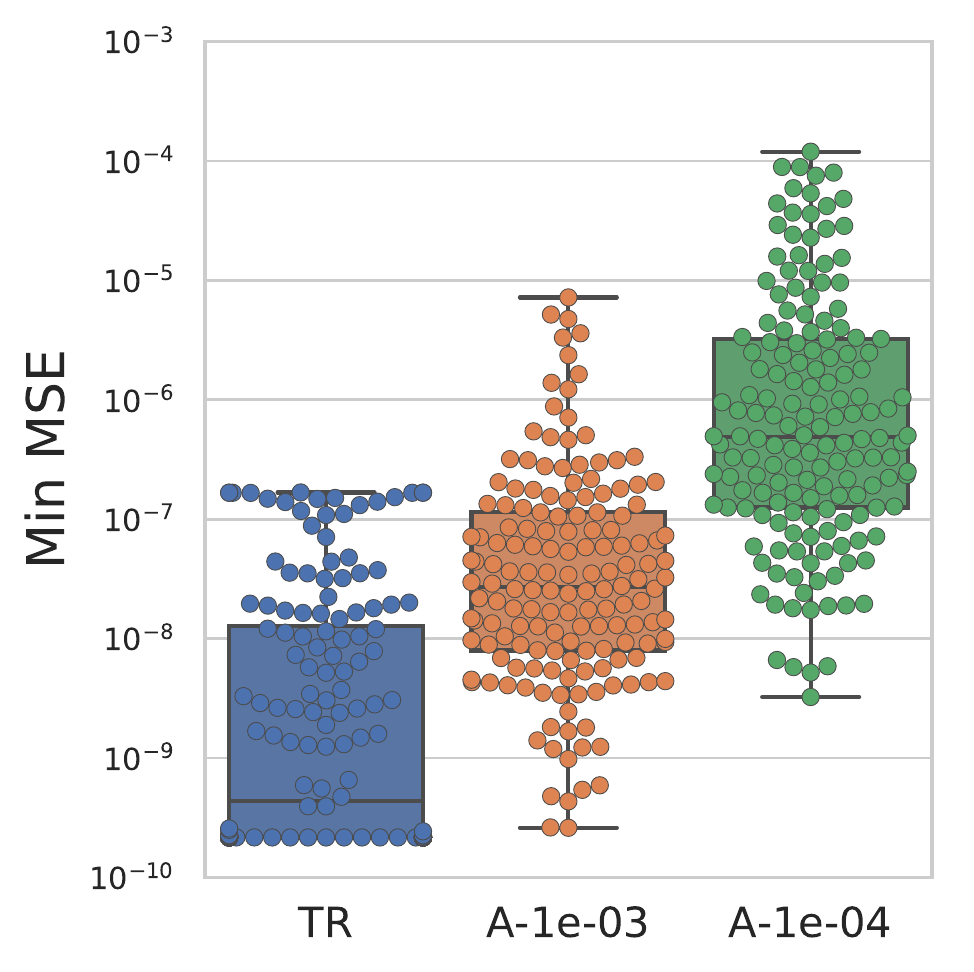}
        \includegraphics[width=0.3\textwidth]{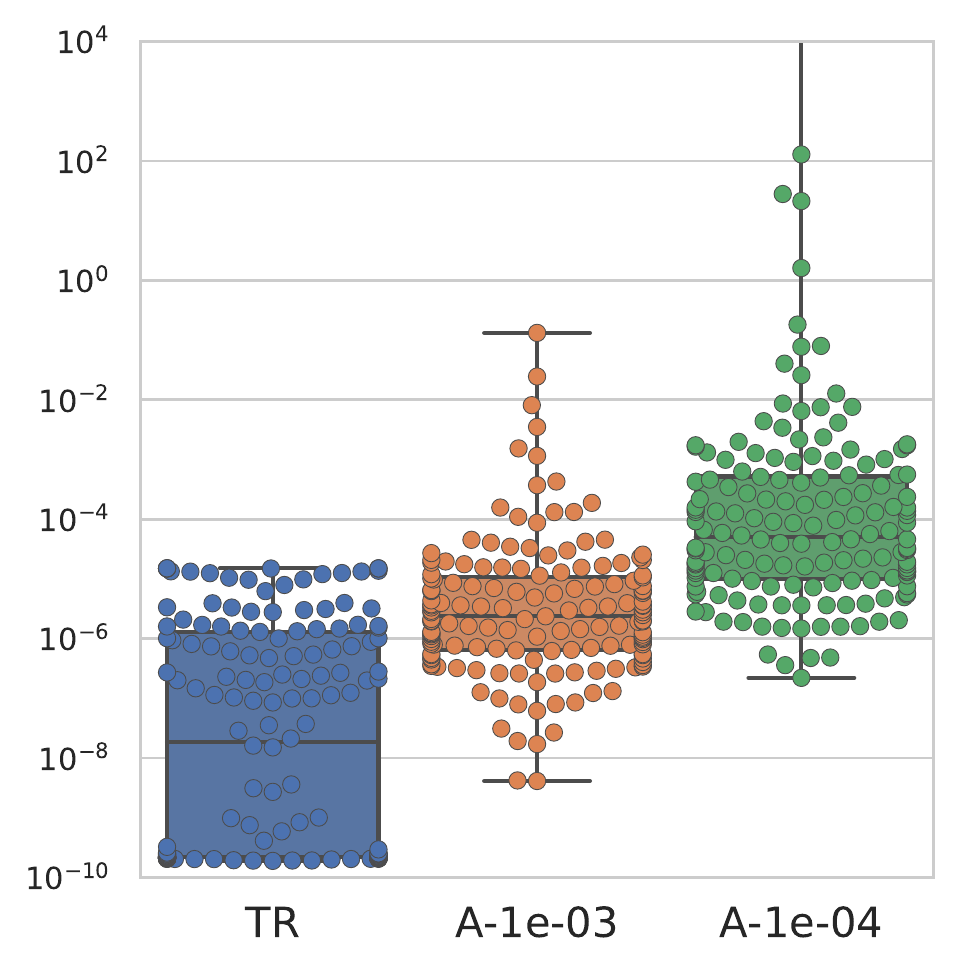}
        \includegraphics[width=0.3\textwidth]{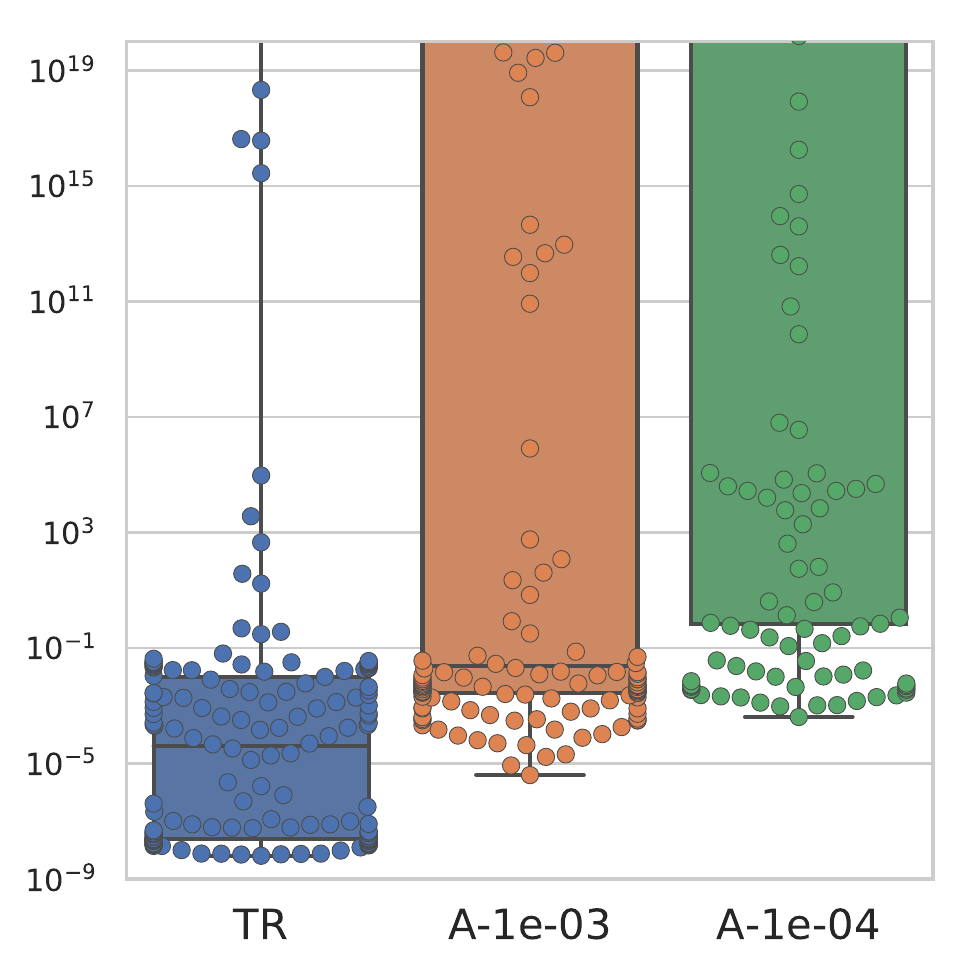}
        \caption{Medium Noise}
    \end{subfigure}\\[3ex]
    \centering
    \begin{subfigure}{1\textwidth}
    \centering
        \includegraphics[width=0.3\textwidth]{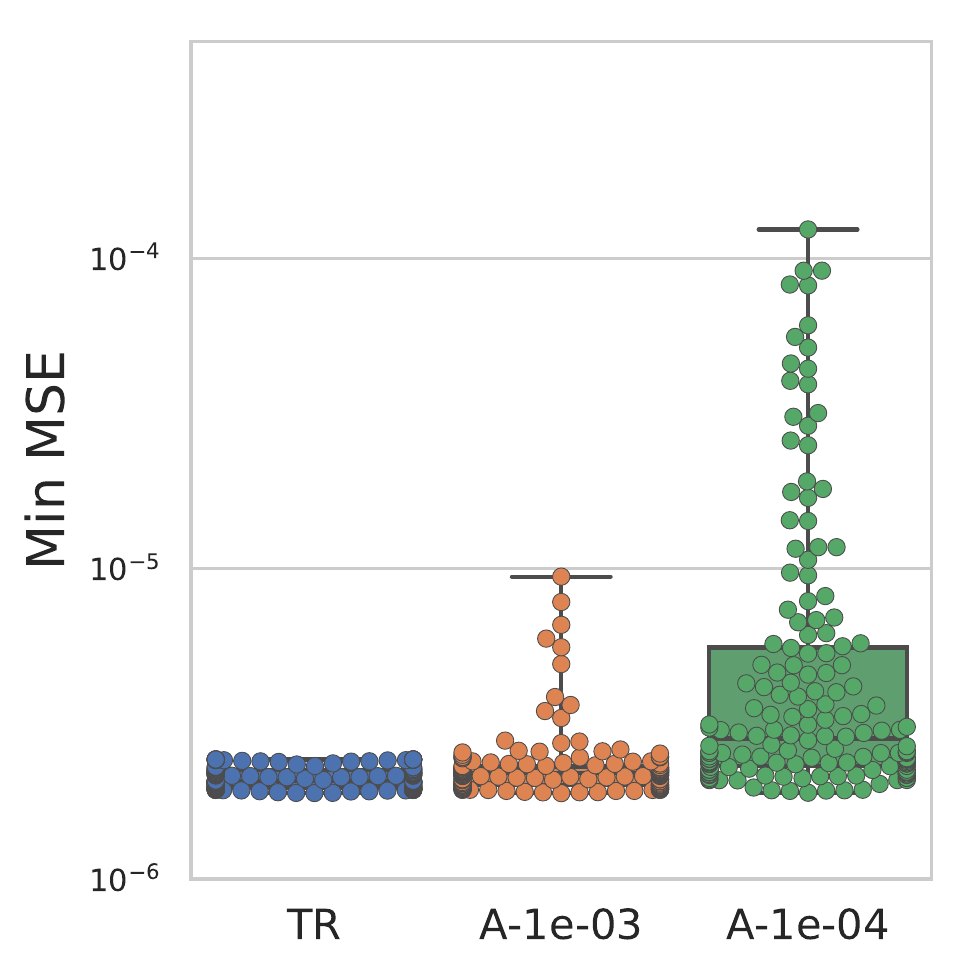}
        \includegraphics[width=0.3\textwidth]{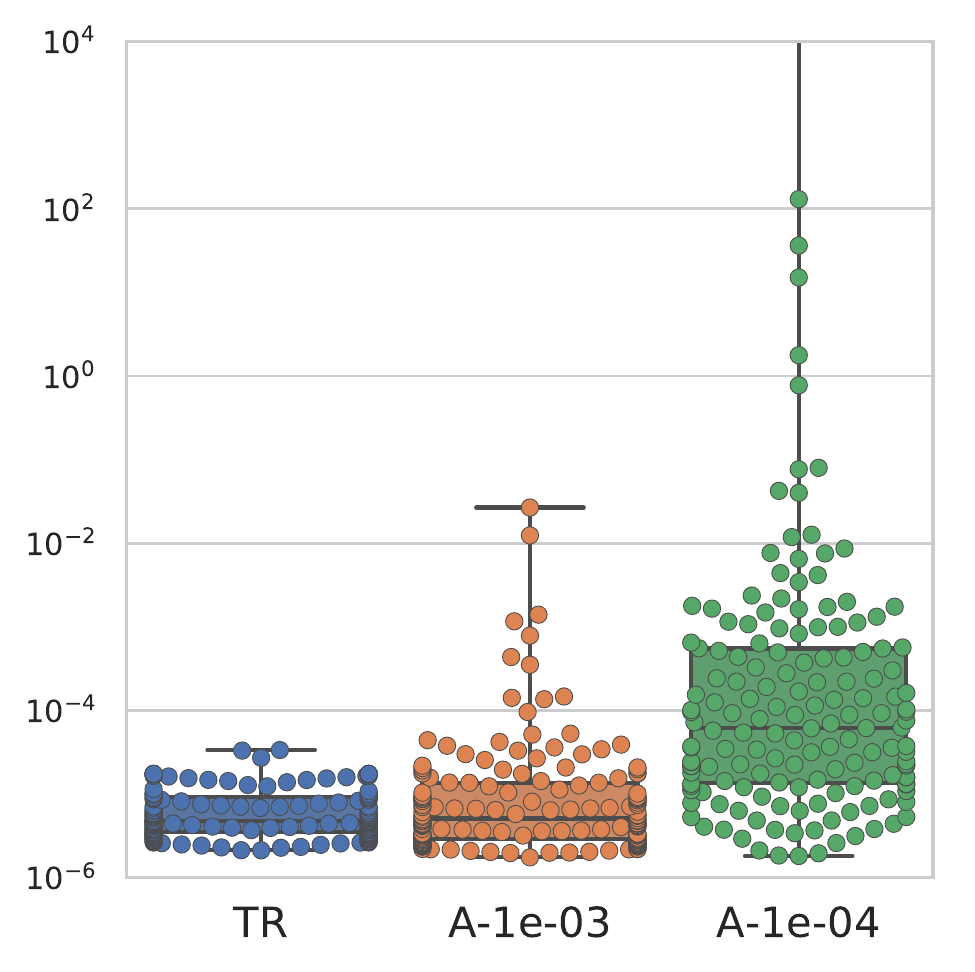}
        \includegraphics[width=0.3\textwidth]{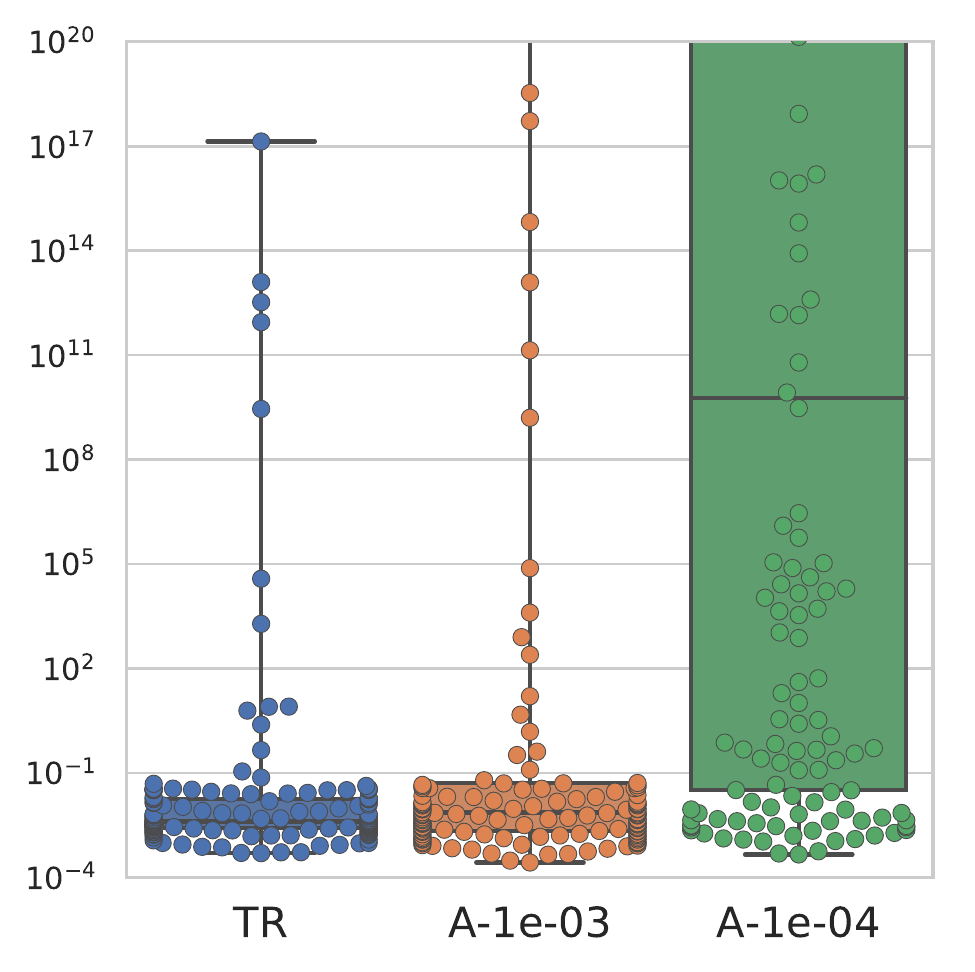}
        \caption{High Noise}
    \end{subfigure}
  \caption{\textbf{Noisy Case:} The minimum testing errors of the one-step (left), multi-step (middle) and 1000-step (right) predictions (aggregated over all configurations) by level of noise.}
  \label{fig.noisy_all}
\end{figure}

In this section, we present the experimental results of the noisy case. Figure \ref{fig.noisy_train} shows the evolution of the training errors of different configurations. Figure \ref{fig.noisy_u} shows the sequential predictions and the squared errors, and Figure \ref{fig.noisy_1000} shows the minimum testing errors (over the training process) of the 1000-step predictions by configuration. In addition, Figure \ref{fig.noisy_scatter} shows the relationship between the training and testing errors (note: the lower and to the left is better). For the two supplementary testing procedures described in Section \ref{app_sup_test}, given the data of solution at $t$, the one-step prediction is made at $t+\Delta t$, i.e., $t+1$, and the multi-step prediction is made at $t + 10\Delta t$, i.e., $t+10$. Figures \ref{fig.noisy_1_evol} \& \ref{fig.noisy_10_evol} show the evolutions of the testing errors \eqref{eq12} and Figures \ref{fig.noisy_1} \& \ref{fig.noisy_10} show the minimum testing errors of the one- and multi-step predictions. To summarize the testing performance, we put the minimum testing errors aggregated over all configurations in Figure \ref{fig.noisy_all}. 

\begin{figure}
    \centering
    \includegraphics[width=0.9\textwidth]{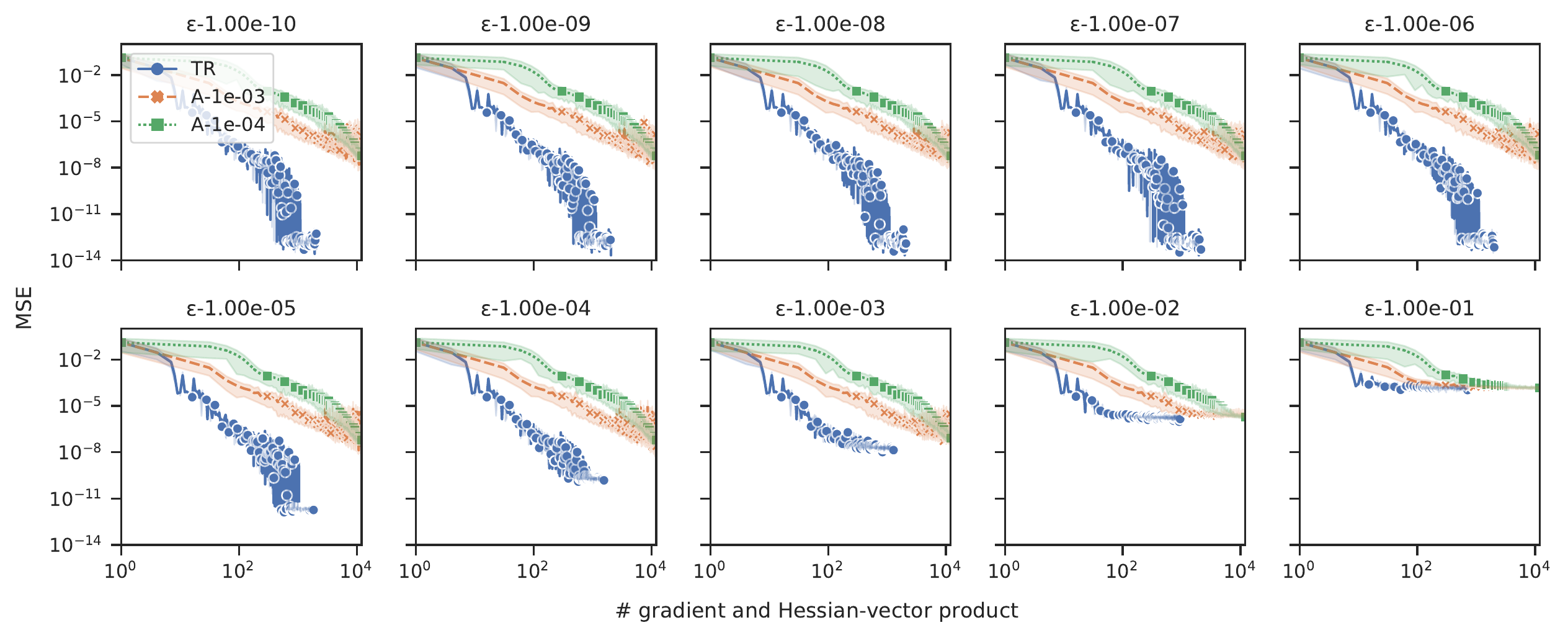}
  \caption{\textbf{Noisy Case:} Sensitivity analysis - the evolution of the MSE loss of stochastic mini-batch by level of noise.}
  \label{fig.all_noise_train}
\end{figure}

\begin{figure}
  \centering
    \begin{subfigure}{1\textwidth}
    \centering
        \includegraphics[width=0.9\textwidth]{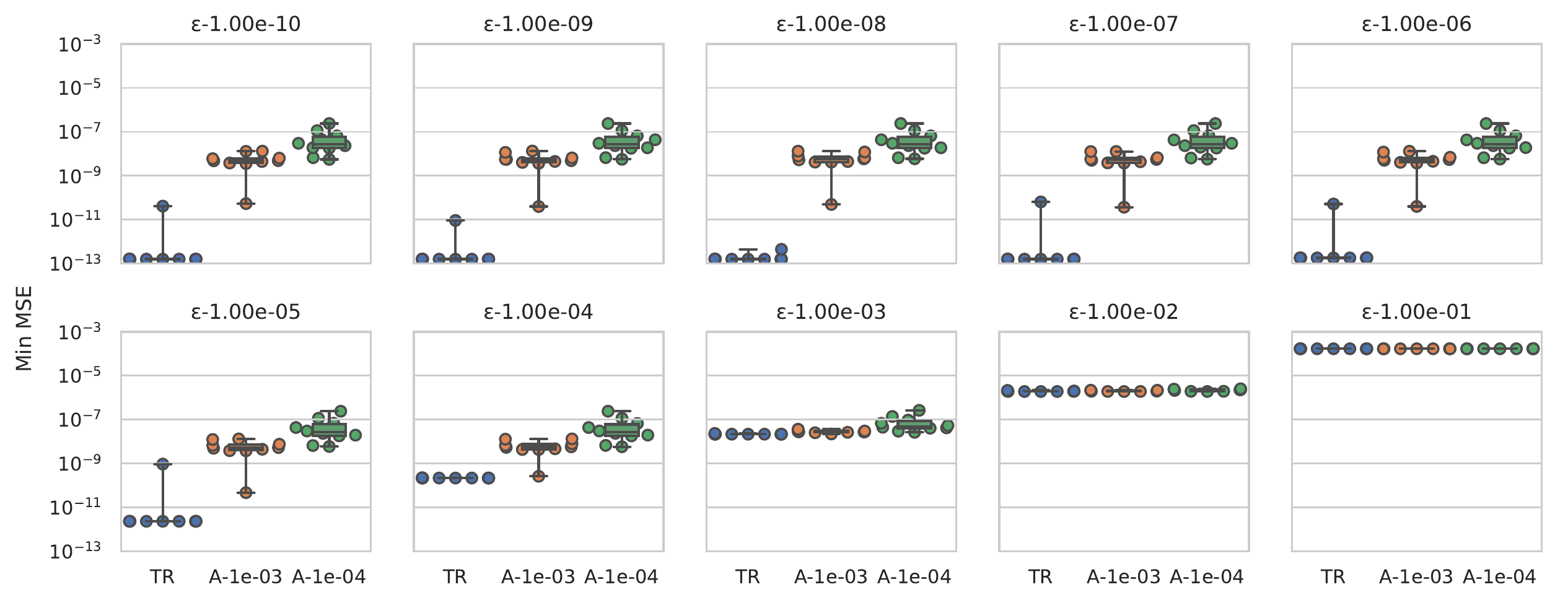}
        \caption{One-Step Prediction}
    \end{subfigure}\\[3ex]
    \centering
    \begin{subfigure}{1\textwidth}
    \centering
        \includegraphics[width=0.9\textwidth]{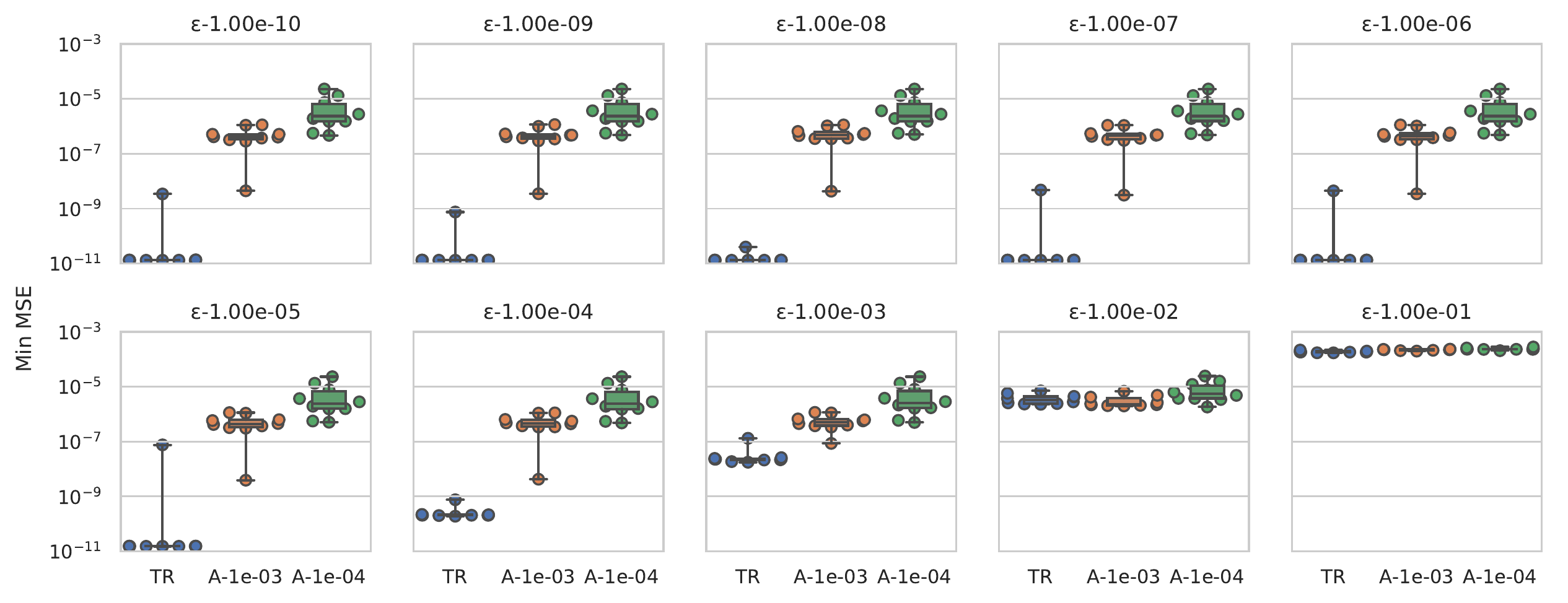}
        \caption{Multi-Step Prediction}
    \end{subfigure}\\[3ex]
    \centering
    \begin{subfigure}{1\textwidth}
    \centering
        \includegraphics[width=0.9\textwidth]{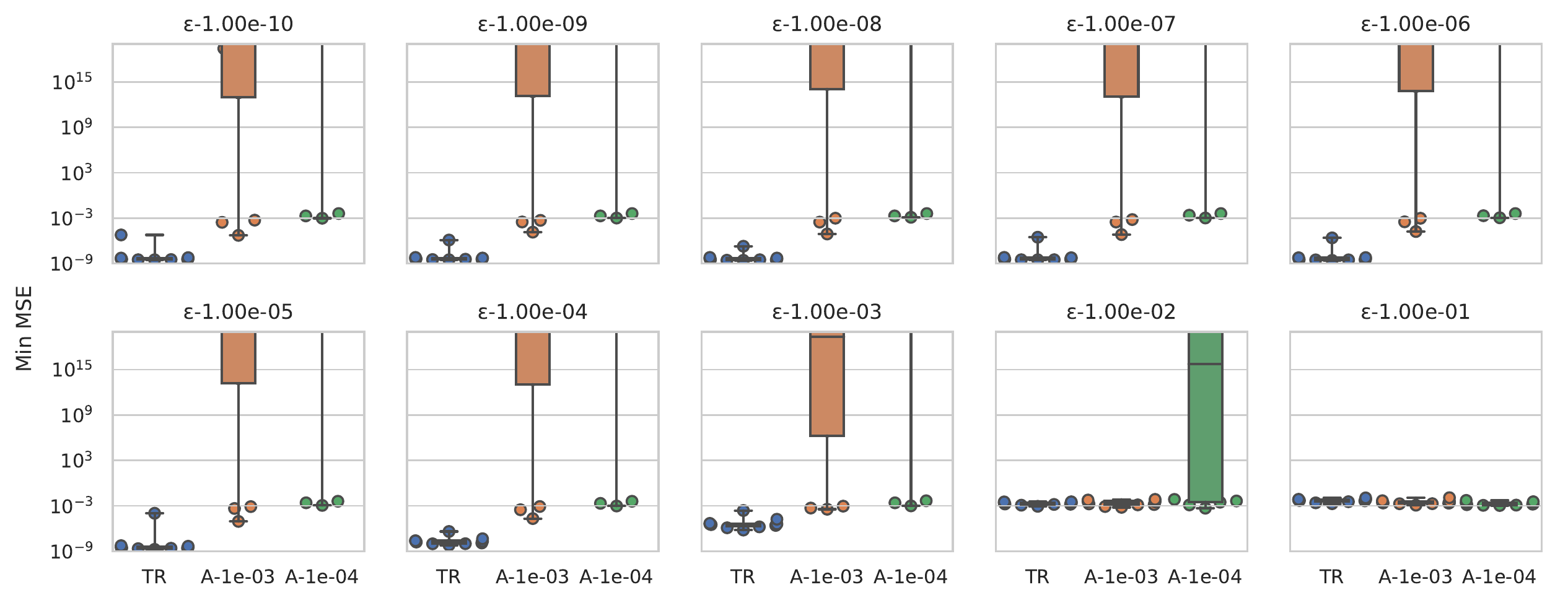}
        \caption{1000-Step Prediction}
    \end{subfigure}
  \caption{\textbf{Noisy Case:} Sensitivity analysis - the minimum testing errors by level of noise.}
  \label{fig.all_noise_1000}
\end{figure}

\begin{figure}[t!]
    \centering
    \includegraphics[width=0.9\textwidth]{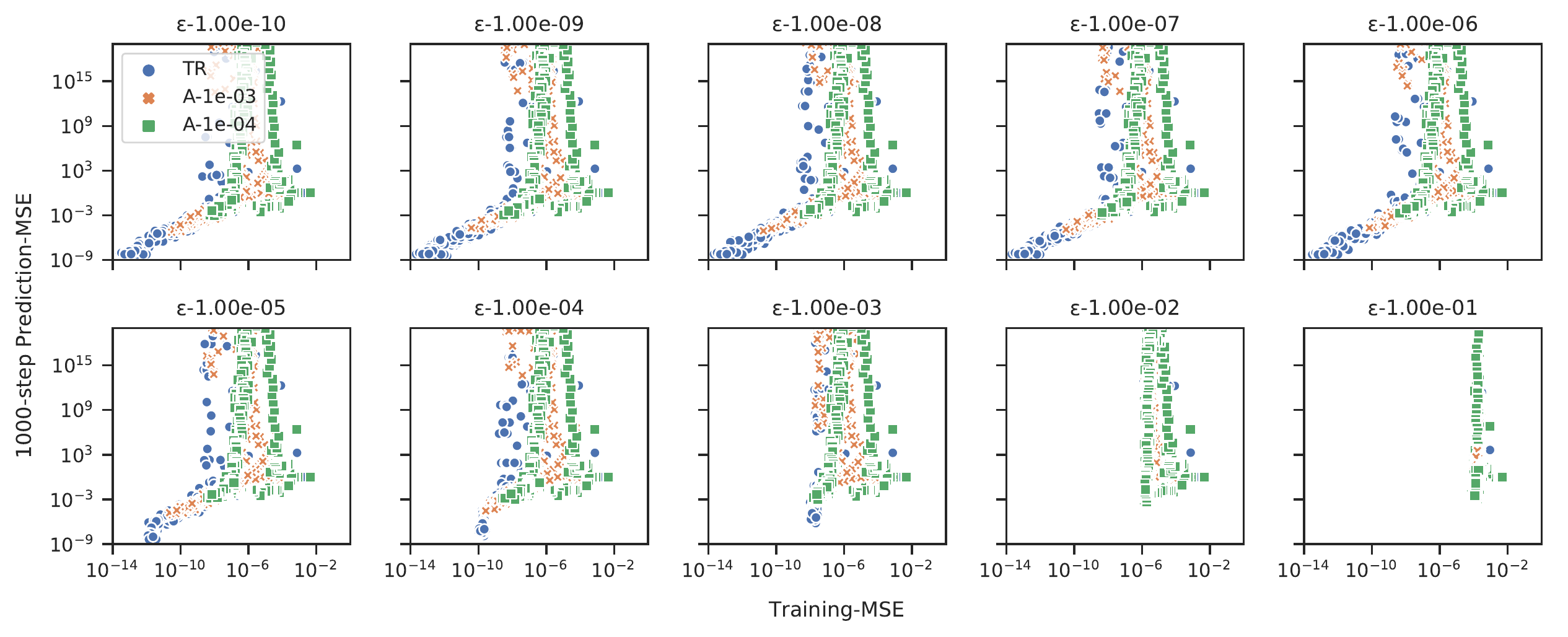}
  \caption{\textbf{Noisy Case:} Sensitivity analysis - training vs. 1000-step prediction (over the training process) by level of noise.}
  \vspace*{6in}
  \label{fig.all_noise_scatter}
\end{figure}

We further conducted experiments on the sensitivity of \FDNet{} to $10$ different levels of the multiplicative noise and used $1$ \FDBlock{} and 16 \FDFilters{} to configure the networks. Figure \ref{fig.all_noise_train} shows the evolution of the training errors, Figure \ref{fig.all_noise_1000} show the minimum testing errors over the training process, and Figure \ref{fig.all_noise_scatter} shows the relationship between the training and testing errors in the training process.


\clearpage

\end{document}